%% file: paper.tex
\documentclass[10pt,twocolumn,letterpaper]{article}

\usepackage{iccv}
\usepackage{times}
\usepackage{epsfig}
\usepackage{graphicx}
\usepackage{amsmath}
\usepackage{amssymb}
\usepackage{gensymb}
\usepackage{subfig}
\usepackage{float}
\usepackage{mathtools}
\usepackage{dblfloatfix}

\usepackage{comment}
\usepackage{array}

\usepackage{verbatim}
\usepackage{xfrac}
\usepackage{subfig}
\usepackage{mathtools}
\usepackage{combelow}
\usepackage{flushend}
\usepackage{booktabs, multicol, multirow}
\usepackage{pifont}
\usepackage[pagebackref=true,breaklinks=true,colorlinks,bookmarks=false]{hyperref}
\newcommand{\cmark}{\checkmark}%
\newcommand{\xmark}{}%
\usepackage{bbm}

\newcommand{\figref}[1]{Fig.~\ref{#1}}
\newcommand{\tabref}[1]{Tab.~\ref{#1}}
\newcommand{\secref}[1]{Sec.~\ref{#1}}

\newcommand{\VG}[1]{\textcolor{black}{#1}}

\iccvfinalcopy % *** Uncomment this line for the final submission

\def\iccvPaperID{3459} % *** Enter the ICCV Paper ID here

\ificcvfinal\fi

\begin{document}

\title{
%Sim2real Transfer using Geometric Self-Supervision
%Geometric Self-Supervision for Sim2real Transfer
%Geometric Self-Supervision for Unsupervised Domain Adaptation
%Self-Supervised Depth for Unsupervised Domain Adaptation
%Self-Supervised Depth as a Proxy Task for Unsupervised Domain Adaptation
%Self-Supervised Depth Helps Sim-to-Real Transfer of Semantic Segmentation
% Self-Supervised Depth for Multi-Task Unsupervised Domain Adaptation
Geometric Unsupervised Domain Adaptation for Semantic Segmentation
}

\author{
Vitor Guizilini \qquad Jie Li \qquad Rare\cb{s} Ambru\cb{s} \qquad Adrien Gaidon \\ % <-this % stops a space
Toyota Research Institute (TRI), Los Altos, CA \\
\texttt{\{first.lastname\}@tri.global}
}

\maketitle

\begin{abstract}
\input{sections/00abstract}
\end{abstract}

\section{Introduction}
\input{sections/01introduction}

\section{Related Work}
\input{sections/02relatedwork}

% \section{Mixed Batch Self-Supervised Adaptation}
\section{Geometric Unsupervised Adaptation}
%\section{Methodology}
\label{sec:methodology}

\input{sections/03methodology}

\section{Experimental Protocol}
\input{sections/04protocol}

\section{Experimental Results}
\input{sections/05experiments}
\section{Conclusion}

\input{sections/06conclusion}

% \input{tables/semantic_synthia_cs}
% \input{figures/scale_pd}

% \input{tables/depth_datasets}
% \input{tables/ablation}

% % % \input{tables/scaling_pd}

% \input{figures/synthia}

% \input{figures/camera}

% \input{figures/baselines}
% \input{figures/kitti}

\clearpage

\appendix

\input{suppmat}

{\small
\bibliographystyle{iccv_template/ieee_fullname}
\bibliography{references}
}

% \clearpage
% \input{suppmat}

\end{document}

%% file: sections/00abstract.tex
Simulators can efficiently generate large amounts of labeled synthetic data with perfect supervision for hard-to-label tasks like semantic segmentation. However, they introduce a domain gap that severely hurts real-world performance.
We propose to use self-supervised monocular depth estimation as a proxy task to bridge this gap and improve sim-to-real unsupervised domain adaptation (UDA).
Our Geometric Unsupervised Domain Adaptation method (GUDA)\footnote{\url{https://github.com/tri-ml/packnet-sfm}} learns a domain-invariant representation via a multi-task objective combining synthetic semantic supervision with real-world geometric constraints on videos.
GUDA establishes a new state of the art in UDA for semantic segmentation on three benchmarks, outperforming methods that use domain adversarial learning, self-training, or other self-supervised proxy tasks.
Furthermore, we show that our method scales well with the quality and quantity of synthetic data while also improving depth prediction.

\vspace{-5mm}

%% file: sections/01introduction.tex
\input{figures/teaser}
Self-supervised learning from geometric constraints is used to learn tasks like depth and ego-motion directly from unlabeled videos~\cite{flynn2016deepstereo,monodepth2,packnet,shu2020featdepth,zhou2018stereo}. 
However, tasks like semantic segmentation and object detection inherently require human-defined labels.
A promising alternative to expensive manual labeling is to use synthetic datasets~\cite{cabon2020vkitti2, desouza2019generating,gaidon2016virtual,Richter_2016_ECCV,ros2016synthia}. Simulators can indeed be programmed to generate large quantities of diverse data with accurate labels (cf.~\figref{fig:teaser}), including for optical and scene flow~\cite{Hur:2020:SSM,advcollab}, object detection~\cite{detection_da}, tracking~\cite{gaidon2016virtual}, action recognition~\cite{desouza2019generating}, and semantic segmentation~\cite{Richter_2016_ECCV,ros2016synthia}. 
However, no simulator is perfect. Hence, effectively using synthetic data requires overcoming the \emph{sim-to-real domain gap}, a distribution shift between a source synthetic domain and a target real one due to differences in content, scene geometry, physics, appearance, or rendering artifacts.
% that limits the applicability of models trained purely with synthetic supervision when transferred to a real-world setting.

The goal of \emph{Unsupervised Domain Adaptation} (UDA) is to improve generalization across this domain gap without any real-world labels. 
Most methods use adversarial learning for pixel or feature-level adaptation~\cite{bousmalis2017unsupervised,dann, hoffman2018cycada,spigan,dada_disc,Volpi_2018_CVPR, Yang2020LabelDrivenRF} or self-training by refining pseudo-labels~\cite{jin2018unsup,iast,usamr,crst,cbst,Saporta2020ESLES}. 
These methods yield clear improvements, but require learning multiple networks beyond the target one, are hard to train (adversarial learning), or limited to semantically close domains (iterative diffusion of high-confidence pseudo-labels).
Alternatively, few works~\cite{sun2019unsupervised, xu2019self} have explored simple image-level self-supervised proxy tasks~\cite{gidaris2018unsupervised, noroozi2016unsupervised, larsson2016learning} to improve generalization across domains, but with only limited success for UDA of semantic segmentation.

% In contrast, we introduce \emph{geometric constraints} commonly used in self-supervised monocular depth estimation for UDA,
In this work, we introduce \textbf{self-supervised monocular depth as a proxy task for UDA in semantic segmentation}.
% thus enabling scalable learning with limitless quantities of labeled synthetic data and raw real-world data.
We propose a multi-task mixed-batch training method combining synthetic supervision with a real-world self-supervised depth estimation objective to learn a domain-invariant encoder.
% Why non-trivial:
Although it is not obvious that \emph{geometric constraints on videos} can help overcome a \emph{semantic gap on images}, our method, called GUDA for Geometric Unsupervised Domain Adaptation, \textbf{outperforms other UDA methods for semantic segmentation}.
% below: not just those + need more space to explain subtlety (constraints / SSL vs sup) --> leave it for rel work.
% that use depth information~\cite{Chen_2019_CVPR,spigan,vu2019dada}.
%
Furthermore, we can directly combine our method with self-trained pseudo-labels, leading to a new state of the art on the standard SYNTHIA-to-Cityscapes benchmark. 
In addition, we show on Cityscapes~\cite{cordts2016cityscapes}, KITTI~\cite{Geiger2012CVPR}, and DDAD~\cite{packnet} that \textbf{our method scales well with both the quantity and quality of synthetic data} (cf.~\figref{fig:teaser}), from SYNTHIA~\cite{ros2016synthia} to VKITTI2~\cite{cabon2020vkitti2}, and a new large-scale high-quality dataset~\cite{parallel_domain}.
Finally, we show that GUDA is also capable of \textbf{state-of-the-art monocular depth estimation} in the real-world domain.

%\begin{itemize}

%\item We introduce \textbf{geometric self-supervision as a tool for unsupervised domain adaptation}, to bridge the gap between real and virtual datasets \textbf{without requiring supervision from real labels}.

%\item By training a \textbf{multi-task semantic and depth network} with a combination of virtual supervision and self-supervision from raw videos, we outperform other UDA methods that use depth as an extra source of supervision for the task of semantic segmentation \textbf{without requiring adversarial learning or additional networks at training or inference time}. 

%\item We show that our proposed GUDA framework can \textbf{use predictions from other UDA techniques} as pseudo-labels at training time to boost performance and achieve a \textbf{new state of the art in the SYNTHIA to Cityscapes benchmark for semantic segmentation}. 

%\end{itemize}

%% file: figures/teaser.tex
\begin{figure}[t!]
\vspace{-3mm}

\textit{
\hspace{0.6cm} SYNTHIA 
\hspace{1.0cm} VKITTI2 
\hspace{0.9cm} Parallel Domain
}
\vspace{-3mm}

% \centering

\subfloat{
\includegraphics[width=0.15\textwidth,height=1.3cm]{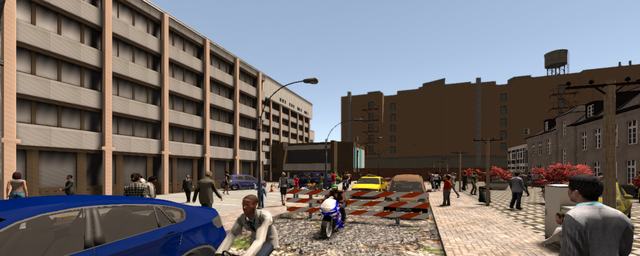}
\includegraphics[width=0.15\textwidth,height=1.3cm]{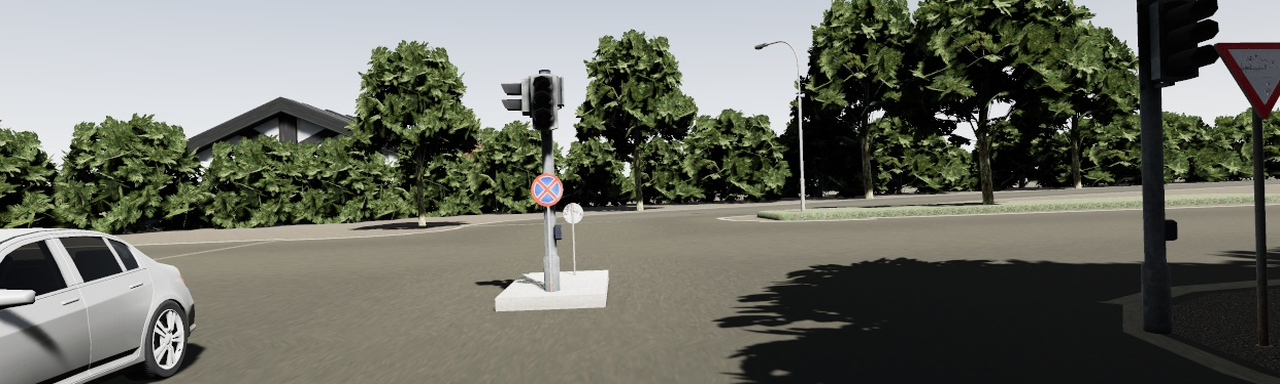}
\includegraphics[width=0.15\textwidth,height=1.3cm]{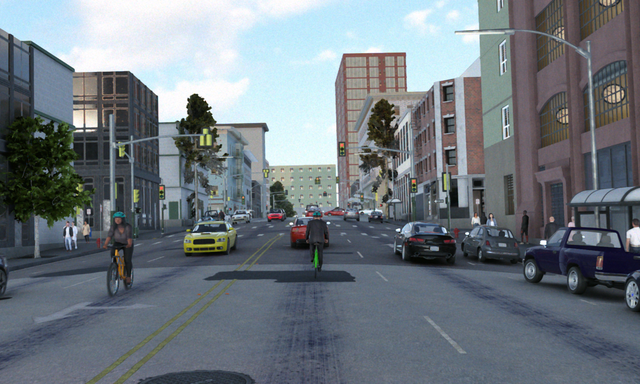} 
}
\vspace{-3mm}
\\ 
\subfloat{
\includegraphics[width=0.15\textwidth,height=1.3cm]{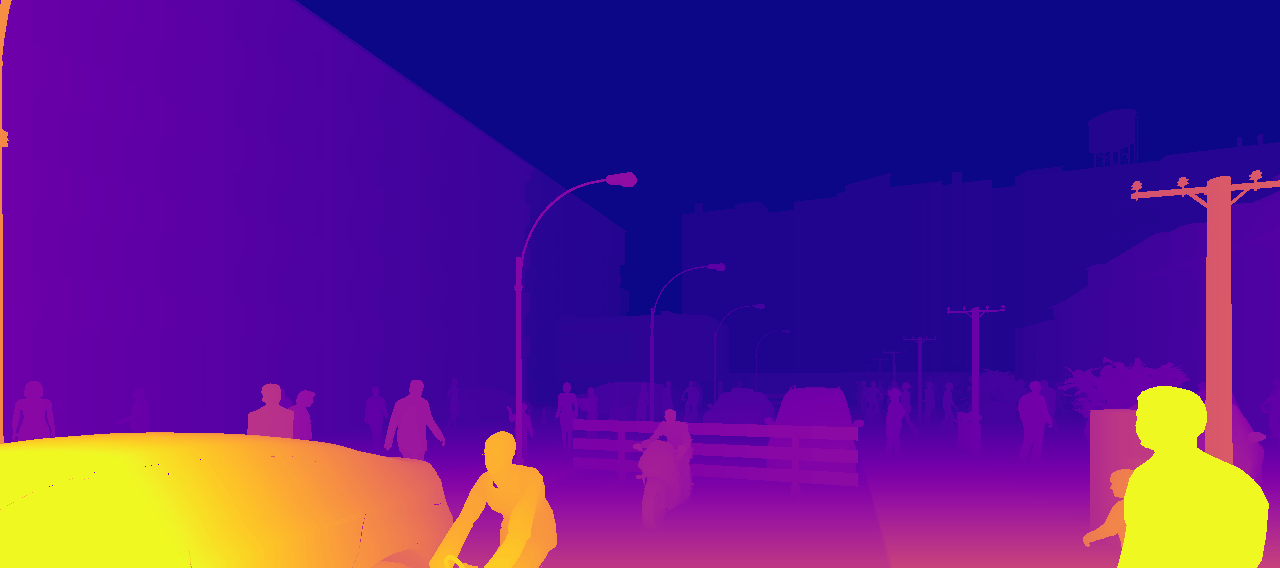}
\includegraphics[width=0.15\textwidth,height=1.3cm]{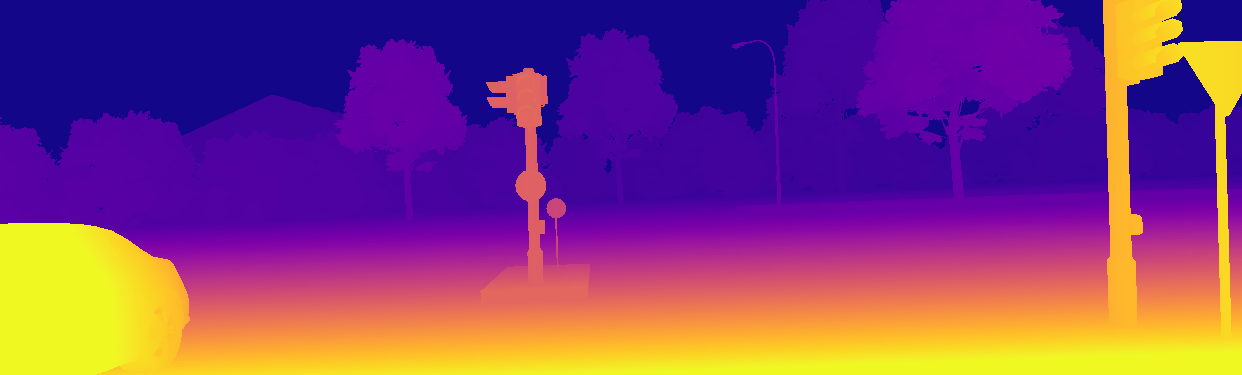}
\includegraphics[width=0.15\textwidth,height=1.3cm]{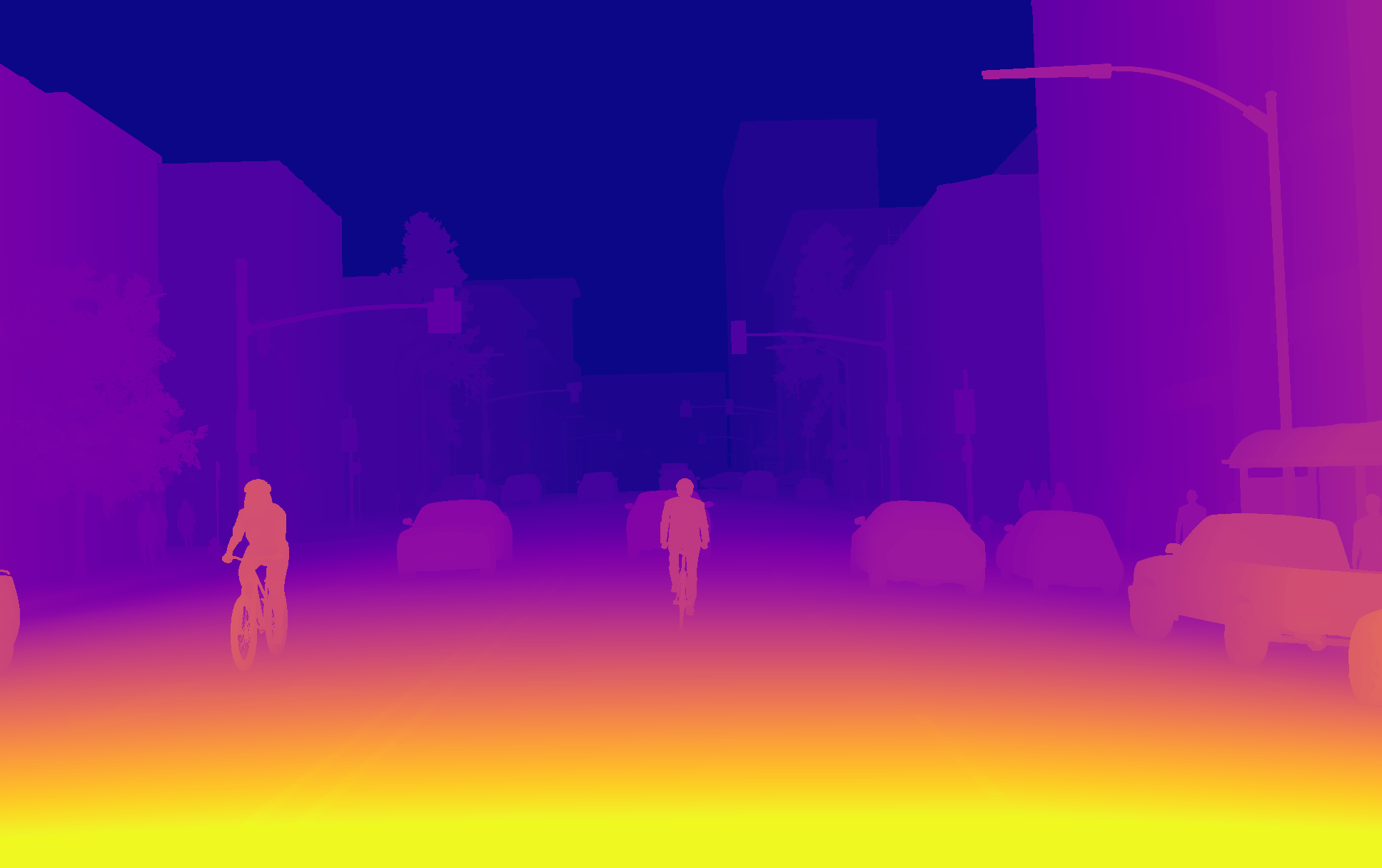}
}
\vspace{-3mm}
\\
\setcounter{subfigure}{0}
\subfloat[Synthetic datasets (ground-truth)]{
\includegraphics[width=0.15\textwidth,height=1.3cm]{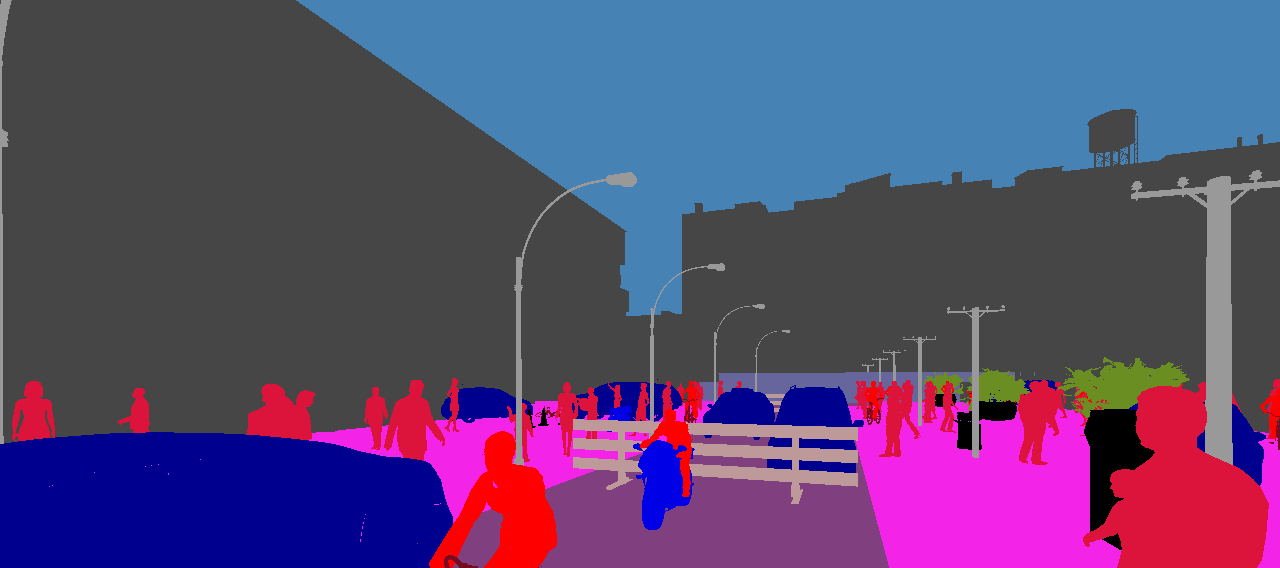}
\includegraphics[width=0.15\textwidth,height=1.3cm]{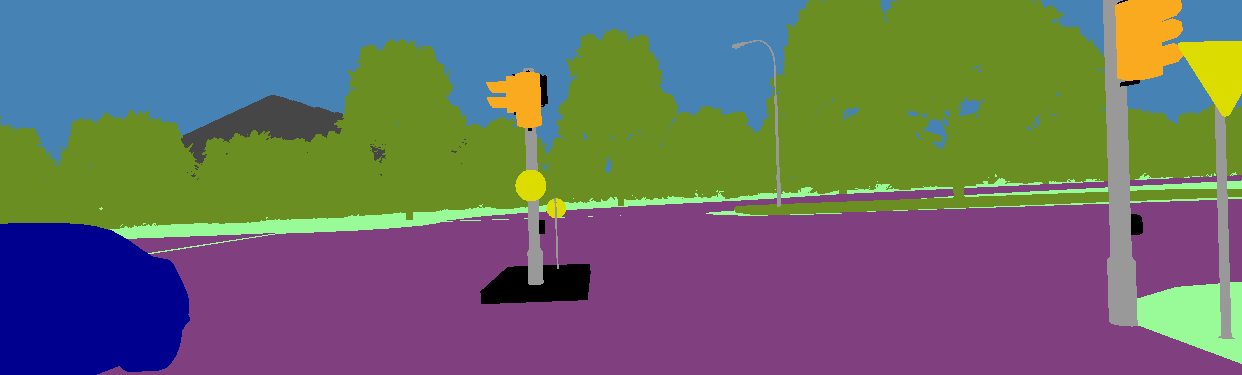}
\includegraphics[width=0.15\textwidth,height=1.3cm]{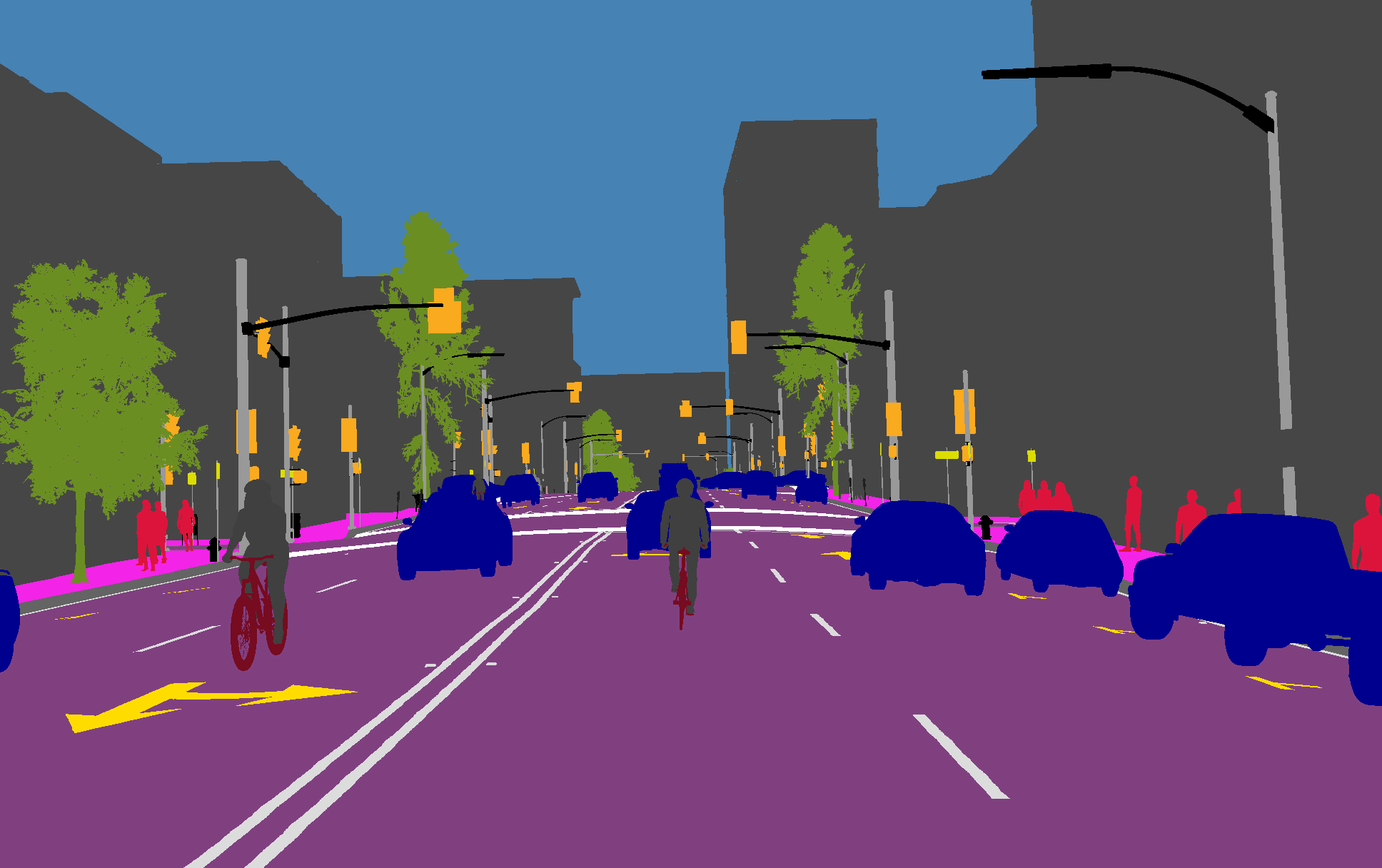}
}

\vspace{1mm}

\textit{
\hspace{0.5cm} Cityscapes 
\hspace{1.3cm} KITTI 
\hspace{1.6cm} DDAD
}
\vspace{-3mm}

\subfloat{
\includegraphics[width=0.15\textwidth,height=1.3cm]{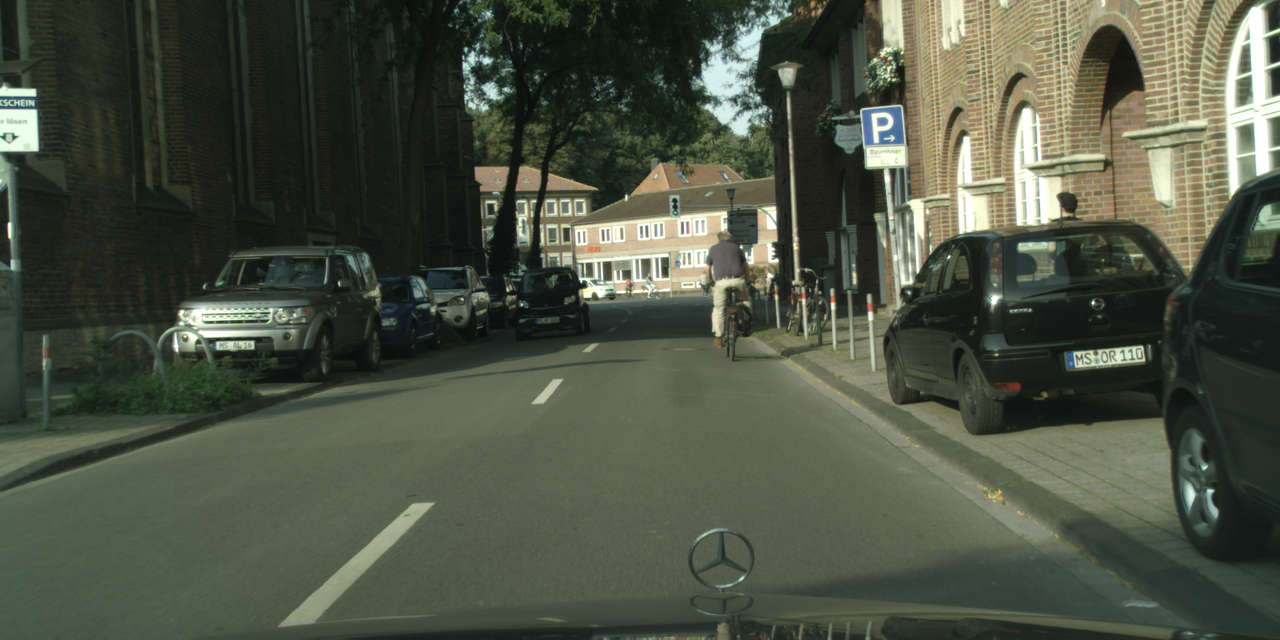}
\includegraphics[width=0.15\textwidth,height=1.3cm]{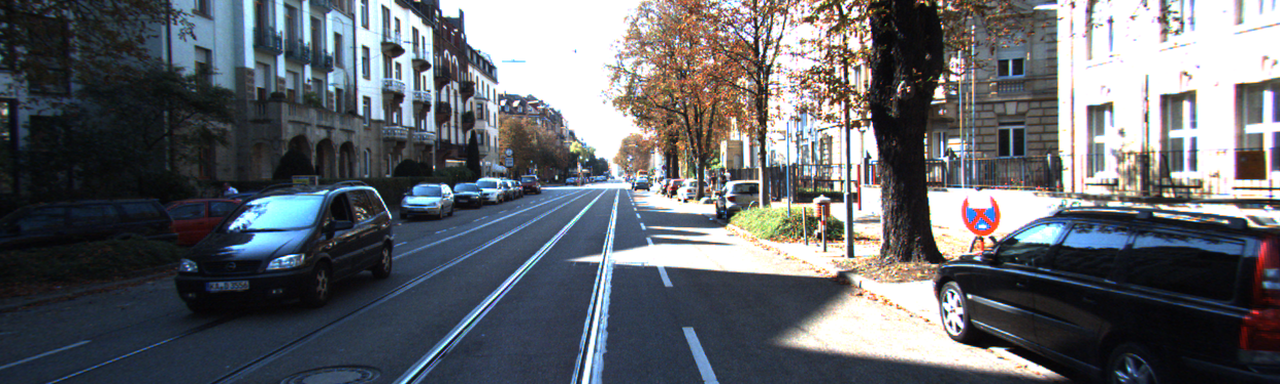}
\includegraphics[width=0.15\textwidth,height=1.3cm]{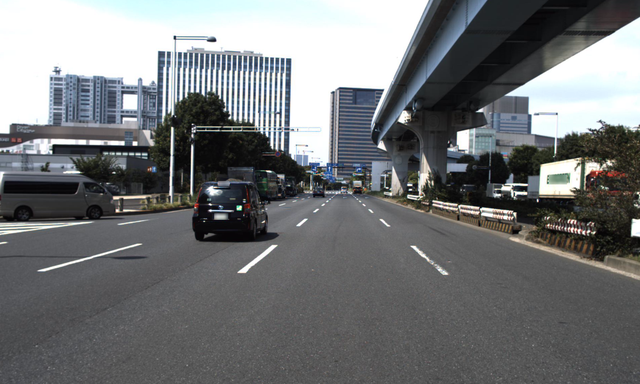}
}
\vspace{-3mm}
\\ 
\subfloat{
\includegraphics[width=0.15\textwidth,height=1.3cm]{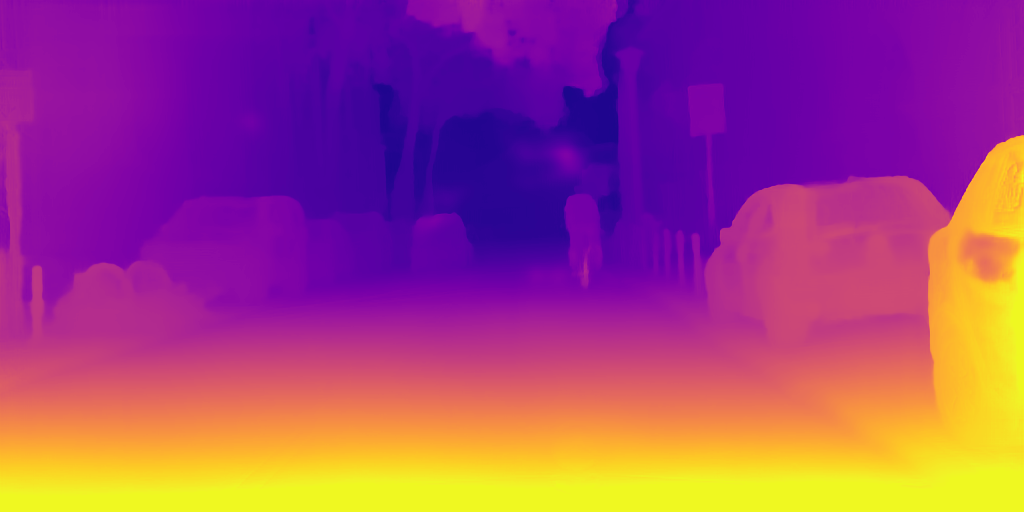} 
\includegraphics[width=0.15\textwidth,height=1.3cm]{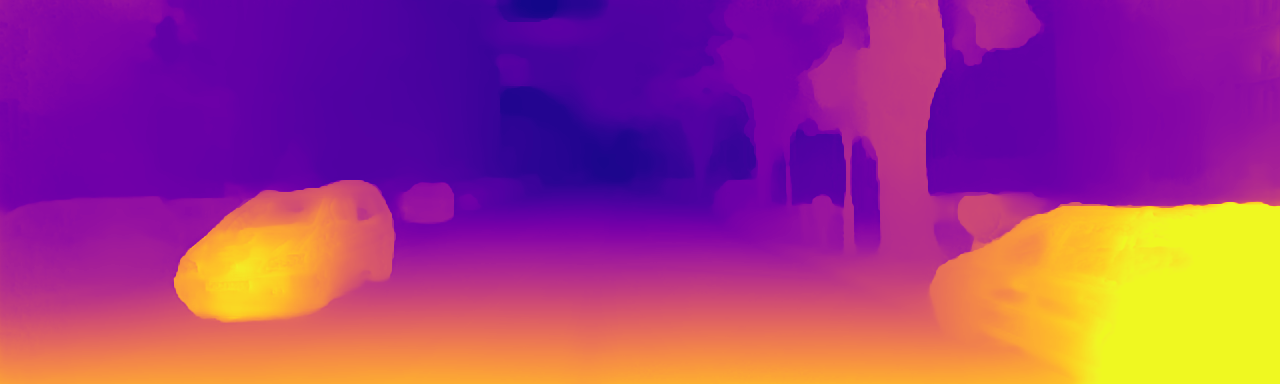}
\includegraphics[width=0.15\textwidth,height=1.3cm]{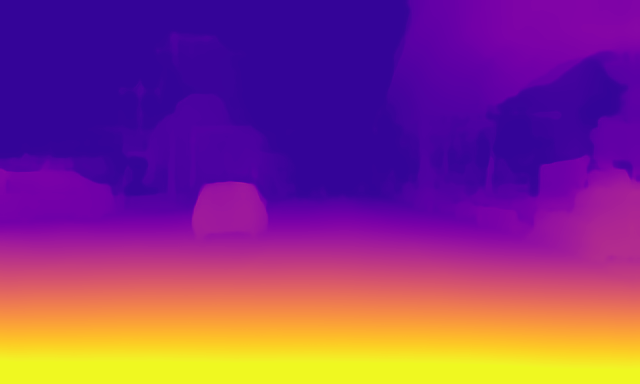}
}
\vspace{-3mm}
\\ 
\setcounter{subfigure}{1}
\subfloat[Real-world datasets (predictions)]{
\includegraphics[width=0.15\textwidth,height=1.3cm]{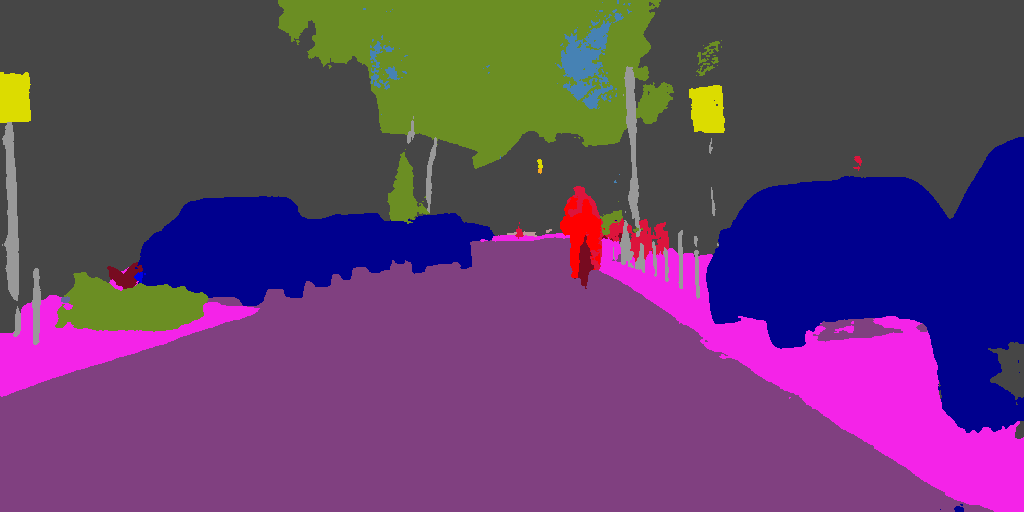} 
\includegraphics[width=0.15\textwidth,height=1.3cm]{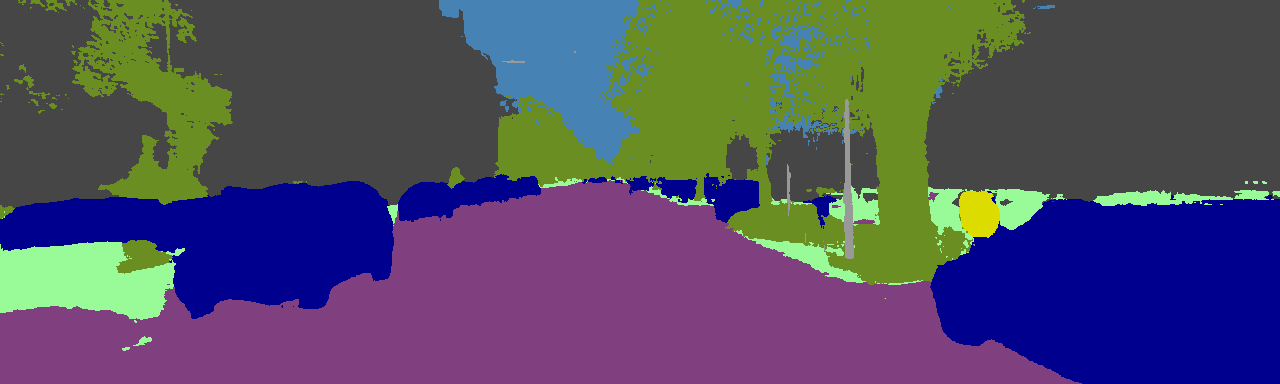}
\includegraphics[width=0.15\textwidth,height=1.3cm]{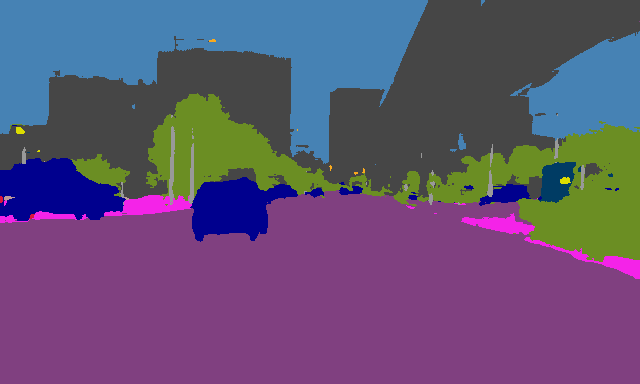}
}
\vspace{-2mm}
% \\
% \subfloat[Input Image]{\includegraphics[width=0.22\textwidth]{images/teaser/media_images_val-cityscapes-val_semseg-stereo-480-rgb_13_7888a1a6.png}} 
% \hspace{2mm}
% \subfloat[Ground-truth semantics]{\includegraphics[width=0.22\textwidth]{images/teaser/media_images_val-cityscapes-val_semseg-stereo-480-semantic_13_ae07d3f7.png}} 
% \vspace{-3mm}
% \\
% \subfloat[Predicted depth]{\includegraphics[width=0.22\textwidth]{images/teaser/media_images_val-cityscapes-val_semseg-stereo-480-inv_depth_30_9e87d512.png}} 
% \hspace{2mm}
% \subfloat[Predicted semantics]{\includegraphics[width=0.22\textwidth]{images/teaser/media_images_val-cityscapes-val_semseg-stereo-480-pred_semantic_6_49b617f9.png}} 
\caption{\textbf{Our GUDA approach} uses geometric self-supervision on \emph{videos} in a multi-task setting to achieve state-of-the-art results in UDA for semantic segmentation.}
% \caption{\textbf{Multi-task predictions} obtained using our proposed approach to unsupervised domain adaptation using geometric self-supervision and pseudo-labels.}
\label{fig:teaser}
\vspace{-6mm}
\end{figure}

%% file: sections/02relatedwork.tex
\subsection{Unsupervised Domain Adaptation
}
% General unsupervised domain adaptation

% Unsupervised domain adaptation (UDA) is an active research area in machine learning. It aims to decrease the gap in model performance between source and target datasets due to differences in data distribution~\cite{wang2018deep, wilson2020survey,patel2015visual}.

Unsupervised domain adaptation (UDA) is an active research area in Computer Vision~\cite{csurka2017domain, wang2018deep, wilson2020survey}. Its main goal is to learn a model on a labeled \emph{source} dataset and a related but statistically different unlabeled \emph{target} dataset where the model is expected to generalize. Common approaches rely on domain-invariant learning or statistical alignment between domains~\cite{dann, tzeng2014deep, yan2017learning}.
% TODO: [Jie] Maybe a bit more discussion here is we have space
% semantic segmentation
In this work, we consider UDA for \emph{semantic segmentation}, for which the labeling process is tedious and expensive. Several synthetic datasets have been proposed to reduce the need for real-world labels, such as SYNTHIA~\cite{ros2016synthia}, GTA5~\cite{Richter_2016_ECCV}, and Virtual KITTI~\cite{gaidon2016virtual}. 

However, models trained on such datasets suffer from a significant performance drop when tested on real-world datasets. 
To overcome this large \emph{sim-to-real} generalization gap, several works have proposed to use \emph{adversarial learning} for distribution alignment at the level of pixels~\cite{hoffman2018cycada,wu2018dcan,fda, zhang2018fully, Yang2020LabelDrivenRF}, features~\cite{hoffman2016fcns,saito2017adversarial,huang2018domain}, or outputs~\cite{adapt_patch, vu2018advent,saito2018maximum}.
% CyCADA~\cite{hoffman2018cycada} and DCAN~\cite{wu2018dcan} use generative adversarial networks (GANs) to align pixel-level distributions between target and source domains.
% FDA~\cite{fda} adapts the Fourier transform to conduct style-transfer on input images. Alternatively, some approaches conduct alignment on the output feature space~\cite{adapt_patch,vu2018advent}.
%
Alternatively, \emph{self-training} (a.k.a. \emph{pseudo-labeling}) has also been successful for UDA~\cite{cbst, crst, cag, iast, usamr, Saporta2020ESLES}. In these works, models are iteratively trained with both ground truth labels in the source domain and inferred pseudo-labels in the target domain, updated as part of an optimization loop.
% no need for details on pseudo-labels (not related to our work)
%Different approaches were proposed to improve the quality of pseudo-labels.
% CBST~\cite{cbst} enforces class balance to avoid the dominance of larger classes, and spatial priors to further refine the pseudo-labels.
% CRST~\cite{crst} improves on top of \cite{cbst} using a pseudo-label entropy regularizer.
% CAG~\cite{cag} filters pseudo labels with scale-invariant properties. 
% IAST~\cite{iast} introduces an instance adaptive module to prevent pseudo-label bias towards easy classes. 
% USAMR \cite{usamr} exploits intra-domain uncertainty using model memory regularization.

% other sources in domain adaptation

% To address 

Although still underexplored, another promising direction is the use of other modalities in the \emph{source} domain to help the unsupervised transfer of semantic segmentation to the target domain.
SPIGAN~\cite{spigan} uses synthetic depth as privileged information to provide additional regularization during adversarial training.
GIO-Ada~\cite{Chen_2019_CVPR} uses geometric information, including depth and surface normals, during style-transfer in the target domain.
DADA~\cite{vu2019dada} predicts depth and semantic segmentation during adversarial training using a shared encoder, and fuses  depth-aware features to improve semantic segmentation predictions.
%Why we are different

% These works provide strong baselines for comparison to our proposed GUDA framework in the context of using depth information. 

%Our method builds on the insights of~\cite{spigan,Chen_2019_CVPR,vu2019dada}, and we also employ dense depth supervision in the source domain. However, they only do so in the source domain, as explicit supervision or to enforce additional constraints during the adaptation stage, while transfer to the target domain is achieved via adversarial adaptation or style transfer. In contrast to~\cite{spigan,Chen_2019_CVPR,vu2019dada}, we also consider depth in the \emph{target} domain through \emph{self-supervision} from geometric cues, and use it as the primary source of domain adaptation. By simultaneously learning the same task in both domains, we produce features that: (i) are capable of performing both tasks (i.e. depth estimation and semantic segmentation), and (ii) are robust to differences in distribution between target and source domains. 

In contrast to~\cite{Chen_2019_CVPR,spigan,vu2019dada}, we do not only use \emph{source} depth information as explicit supervision or to enforce additional constraints during the adaptation stage. Instead, we \emph{infer} and leverage depth in the \emph{target} domain through \emph{self-supervision} from geometric \emph{video-level} cues, and use it as the primary source of domain adaptation. By simultaneously learning depth estimation in both domains, we produce features that are discriminative enough to perform this task while being robust to general differences in distribution between target and source domains. 

%as explicit supervision or to enforce additional consistency during the adaptation stage. Ours, on the other hand, also considers depth in the target domain through self-supervision from geometric cues, and use it as the primary source of adaptation. Because of that, we eliminate the need for additional networks, often trained using adversarial learning that is notoriously unstable and creates extra computational costs. 

\subsection{Self-Supervised Learning}

%\textcolor{red}{Self-supervised learning also has been applied to domain adaptation (talk about colorization and other techniques). Then move on to self-supervised depth and ego-motion learning. }
% self supervised learning
Self-supervised learning (SSL) has recently shown promising results in feature extraction through the definition of auxiliary tasks, using only unlabeled data as input~\cite{gidaris2018unsupervised,larsson2016learning,noroozi2016unsupervised,wolfram-ssl}. Typical auxiliary tasks look at different ways of reconstructing input data, such as rotation~\cite{gidaris2018unsupervised},  patch jigsaw puzzles~\cite{noroozi2016unsupervised}, or image colorization~\cite{larsson2016learning}.
Only few works have used SSL as a tool for domain adaptation \cite{ghifary2016deep, carlucci2019domain, sun2019unsupervised, xu2019self}, reporting results far from the state of the art (cf. Tab. \ref{tab:small_table}).
\cite{ghifary2016deep} proposes an auxiliary image reconstruction task on the target domain for image classification~\cite{ghifary2016deep}, whereas~\cite{carlucci2019domain,sun2019unsupervised,xu2019self} explore different image-level pretext tasks to improve standard adversarial domain adaptation.

In this paper, we build upon recent developments in self-supervised learning in \emph{videos} for monocular \emph{depth} and \emph{ego-motion} estimation \cite{monodepth2,packnet,shu2020featdepth,zhou2017unsupervised}. We show that these SSL tasks help UDA, leveraging strong geometric priors from videos to adapt features in a multi-task setting. 

%In this paper we introduce a new SSL tool that leverages recent progresses in self-supervised learning for monocular depth estimation 

%In this paper we introduce self-supervised geometric priors as a tool for unsupervised domain adaptation, 
%Nevertheless, none of these methods claims the most recent state-of-the-art in semantic segmentation, indicating that it is non-trivial to apply SSL techniques in Domain Adaptation algorithms.
%Besides, all these methods consider SSL tasks designed based on single frame image purely. In this method, we proposed a novel approach to adopt SSL in domain adaptation context leveraging different modality and go beyond single frame setting.

%% file: sections/03methodology.tex
\input{figures/diagram2}

A diagram of our proposed architecture for geometric unsupervised  domain adaptation (GUDA) is shown in \figref{fig:diagram}. It is composed of three networks: \emph{Depth} $f_D : I \rightarrow \hat{D}$, that takes an input image $I$ and outputs a predicted depth map $\hat{D}$; \emph{Semantic} $f_S : I \rightarrow \hat{S}$, that takes the same input image and outputs a predicted semantic map $\hat{S}$; and \emph{Pose} $f_T : \{ I_a, I_b \} \rightarrow \hat{T}_a^b$, that takes a pair of images and outputs the rigid transformation $\hat{T}$ between them.  The depth and semantic networks share the same encoder $f_E : I \rightarrow \hat{F}$, such that $f_D : f_E(I) \rightarrow \hat{D}$ and $f_S : f_E(I) \rightarrow \hat{S}$ both decode the same latent features $\hat{F}$ into their respective tasks. 
See \secref{sec:archi} for architecture details.
%Following related work~\cite{zhou2017unsupervised}, the depth network outputs $4$ different scales with increasing resolution.

%This is a key aspect of GUDA, as it enables the adaptation of semantic features from a synthetic domain using depth and ego-motion losses calculated in the target domain. Our hypothesis is that, since depth and semantic decoders share the same encoded features, the adaptation generated by jointly training depth in both source and target domains will also adapt the shared semantic features. Because the network features themselves are adapted, GUDA does not require additional translation networks for domain alignment. Furthermore, by considering only self-supervision from monocular structure-from-motion cues in the target domain, we completely eliminate the need for real-world semantic or depth labels.
During training we employ a mixed-batch approach, in which at each iteration real $\mathcal{B}_R$ and virtual $\mathcal{B}_V$ batches are received and processed to generate corresponding real $\mathcal{L}_R$ (\secref{sec:real_sample}) and virtual $\mathcal{L}_V$ (\secref{sec:virtual_sample}) losses, depending on the available information. The final loss is defined as:
\begin{equation}
\mathcal{L} = \mathcal{L}_{R} + \lambda_V \mathcal{L}_{V}
\end{equation}
where $\lambda_V$ is a coefficient used to balance real and virtual losses during optimization. The next sections detail how each of these losses is calculated.

% \begin{align} 
% \hat{D}_t &= f_D(I_t)  \\
% T^{t-1}_t &= f_T(I_t, I_{t-1})  \\
% \hat{S}_t &= f_S(I_t) \\
% \end{align}

\subsection{Real (Target) Sample Processing}
\label{sec:real_sample}

Real-world samples are assumed to contain only unlabeled image sequences $\mathcal{I}_t$, in the form of the current frame $I_t$ and a temporal context $\{I_{t-s}, \cdots, I_{t+s} \}$. In all experiments we considered a temporal context of $s=1$, resulting in $\mathcal{I}_t = \{ I_{t-1}, I_t, I_{t+1} \}$. For simplicity, we also assume known and constant camera intrinsics $K$ for all frames, however this assumption can be relaxed to include the simultaneous learning of the projection model~\cite{gordon2019depth,vasiljevic2020neural}. Following~\cite{sun2019unsupervised}, we use an auxiliary self-supervised task in the target domain to help adapt features learned in the source domain. Specifically, depth and ego-motion learning via self-supervised photometric consistency in videos has been shown to produce results competitive with supervised learning in some domains~\cite{packnet,zhou2017unsupervised}. Leveraging this insight, we define our target domain loss as:
% Since there are no explicit labels for supervised training, we focus instead on geometric cues from image sequences, which has been shown to provide enough supervisory signals for depth and pose learning~\cite{zhou2017unsupervised}. The final loss is of the form:
\begin{equation}
% \mathcal{L}_{R}(\mathcal{I}) = \mathcal{L}_{P}(\mathcal{I}, \hat{D}, \hat{\mathcal{T}})
\mathcal{L}_{R} = \mathcal{L}_{P} + \lambda_{PL} \mathcal{L}_{PL}
\end{equation}
where $\mathcal{L}_{P}$ is the self-supervised photometric loss described in Section \ref{sec:self-sup-loss}, and $\mathcal{L}_{PL}$ is an optional pseudo-label loss described in Section \ref{sec:pseudo-label-loss}, with weight coefficient $\lambda_{PL}$.

% \begin{align}
% \mathcal{L}_{depth}(\mathcal{I}) = & \\
% \mathcal{L}_{photo}(\mathcal{I}) = &
% \mathcal{L}_{normal}(\mathcal{I}) = &
% \end{align}

\subsubsection{Self-Supervised Photometric Loss}
\label{sec:self-sup-loss}

Following previous works~\cite{garg2016unsupervised, zhou2017unsupervised}, the self-supervised depth and ego-motion objective can be formulated as a novel view synthesis problem, in which a target image $I_t$ is reconstructed using information from a reference image $I_{t'}$ given a predicted depth map $\hat{D}_t$ and relative transformation $T_t^{t'}$ between images:
\begin{equation}
\label{eq:projection}
\hat{I}_t = I_{t'} \Big\langle \pi \Big( \hat{D}_t, \hat{T}^{t'}_t, K \Big) \Big\rangle
\end{equation}
where $\pi$ is the \emph{projection operation} determined by camera geometry and $\langle \rangle$ is the \emph{bilinear sampling operator}, that is locally sub-differentiable and thus can be used as part of an optimization pipeline. To measure the reconstruction error we use the standard photometric loss~\cite{photo_loss}, with a structural similarity (SSIM) component~\cite{wang2004image} and the L1 distance in pixel space, weighted by $\alpha=0.85$:
\begin{equation}
\small
\mathcal{L}_P(I_t, \hat{I}_t) = \alpha \frac{\Big( 1 - \text{SSIM}(I_t, \hat{I}_t) \Big)}{2} + (1 - \alpha) ||I_t - \hat{I}_t||_1
\end{equation}
This loss is calculated for every image $I_{t'} \in \mathcal{I}_t$ and averaged for all pixels between multiple scales, after upsampling to the highest resolution. Following~\cite{monodepth2}, we use auto-masking and minimum reprojection error to mitigate effects caused by occlusions and dynamic objects.

\subsubsection{Pseudo-Label Distillation}
\label{sec:pseudo-label-loss}
Self-training methods~\cite{iast,usamr,cbst} are currently the dominant framework to address unsupervised domain adaptation for several different tasks~\cite{roychowdhury2019selftrain}. They work by iteratively refining high-confidence pseudo-labels in the target domain using supervision in the source domain. This source of domain adaptation can in principle be used to augment our proposed domain adaptation via geometric self-supervision.

Here we propose a simple yet effective way to distill information from self-training methods (or any other UDA method) into GUDA, by using pre-calculated pseudo-labels as supervision in the target domain. The resulting loss is similar to the supervised semantic loss described in \secref{sec:sup-semantic-loss}, using the predicted semantic map $\hat{S}$ from the real sample and the pseudo-label $S^{PL}$ pre-calculated from the same input image $I$ as ground-truth:
\begin{equation}
\mathcal{L}_{PL} = \mathcal{L}_S (\hat{S}, S^{PL})
\end{equation}
In our ablation analysis (\tabref{tab:ablation}), we show that the combination of GUDA with pseudo-label supervision from a self-training method~\cite{usamr} achieves the best results, surpassing other methods and establishing a new state of the art in unsupervised domain adaptation for semantic segmentation.

\subsection{Virtual (Source) Sample Processing}
\label{sec:virtual_sample}

Virtual samples consist of input images $I_t$ with corresponding dense annotations for all considered tasks, i.e. depth maps $D_t$ and semantic labels $S_t$. If sequential data is available we also assume a temporal context $\mathcal{I}_t = \{ I_{t-1}, I_t, I_{t+1} \}$, corresponding ground-truth rigid transformation between frames $\mathcal{T}_t= \{ T_t^{t-1}, T_t^{t+1} \}$, and constant camera intrinsics $K$. The availability of supervision allows the learning of both semantic and depth tasks in the source domain, encoding this information into the shared encoder $f_E$ and the respective decoders $f_D$ and $f_S$. We define our source domain loss as follows:
\begin{equation}
\mathcal{L}_{V} = \mathcal{L}_{D} + \lambda_S \mathcal{L}_S + \lambda_N \mathcal{L}_{N} + \lambda_{PP} \mathcal{L}_{PP}
\end{equation}
where $\mathcal{L}_S$ is a supervised semantic loss (Sec.~\ref{sec:sup-semantic-loss}), $\mathcal{L}_D$ is a supervised depth loss (\secref{sec:sup-depth-loss}), $\mathcal{L}_N$ is a surface normal regularization term (Sec. \ref{sec:sup-normal-loss}), and $\mathcal{L}_{PP}$ is an optional partially-supervised photometric loss (Sec.~\ref{sec:partial-sup-loss}), each weighted by their corresponding coefficient.

\subsubsection{Supervised Semantic Loss}
\label{sec:sup-semantic-loss}

Following \cite{wu2016bridging-bootstrapped,pohlen2017full-bootstrapped}, we supervise semantic segmentation in the source domain using a bootstrapped cross-entropy loss between predicted $\hat{S}$ and ground-truth $S$ labels:
% \begin{equation}
% \mathcal{L}_S = - \frac{1}{HW} \sum_{u,v}^{H,W} \sum _{c=1}^C \mathbbm{1}_{ [ c=y_{u,v} ] } \log \Big( T(x;\theta_T)_{u,v} \Big)
% \end{equation}
\begin{equation}
\mathcal{L}_S = - \frac{1}{K} \sum_{u=1}^{H} \sum_{v=1}^{W} \sum _{c=1}^C \mathbbm{1}_{ [ c=y_{u,v}, p_{u,v}^c<t]} \log \Big( p_{u,v}^c \Big)
\end{equation}
where $p_{u,v}^c$ denotes the predicted probability of pixel $(u,v)$ belonging to class $c$. The term $t$ is a run-time threshold so that only the worst $K$ performing predictions are considered. We adopt $K=0.3\times H\times W$ in our experiments.

\input{figures/normals}

\subsubsection{Supervised Depth Loss}
\label{sec:sup-depth-loss}

Our supervised objective loss is the \textit{Scale-Invariant Logarithmic} loss (SILog)~\cite{eigen2014depth}, composed by the sum of the variance and the weighted squared mean of the error in log space $\Delta d = \log d - \log \hat{d}$:
\begin{equation}
\mathcal{L}_D =
\frac{1}{P} \sum_{d \in D} \Delta d ^2 - \frac{\lambda}{P^2} \left(\sum_{d \in D} \Delta d \right)^2
\end{equation}
where $P$ is the number of pixels $d \in D$ with valid depth information. The coefficient $\lambda$ balances variance and error minimization, and following previous works~\cite{lee2019big} we use $\lambda=0.85$ in all experiments.

\subsubsection{Surface Normal Regularization}
\label{sec:sup-normal-loss}

Because depth estimates are produced on a per-pixel basis, it is common to enforce an additional smoothness loss~\cite{godard2017unsupervised} to maintain local consistency. Here we propose an alternative to the smoothness loss, that leverages the dense depth supervision available in synthetic datasets and minimizes instead the difference between surface normal vectors produced by ground-truth and predicted depth maps. Note that, differently from other methods~\cite{eigen_normals,cross_stitch}, we are not explicitly predicting surface normals as an additional task, but rather using them as regularization to enforce certain structural properties in the predicted depth maps. For any pixel $\mathbf{p} \in D$, its surface normal vector $\mathbf{n} \in \mathbb{R}^3$ is calculated as:
\begin{equation}
\mathbf{n} = \Big( \mathbf{P}_{u+1,v} - \mathbf{P}_{u,v} \Big) \times \Big( \mathbf{P}_{u,v+1} - \mathbf{P}_{u,v} \Big)
\end{equation}
where $\mathbf{P} = \phi(\mathbf{p}, d, K)$ is the point obtained by \emph{unprojecting} $\mathbf{p}$ from the camera frame of reference into 3D space, given its depth value $d$ and intrinsics $K$. As a measure of proximity between two surface normal vectors we use the \emph{cosine similarity} metric, defined as:
\begin{equation}
\mathcal{L}_{N} = \frac{1}{2P} \sum_{\mathbf{p} \in D} \Big( 1 - \frac{\hat{\mathbf{n}} \cdot \mathbf{n}}{||\hat{\mathbf{n}}|| \hspace{0.2em} ||\mathbf{n}||} \Big)
\end{equation}
where $\mathbf{n}$ and $\hat{\mathbf{n}}$ are unitary vectors representing respectively ground-truth and predicted surface normals for each pixel $\mathbf{p} \in D$. In  \figref{fig:normals} we show the impact that our proposed surface normal regularization has in the resulting predicted depth maps. While pure self-supervision on the target domain produces accurate depth maps, the surface normals are locally inconsistent and fail to properly capture object geometry. Introducing synthetic depth supervision significantly improves consistency, and our proposed regularization further sharpens object boundaries and enables the proper modeling of far away areas, including the sky. 

\subsubsection{Partially-Supervised Photometric Loss}
\label{sec:partial-sup-loss}

If image sequences are also provided as part of the synthetic sample, we can use the same self-supervised photometric loss described in \secref{sec:self-sup-loss} as additional training signal for the depth and pose networks. Furthermore, because we have individual dense depth and pose supervision, we can decouple each task within the self-supervised photometric loss. This is achieved by modifying Eq. \ref{eq:projection} to define $\hat{I}_t^D = I_{t'} \Big\langle \pi \Big( D_t, \hat{T}^{t'}_t, K \Big) \Big\rangle$ as the reconstructed target image obtained using ground-truth depth and predicted pose. Similarly, we can define $\hat{I}_t^T = I_{t'} \Big\langle \pi \Big( \hat{D}_t, T^{t'}_t, K \Big) \Big\rangle$ as the reconstructed target image obtained using predicted depth and ground-truth pose. The final partially-supervised photometric loss is calculated as:
\begin{equation}
\mathcal{L}_{PP} = \frac{1}{3} 
\Big( 
\mathcal{L}_P(I_t, \hat{I}_t) + 
\mathcal{L}_P(I_t, \hat{I}_t^D) + 
\mathcal{L}_P(I_t, \hat{I}_t^T)
\Big)
\end{equation}

%% file: figures/diagram2.tex
\begin{figure*}[t!]
\vspace{-6mm}
\centering
\includegraphics[width=0.95\textwidth,trim=0 0 0 0, clip]{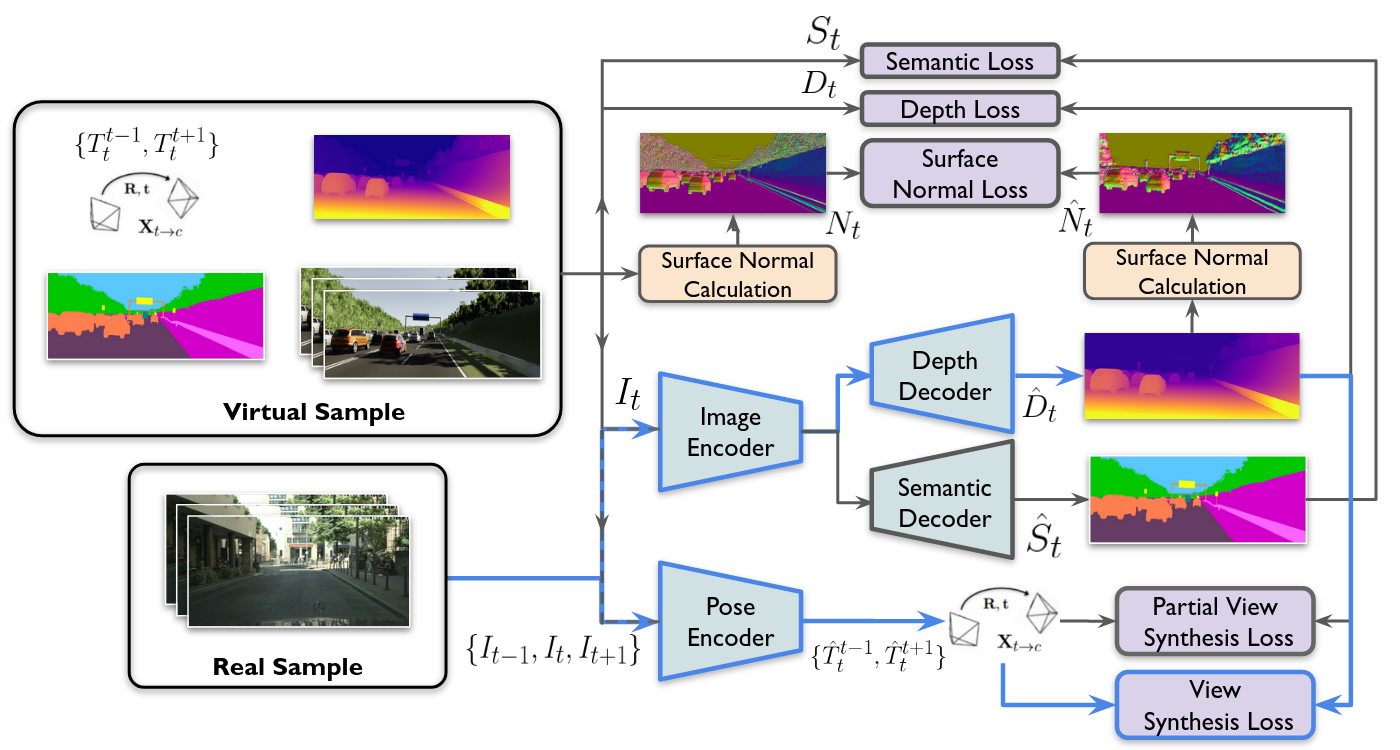}
\vspace{-2mm}
\caption{\textbf{Diagram of our proposed multi-task multi-domain GUDA architecture} for geometric unsupervised domain adaptation using mixed-batch training of real (Sec.~\ref{sec:real_sample}) and virtual (Sec.~\ref{sec:virtual_sample}) samples. The common paths during training (self-supervised) are in blue, whereas other ones (gray) use synthetic supervision.}
\label{fig:diagram}
\vspace{-4mm}
\end{figure*}

%% file: figures/normals.tex
\begin{figure}[t!]
\captionsetup[subfigure]{justification=centering}
\vspace{-3mm}
\small
\centering
\hspace{0.1mm}
\subfloat{\includegraphics[width=0.15\textwidth]{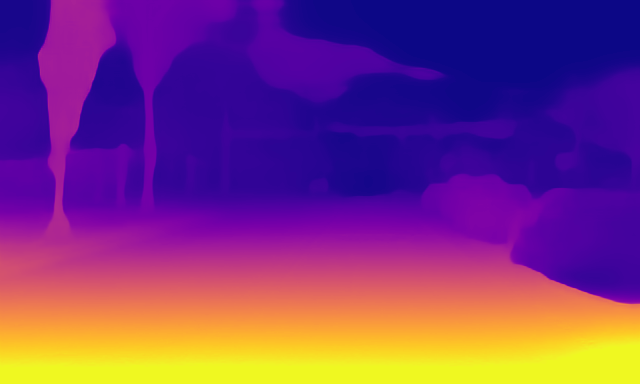}}
\hspace{0.9mm}
\subfloat{\includegraphics[width=0.15\textwidth]{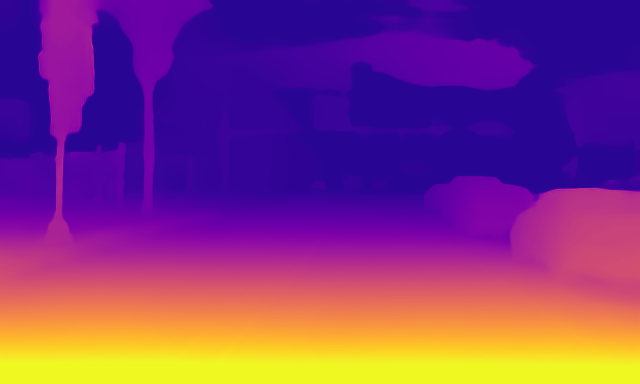}} 
\hspace{0.9mm}
\subfloat{\includegraphics[width=0.15\textwidth]{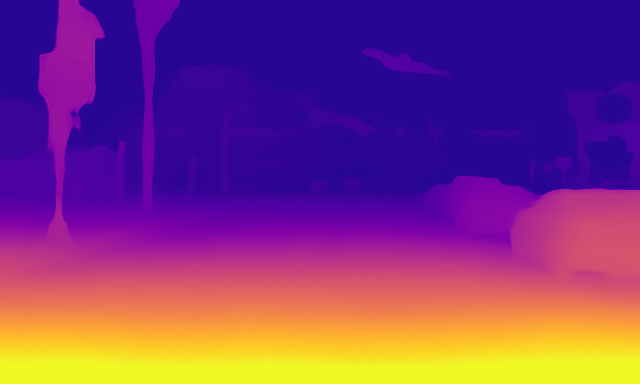}} 
\\ \vspace{-3mm}
\setcounter{subfigure}{0}
\subfloat[Real-world \\ self-supervision]{
\includegraphics[width=0.15\textwidth]{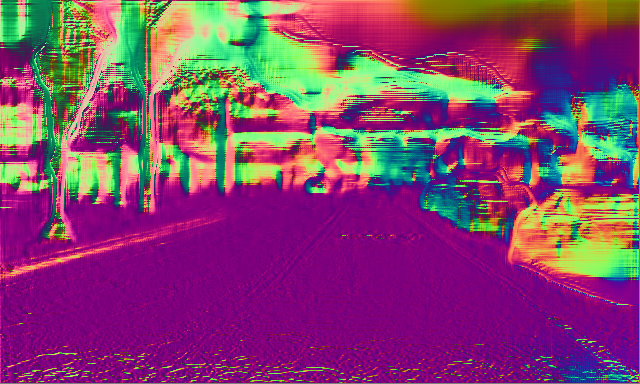}} 
\hspace{0.1mm}
\subfloat[Synthetic depth \\ supervision]{
\includegraphics[width=0.15\textwidth]{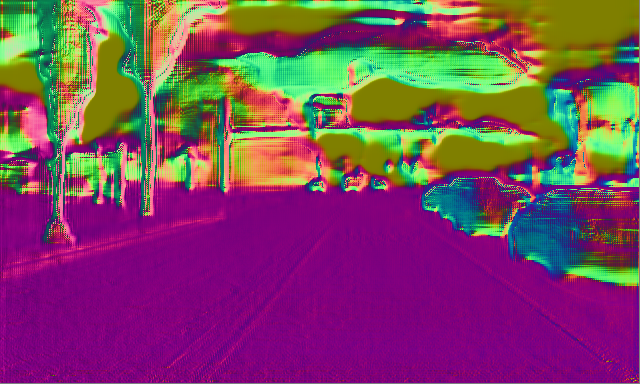}}
\hspace{0.1mm}
\subfloat[Synthetic depth + normal supervision]{
\includegraphics[width=0.15\textwidth]{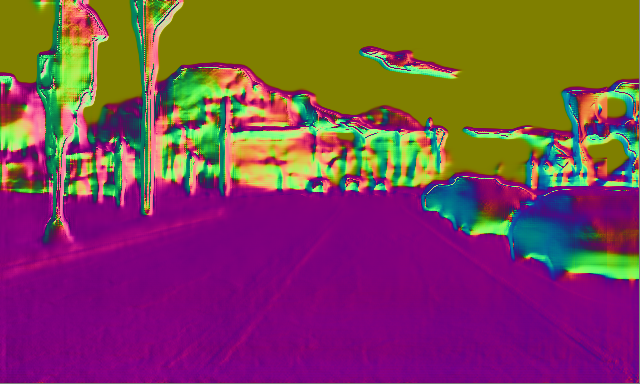}} 
\\
\caption{\textbf{Effects of surface normal regularization} from synthetic datasets on a real-world sample. In (a) only self-supervision on the real images was used during training, in (b) a synthetic depth loss was added, and in (c) our proposed surface normal regularization loss was added.}
\label{fig:normals}
\vspace{-4mm}
\end{figure}

%% file: sections/04protocol.tex
\subsection{Implementation Details}

Our models
%\footnote{Training and inference code will be made available upon publication.}
are implemented using Pytorch~\cite{paszke2017automatic} and trained across eight V100 GPUs with a batch size \VG{$b=8$ (1 per GPU)} for both source and target datasets. \VG{The shared depth and semantic encoder takes as input one RGB image (the middle frame), and the pose network takes three consecutive RGB images.} We use the Adam optimizer~\cite{adamw}, with $\beta_1=0.9$, $\beta_2=0.999$, and starting learning rate $lr=10^{-4}$. The length of each epoch is determined by the target dataset, and at each iteration samples are randomly selected from both source and target datasets.  Our training schedule consists of $20$ epochs at half resolution, followed by $20$ epochs at full resolution, halving the learning rate for the final $10$ epochs. \VG{This multi-scale schedule is used to increase convergence speed and has no meaningful impact in final model performance, as shown in Tab. \ref{tab:ablation}}. If pseudo-labels are available, the model is refined for another $10$ epochs with the introduction of $\mathcal{L}_{PL}$, halving the learning rate for the final $5$ epochs. Through grid search we have determined the loss coefficients to be $\lambda_V=1$, $\lambda_S=0.001$, $\lambda_N=0.01$, $\lambda_{PP}=0.005$ and $\lambda_{PL}=0.01$. Results are stable, starting to vary with coefficient changes of around one order of magnitude.

\subsection{Networks}
\label{sec:archi}
Unless noted otherwise, we use a ResNet101 \cite{he2016deep} with ImageNet \cite{Deng09imagenet} pre-trained weights as the shared backbone for the semantic and depth decoders. The depth decoder follows \cite{monodepth2} and outputs inverse depth maps at four different resolutions. The semantic decoder is similar and outputs semantic logits at a single resolution, obtained by concatenating the four output scales (upsampled to the highest resolution) followed by a final convolution layer.
Our pose network also follows \cite{monodepth2} and uses a ResNet18 encoder (also pre-trained on ImageNet), followed by a series of convolutions that output a 6-dimensional vector containing translation and rotation in Euler angles. For more details we refer the reader to the supplementary material. 

\vspace*{2mm}
\subsection{Datasets}

\vspace*{1mm}
\subsubsection{Real Datasets}
\vspace*{1mm}

\paragraph{Cityscapes~\cite{cordts2016cityscapes}}
The Cityscapes dataset is a widely used benchmark for semantic segmentation evaluation. 
For self-supervision on the target domain, following~\cite{packnet}, we use the $2975$ training images (without the labels) with their corresponding $30$-frame sequences, for a total of $2975 \times 30 = 89250$ images. \VG{In Tab. \ref{tab:ablation} we ablate the impact of training with fewer frames, such as 2-frame sequences (the minimum required for monocular self-supervised learning)}. We evaluate our semantic segmentation performance on the official $500$ annotated validation set.

%training split and include the 
%of the left camera $17$ Hz image sequences ($88250$ frames in total).
%For self-supervision on the real domain we follow \cite{packnet} and consider $88250$ training frames, and evaluate semantic segmentation on the official $500$ annotated validation frames.

\paragraph{KITTI \cite{geiger2013vision}}
The KITTI dataset is considered the standard benchmark for depth evaluation. We use the \textit{Eigen} split filtered according to \cite{zhou2016learning}, resulting in 39810 training, 888 validation and 697 test images, with corresponding LiDAR-projected depth maps (used only for evaluation). To evaluate semantic segmentation we use the 200 annotated frames found in \cite{Alhaija2018IJCV}, mapped to the Cityscapes ontology \cite{cordts2016cityscapes}.

\paragraph{DDAD~\cite{packnet}}
The Dense Depth for Automated Driving dataset is a challenging depth evaluation benchmark, with denser ground-truth depth maps and longer ranges of up to 250m. We consider only the front camera, resulting in $150$ training sequences with $12560$ images and $50$ validation sequences with $3950$ images, from which $50$ are semantically labeled (the middle frame in each sequence).

\input{tables/table_small}

\subsubsection{Synthetic Datasets}

\paragraph{SYNTHIA \cite{ros2016synthia}}
The SYNTHIA dataset contains scenes generated from an autonomous driving simulator of urban scenes. For a fair comparison with other methods we used the SYNTHIA-RAND-CITYSCAPES subset, with $9400$ images and semantic labels compatible with Cityscapes. 

\paragraph{VKITTI2 \cite{cabon2020vkitti2}}
The VKITTI2 dataset was recently released as a more photo-realistic version of \cite{gaidon2016virtual}, containing reconstructions of five sequences found in the KITTI odometry benchmark \cite{Geiger2012CVPR} for a total of $2156$ samples.
%In our experiments we used all five \textit{clone} sequences of camera $0$ for training, including RGB images, camera intrinsics and pose, semantic labels and depth maps, for a total of $2156$ samples. 

\vspace*{-2mm}
\paragraph{Parallel Domain \cite{parallel_domain}}
The Parallel Domain dataset\footnote{\url{https://paralleldomain.com/public-datasets/}} is procedurally-generated and contains fully annotated photo-realistic renderings of urban driving scenes, including multiple cameras and LiDAR sensors. It contains $5000$ $10$-frame sequences, with $4200$/$800$ training and validation splits. More details and examples can be found in the supplementary material. 

\vspace*{-2mm}
\paragraph{GTA5} \cite{Richter_2016_ECCV}
The GTA5 dataset is a street view synthetic dataset rendered from the GTA5 game engine, containing $24966$ images and semantic labels defined on $19$ classes compatible with Cityscapes.

% \noindent\textbf{CARLA.}
% Talk about CARLA here (?)

% \input{tables/datasets}

% \input{tables/depth}
% \input{tables/semantic}

% \input{figures/baselines}
% \input{figures/kitti}

% \input{figures/camera}

%\input{tables/kitti_depth}
%\input{tables/kitti_semantic}

%\input{tables/ddad_depth}
%\input{tables/ddad_semantic}

%\input{tables/cityscapes_depth}
%\input{tables/cityscapes_semantic}

%% file: tables/table_small.tex
\begin{table}[t!]
\renewcommand{\arraystretch}{0.85}
\centering
{
\small
\setlength{\tabcolsep}{0.45em}
% \begin{tabular}{l||c|c|c|c|c|c|c|c|c|c|c|c|c|c|c|c||c|c|}
\begin{tabular}{l||ccccc||c|c}
\toprule
\multirow{4}{*}{\textbf{Method}}
& \multirow{4}{*}{\rotatebox{90}{Sty.Trans.}}
& \multirow{4}{*}{\rotatebox{90}{Advers.}}
& \multirow{4}{*}{\rotatebox{90}{Depth}}
& \multirow{4}{*}{\rotatebox{90}{Self-Sup.}} 
& \multirow{4}{*}{\rotatebox{90}{Ps.-Label}} 
% & \multirow{4}{*}{A}
% & \multirow{4}{*}{D}
% & \multirow{4}{*}{PL} 
% & \multirow{4}{*}{SS} 
% & \multirow{3}{*}{\rotatebox{90}{Wall*}}
% & \multirow{3}{*}{\rotatebox{90}{Fence*}}
% & \multirow{3}{*}{\rotatebox{90}{Pole*}}
% & \multirow{3}{*}{\rotatebox{90}{T.Light}}
% & \multirow{3}{*}{\rotatebox{90}{T.Sign}}
% & \multirow{3}{*}{\rotatebox{90}{Vegt.}}
% & \multirow{3}{*}{\rotatebox{90}{Sky}}
% & \multirow{3}{*}{\rotatebox{90}{Person}}
% & \multirow{3}{*}{\rotatebox{90}{Rider}}
% & \multirow{3}{*}{\rotatebox{90}{Car}}
% & \multirow{3}{*}{\rotatebox{90}{Bus}}
% & \multirow{3}{*}{\rotatebox{90}{Motor.}}
% & \multirow{3}{*}{\rotatebox{90}{Bike}}
& \multirow{4}{*}{\rotatebox{90}{mIoU}}
& \multirow{4}{*}{\rotatebox{90}{mIoU*}}
\\
% & & & & & & & & & & & & & & & & & & \\
& & & & & & & \\
& & & & & & &\\

% & & & & & & & & & & & & & & & & & & \\
& & & & & & & \\
% \midrule
%\parbox[t]{2mm}{\multirow{17}{*}{\rotatebox[origin=c]{90}{}}}

% \bottomrule
% \multicolumn{10}{l}{\textbf{asdfasdfasf}}
\toprule
Source (SY) & & & & && 
31.7 & 36.7 \\
Source (PD) & & & & & & 
38.1 & 44.0 \\
Target & & & & & & 
72.9 & 77.8 \\

% \toprule
% \toprule
% \multicolumn{3}{l}{\textbf{(a) Comparison with other depth-based UDA methods (SYNTHIA $\rightarrow$ Cityscapes)}} \\
\toprule
\multicolumn{8}{l}{\textbf{\textit{(a) SYNTHIA $\rightarrow$ Cityscapes}}} \\
\toprule
SPIGAN \cite{spigan} & \checkmark&\checkmark & \checkmark & & & 
36.8 & 42.4 \\
GIO-Ada~\cite{Chen_2019_CVPR} & \checkmark& \checkmark & \checkmark & & & 
37.3 & 43.0 \\
DADA \cite{vu2019dada} &&  \checkmark & \checkmark & & & 
\underline{42.6} & \underline{49.8} \\

\cmidrule{1-8}

\textbf{GUDA} & & & \checkmark & \checkmark & & 
\textbf{44.5} & \textbf{50.9} \\

% \toprule
% \multicolumn{3}{l}{\textbf{(b) Comparison with other UDA methods (SYNTHIA $\rightarrow$ Cityscapes)}} \\
\toprule
CLAN \cite{clan}  & & \checkmark & & &  & 
---& 47.8\\
Xu et al.\cite{xu2019self} &  & \checkmark & & \checkmark & & 
38.8& ---\\
CBST \cite{cbst}&  &  & & & \checkmark & 
42.5& 48.4\\

% CAG \cite{cag}&  & & &\checkmark & & 
% 42.6 & 49.4 \\ % 
CRST \cite{crst} & & & & & \checkmark & 
43.8 & 50.1 \\
ESL \cite{Saporta2020ESLES}&  &  & & & \checkmark & 
43.5& 50.7\\
FDA \cite{fda}& \checkmark & & & & & 
--- & 52.5 \\
CCMD \cite{ccmd} & & & & & \checkmark & 
45.2 & 52.6 \\
Yang el al. \cite{Yang2020LabelDrivenRF} &\checkmark &\checkmark& & &  & 
--- & 53.1 \\
USAMR \cite{usamr} & & & & \checkmark &\checkmark & 
46.5 & 53.8 \\
IAST \cite{iast} & & & & & \checkmark & 
\underline{49.8} & \underline{57.0} \\

\cmidrule{1-8}

\textbf{GUDA+PL} &  & &\checkmark & \checkmark & \checkmark & 
\textbf{51.0} & \textbf{57.9} \\

% \toprule
% \multicolumn{3}{l}{\textbf{(c) Comparison with the state of the art (Varying Sources $\rightarrow$ Cityscapes)}} \\
\toprule
\multicolumn{8}{l}{\textbf{(b) \textit{Varying Sources (G5, PD) $\rightarrow$ Cityscapes}}} \\
\toprule
% Unserpervised domain adaptation through self-supervision
UDAS (G5) \cite{sun2019unsupervised} &\checkmark & \checkmark & & \checkmark & &
44.3
 & 49.2\\
USAMR (G5) \cite{usamr} & & & & \checkmark & \checkmark& 
53.1 & 58.0 \\
IAST (G5) \cite{iast}&  & & & & \checkmark & 
\underline{55.6} & \underline{61.2} \\

\cmidrule{1-8}

\textbf{GUDA(PD)+PL(G5)} & & & \checkmark & \checkmark & \checkmark & 
\textbf{57.2} & \textbf{63.2} \\

% \toprule
% \toprule

\bottomrule

\end{tabular}
}
\caption{\textbf{Semantic segmentation results on \emph{Cityscapes}} using different unsupervised domain adaptation (UDA) methods and synthetic datasets. \emph{mIoU} considers all 16 classes, and \emph{mIoU*} only the 13 SYNTHIA classes. \emph{Source} shows results without any adaptation, and \emph{Target} shows results with semantic supervision on the target domain. Synthetic datasets include: \emph{SYNTHIA (SY)}, \emph{Parallel Domain (PD)}, and \emph{GTA5 (G5)}. 
%Note that \emph{GTA5} does not contain depth information, and thus cannot be used with GUDA. 
Detailed per class results are given in the supplementary material.} 
\label{tab:small_table}
\vspace{-5mm}
\end{table}

%% file: sections/05experiments.tex
\subsection{Semantic Segmentation}

First, we evaluate our proposed GUDA framework on the task of unsupervised domain adaptation for semantic segmentation using the Cityscapes dataset. We consider three different scenarios, with results shown in \tabref{tab:small_table}. In (a) we use the SYNTHIA dataset as source and compare against other methods that use depth in the source domain as additional supervision, either as regularization~\cite{spigan}, to reduce domain shift at different feature levels~\cite{Chen_2019_CVPR}, or by sharing a lower-level representation like us. From these results, we see that \textbf{GUDA outperforms all previous methods}, even though it does not leverage additional translation networks or adversarial training. These results confirm that training the depth network in both domains thanks to self-supervised geometric constraints on \emph{target videos} improves the generalization of the shared intermediate representation.

A more detailed analysis (available in the supplementary material) shows that \textbf{GUDA excels in classes with well-defined geometries}, such as \emph{road}, \emph{sidewalk}, and \emph{building}. Interestingly, it also performs well on \emph{sky}, most likely due to our proposed surface normal regularization (\secref{sec:sup-normal-loss}), which can properly model such areas (\figref{fig:normals}). We also note that GUDA's smallest improvements are on rarer dynamic classes (e.g., \emph{motorcycle}). This stems from the photometric loss being unable to model dynamic object motion due to a static world assumption~\cite{monodepth2,guizilini2020semantically}.  

To overcome this limitation, we introduce pseudo-label supervision (\secref{sec:pseudo-label-loss}) from USAMR~\cite{usamr}, obtained by evaluating the official pre-trained model (mIoU 46.5) on our 89250 training images.
In this configuration, \textbf{GUDA achieves state-of-the-art results}, outperforming UDA methods relying on style-transfer~\cite{fda,Yang2020LabelDrivenRF}, adversarial learning~\cite{clan}, self-training~\cite{cbst, iast, ccmd, Saporta2020ESLES}, or other forms of (non-geometric) self-supervision~\cite{xu2019self,sun2019unsupervised}.
\figref{fig:lambda} analyzes the interplay between geometric self-supervision (GS) and pseudo-labels (PL), indicating an optimal pseudo-labeling loss weight $\lambda_{PL}=0.01$.

A detailed ablation study of our proposed architecture can be found in \tabref{tab:ablation}. It shows that (i) geometric supervision by itself improves performance, (ii) all components help, and (iii) the benefits of our method are not due simply to using more target data (video frames), although GUDA can benefit from them in contrast to other approaches.

\input{figures/lambda}

\input{figures/vkitti2_kitti}

\input{figures/pd_ddad}

Finally, in \tabref{tab:small_table} (b) we present results considering different source datasets. Because GTA5~\cite{Richter_2016_ECCV} does not provide depth labels, we instead report results using the photorealistic \emph{Parallel Domain} dataset, with GTA5 pseudo-labels (mIoU 53.1) from USAMR~\cite{usamr}. In this configuration, \textbf{GUDA once more outperforms other considered methods}, improving upon the state of the art when different source datasets are considered.

% \VG{Good place to include a brief discussion vs other self-sup. methods.}

\input{tables/ablation}

\input{figures/scale}

To evaluate how GUDA generalizes to other source and target datasets, we also provide semantic segmentation results from \emph{VKITTI2} to \emph{KITTI} (\figref{fig:vkitti2_kitti}) and \emph{Parallel Domain} to \emph{DDAD} (\figref{fig:pd_ddad}). To our knowledge, no current method reports results on these benchmarks for this task. Hence, we compare to the standard \emph{Domain Adversarial Neural Network} (DANN) baseline~\cite{dann}, which uses a gradient reversal layer to learn discriminative features for the main task on the source domain while maximizing domain confusion.
% repetition? 
% Note that this is similar to the underlying structure of GUDA, however we instead reduce domain shift by simultaneously learning depth features --- and, by extension, shared semantic features --- in both domains. 

\input{figures/cityscapes}

\input{tables/depth_datasets}

As expected, DANN improves over source-only results by a significant margin ($+7.0/+5.1$ mIoU). Nevertheless, the stronger geometric supervision from GUDA still substantially outperforms DANN ($+12.65/+8.58$ mIoU), with similar trends as observed in previous experiments. In \figref{fig:scale} we show how GUDA and DANN scale with improvements in data quality (from \emph{SYNTHIA} to \emph{Parallel Domain}) and data quantity (different subsets of \emph{Parallel Domain}). Assuming linear improvement, GUDA would only require 200k synthetic samples to fully overcome the domain gap, whereas DANN would require 350k samples.
We also show in Tab.~\ref{tab:ablation} that GUDA and DANN can be combined, resulting in a $+0.2/0.3\%$ improvement.

\subsection{Depth Estimation}

As stated previously, GUDA achieves domain adaptation by jointly learning depth features in both domains, using a combination of dense synthetic supervision and geometric self-supervision on real-world images. To further demonstrate this property, in this section we analyze how GUDA impacts the task of monocular depth estimation itself and improves upon the standard approach of model fine-tuning. Similar to previous experiments, we evaluate on \emph{VKITTI2} to \emph{KITTI} and \emph{Parallel Domain} to \emph{DDAD}, noting that each of these combinations have similar sensor configuration (intrinsics and extrinsics), which makes them particularly suitable for domain adaptation experiments. The same training schedule and architecture was used, with the inclusion of experiments using a ResNet18 backbone to facilitate comparison with other methods.

\input{figures/depth}

Quantitative results are shown in  \tabref{tab:depth_datasets}, with qualitative examples in \figref{fig:kitti_depth}. The first noticeable aspect is that direct transfer (\emph{Source only}) not only produces relatively accurate, but also \emph{scale-aware} results, due to similarities in vehicle extrinsics and camera parameters. As expected, this behavior is not observed when only real-world information (\emph{Target only}) is used, however it is also not observed when fine-tuning, indicating a catastrophic forgetting of the scale factor. In contrast, and although not our primary goal, \textbf{GUDA preserves the scale learned from synthetic supervision} and also \textbf{significantly improves depth estimation performance} relative to the standard fine-tuning approach. In alignment with recent observations~\cite{packnet}, switching to a larger backbone improves results even further.
%, especially on a saturated dataset such as KITTI.

% \input{tables/semantic_synthia_cs}

%% file: figures/lambda.tex
\begin{figure}[t!]
\vspace{-2mm}
\centering
\includegraphics[width=0.4\textwidth]{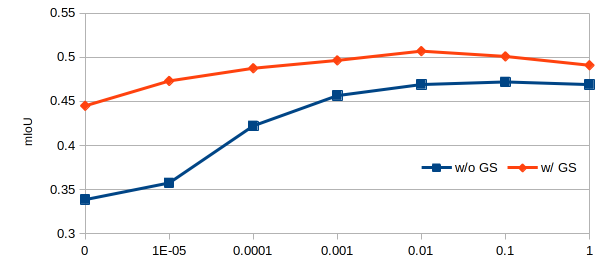}
\caption{\textbf{Effects of pseudo-label supervision}, with and without our proposed geometric self-supervision (GS) for different values of $\lambda_{PL}$. When GS is not used (blue line), there is no benefit in adding virtual supervision, and as $\lambda_{PL}$ increases results converge to those using only pseudo-label supervision (Tab. \ref{tab:ablation}). When GS is used (red line), results are consistently higher and start to degrade after $\lambda_{PL}=0.01$.}
\label{fig:lambda}
\end{figure}

%% file: figures/vkitti2_kitti.tex
\begin{figure}[t!]
\centering
\includegraphics[width=0.49\textwidth,height=5cm,trim=15 10 0 0,clip]{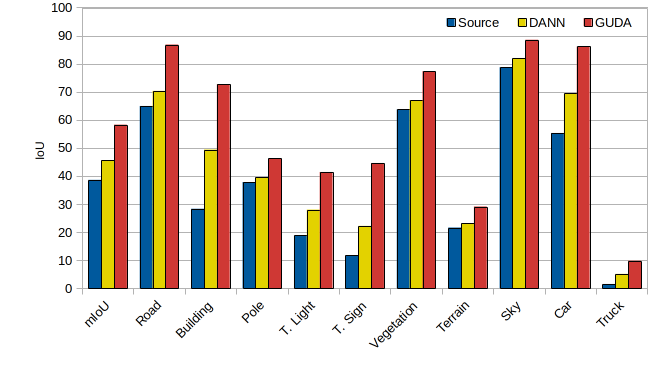}
\vspace{-6mm}
\caption{\textbf{Semantic segmentation results on \textit{VKITTI2} $\rightarrow$ \textit{KITTI}}, using GUDA and DANN \cite{dann}. Detailed numbers are available in the supplementary material.}
\label{fig:vkitti2_kitti}
\end{figure}

%% file: figures/pd_ddad.tex
\begin{figure}[t!]
\vspace{-1mm}
\centering
\includegraphics[width=0.49\textwidth,height=5cm,trim=20 10 0 0,clip]{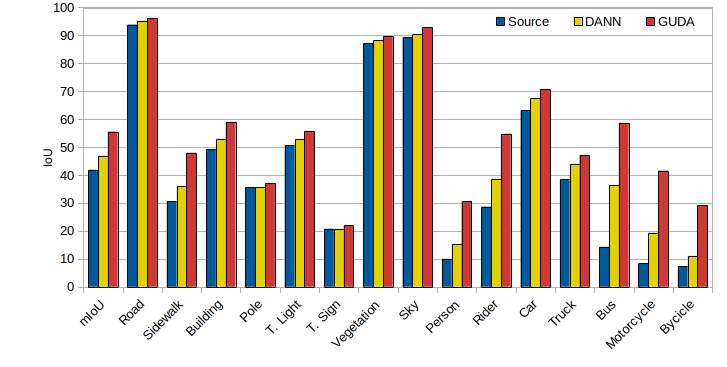}
\vspace{-6mm}
\caption{\textbf{Semantic segmentation results on \textit{Parallel Domain} $\rightarrow$ \textit{DDAD}}, using GUDA and DANN \cite{dann}. Detailed numbers are available in the supplementary material.}
\label{fig:pd_ddad}
\vspace{-4mm}
\end{figure}

%% file: tables/ablation.tex
\begin{table}[t!]
\vspace*{-1mm}
\small
\renewcommand{\arraystretch}{0.9}
\centering
{
\small
\setlength{\tabcolsep}{0.2em}
\begin{tabular}{l|ccc|ccc||cc}

\toprule
\multirow{2}{*}{\textbf{Variation}} & \multicolumn{3}{c}{Real} & \multicolumn{3}{|c||}{Virtual} & \multicolumn{2}{c}{} \\
% \cmidrule{2-7}
& GT & GS & PL & D & S & N & mIoU & mIoU* 
\\
\toprule
% \midrule
% \textbf{Ours} & \xmark & \xmark & \xmark & \xmark & \xmark & \xmark & --- & --- \\
% \midrule
\VG{{DADA}} & & & & & & & 42.6 & 49.8 \\
% DADA 2 frames all classes: 0.89, 0.43, 0.81, 0.1 , 0.01, 0.29, 0.1 , 0.14, 0.81, 0.83, 0.54, 0.18, 0.8 , 0.35, 0.17, 0.36
\VG{{DADA (2 frames)}} & & & & & & & 42.5 & 49.3 \\
% DADA all frames all classes: 0.91, 0.46, 0.81, 0.06, 0.  , 0.28, 0.07, 0.09, 0.81, 0.85, 0.57, 0.21, 0.83, 0.34, 0.19, 0.29
\VG{{DADA (30 frames)}} & & & & & & & 42.4 & 49.5 \\
\midrule
Source only & 
\xmark & \xmark & \xmark & \xmark & \cmark & \xmark & 32.1 & 37.2 \\
Target only (GT) & 
\cmark & \xmark & \xmark & \xmark & \xmark & \xmark & 73.9 & 78.7 \\
Target only (PL) & 
\xmark & \xmark & \cmark & \xmark & \xmark & \xmark & 47.1 & 54.3 \\
Virtual only & 
\xmark & \xmark & \xmark & \cmark & \cmark & \cmark & 33.8 & 38.8 \\
\midrule
\textbf{GUDA - N} & 
\xmark & \cmark & \xmark & \cmark & \cmark & \xmark & 43.4 & 48.0 \\
\VG{\textbf{GUDA (2 frames)}} & 
\xmark & \cmark & \xmark & \cmark & \cmark & \cmark & 43.8 & 50.5 \\
\VG{\textbf{GUDA (single res.)}} & 
\xmark & \cmark & \xmark & \cmark & \cmark & \cmark & 44.3 & 50.8 \\
\textbf{GUDA} & 
\xmark & \cmark & \xmark & \cmark & \cmark & \cmark & 44.5 & 50.9 \\
\VG{\textbf{GUDA + DANN}} & 
\xmark & \cmark & \xmark & \cmark & \cmark & \cmark & 44.8 & 51.1 \\
\textbf{GUDA + PL - GS} & 
\xmark & \xmark & \cmark & \cmark & \cmark & \cmark & 46.8 & 54.1 \\
\textbf{GUDA + PL} & 
\xmark & \cmark & \cmark & \cmark & \cmark & \cmark & \textbf{51.0} & \textbf{57.9} \\
% \midrule
% \midrule
% \parbox[t]{2mm}{\multirow{7}{*}{\rotatebox[origin=c]{90}{\textit{Parallel Domain}}}} & Source only (10k) & 
% \xmark & \xmark & \xmark & \xmark & \cmark & \xmark & 32.3 & 37.3 \\
% & Source only (25k) & 
% \xmark & \xmark & \xmark & \xmark & \cmark & \xmark & 35.8 & 40.4 \\
% & Source only (42k) & 
% \xmark & \xmark & \xmark & \xmark & \cmark & \xmark & 37.5 & 43.4 \\
% \cmidrule{2-10}
% & \textbf{GUDA (10k)} & 
% \xmark & \xmark & \cmark & \cmark & \cmark & \cmark & 46.3 & 50.1 \\
% & \textbf{GUDA (25k)} & 
% \xmark & \xmark & \cmark & \cmark & \cmark & \cmark & 47.6 & 53.2 \\
% & \textbf{GUDA (42k)} & 
% \xmark & \xmark & \cmark & \cmark & \cmark & \cmark & 50.5 & 57.9 \\
% & \textbf{GUDA (42k) + PL (G5)} & 
% \xmark & \cmark & \cmark & \cmark & \cmark & \cmark & \textbf{57.2} & \textbf{63.7} \\

\bottomrule

\end{tabular}
}
\vspace*{-1mm}
\caption{\textbf{Ablation study of our proposed method (\emph{SYNTHIA} $\rightarrow$ \emph{Cityscapes})}. In the \textit{Real} columns, \textit{GT} refers to the use of ground-truth semantic labels, \textit{PL} to the use of USAMR \cite{usamr} semantic pseudo-labels (\secref{sec:pseudo-label-loss}), and \textit{GS} to the use of geometric self-supervision (\secref{sec:self-sup-loss}). In the \textit{Virtual} columns, \textit{D}, \textit{S}, and \textit{N} refer respectively to semantic, depth, and surface normal supervision (\secref{sec:virtual_sample}). \VG{\textit{DANN} refers to the use of an additional domain adversarial loss \cite{dann}, and \textit{single res.} to training only in full resolution. We also provide DADA results trained using 2 and 30 context frames (rather than 1), as well as GUDA results using only the minimum 2 context frames (rather than 30).}}
\label{tab:ablation}
\vspace{-3mm}
\end{table}

%% file: figures/scale.tex
\begin{figure}[t!]
\centering
\includegraphics[width=0.45\textwidth,trim=0 0 0 0,clip]{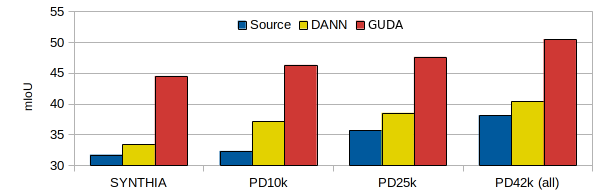}
\caption{\textbf{Performance improvement on \emph{Cityscapes}} with increasing data quality (SYNTHIA $\rightarrow$ PD) and quantity.}
\label{fig:scale}
\vspace{-4mm}
\end{figure}

%% file: figures/cityscapes.tex
\begin{figure*}[t!]
\centering
\vspace{-7mm}
% \raisebox{9mm}{\rotatebox[origin=c]{90}{\textit{Input}}}
\subfloat{
\includegraphics[width=2.7cm,height=1.6cm]{
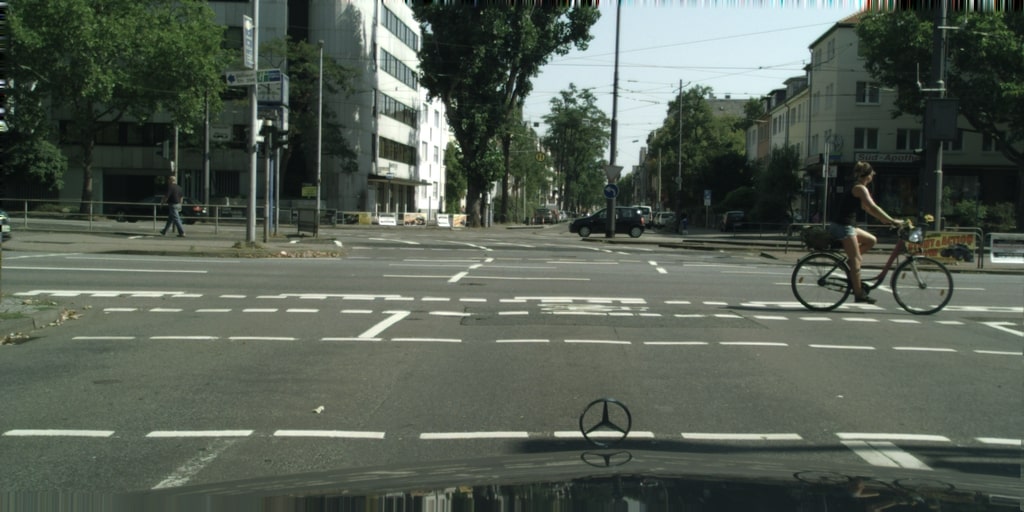}}
\subfloat{
\includegraphics[width=2.7cm,height=1.6cm]{
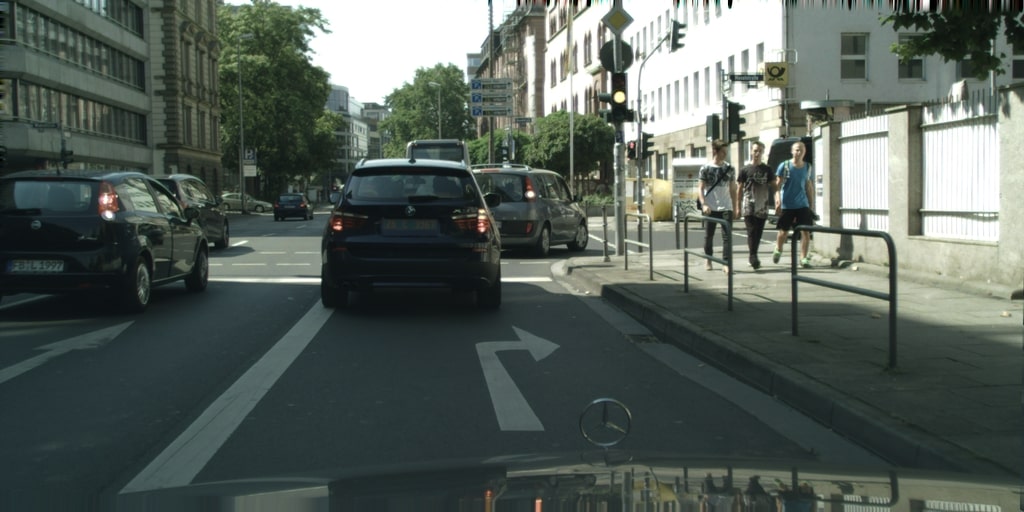}}
\subfloat{
\includegraphics[width=2.7cm,height=1.6cm]{
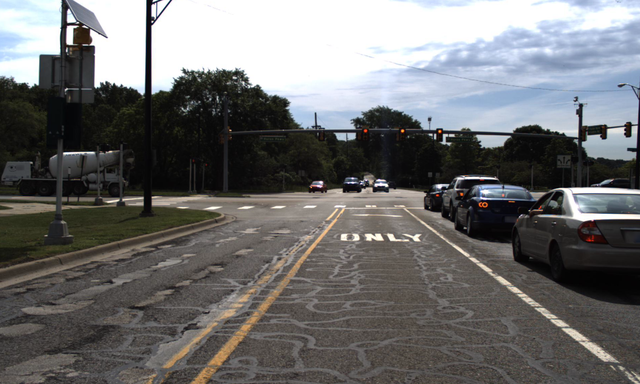}}
\subfloat{
\includegraphics[width=2.7cm,height=1.6cm]{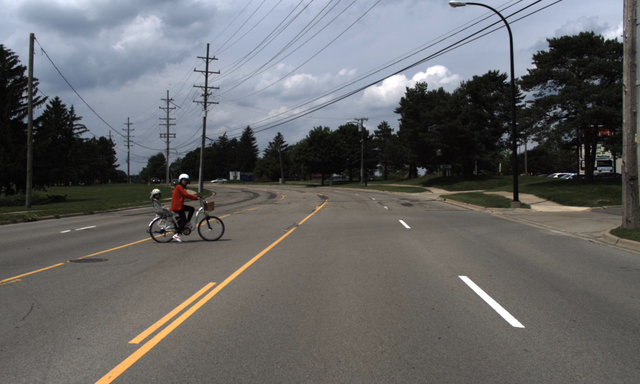}}
\subfloat{
\includegraphics[width=2.7cm,height=1.6cm]{
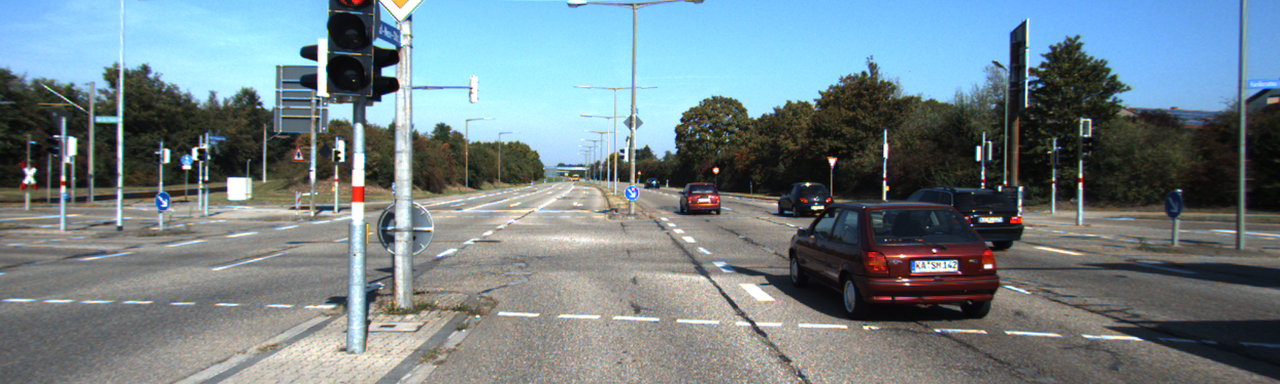}}
\subfloat{
\includegraphics[width=2.7cm,height=1.6cm]{
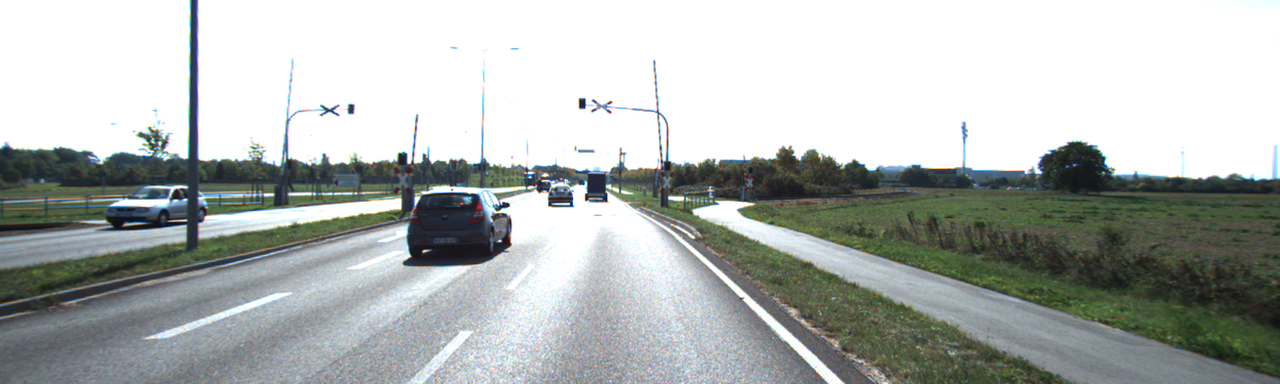}}
\\ \vspace{-3mm}
% \raisebox{9mm}{\rotatebox[origin=c]{90}{\textit{GUDA}}}
\subfloat{
\includegraphics[width=2.7cm,height=1.6cm]{
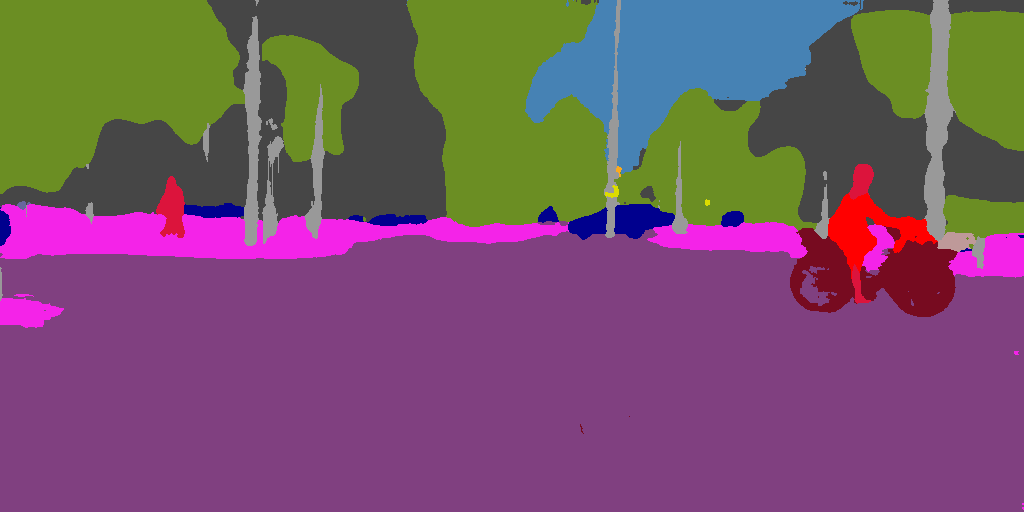}}
\subfloat{
\includegraphics[width=2.7cm,height=1.6cm]{
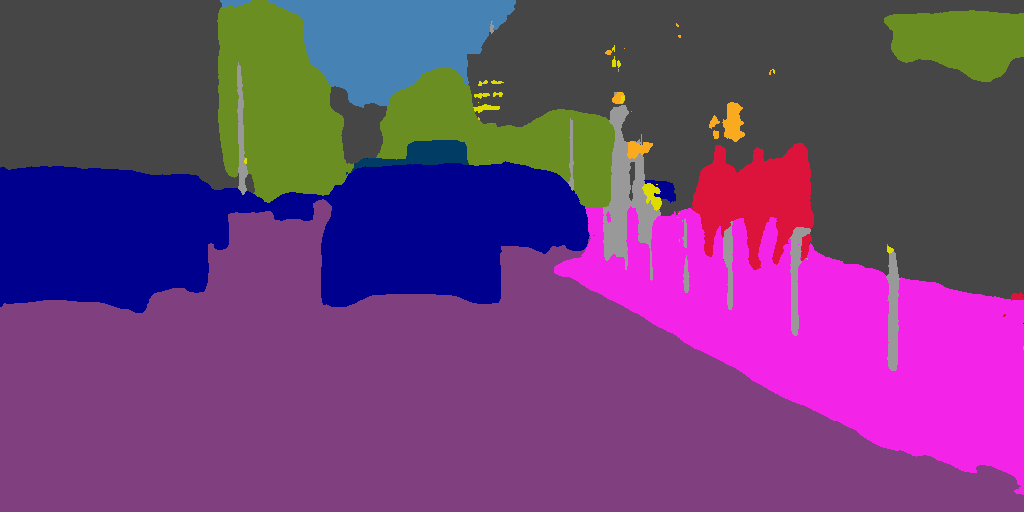}}
\subfloat{
\includegraphics[width=2.7cm,height=1.6cm]{
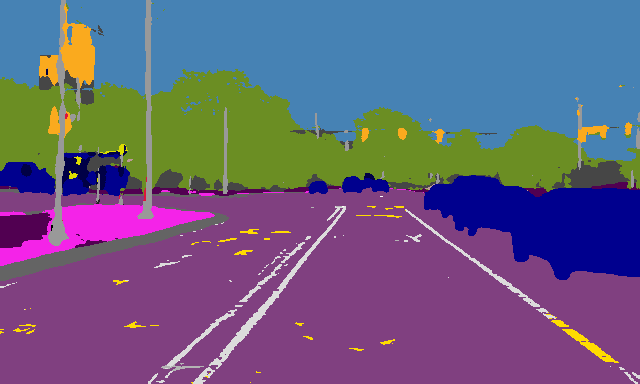}}
\subfloat{
\includegraphics[width=2.7cm,height=1.6cm]{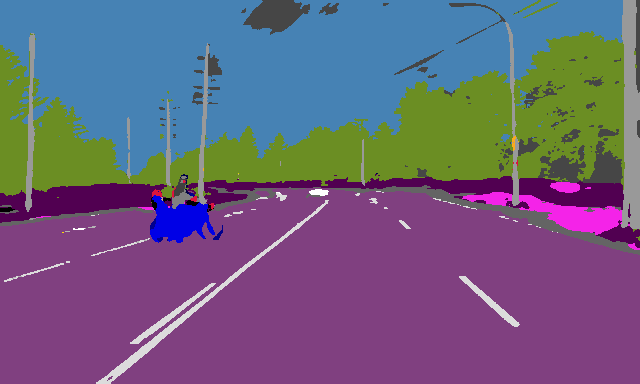}}
\subfloat{
\includegraphics[width=2.7cm,height=1.6cm]{
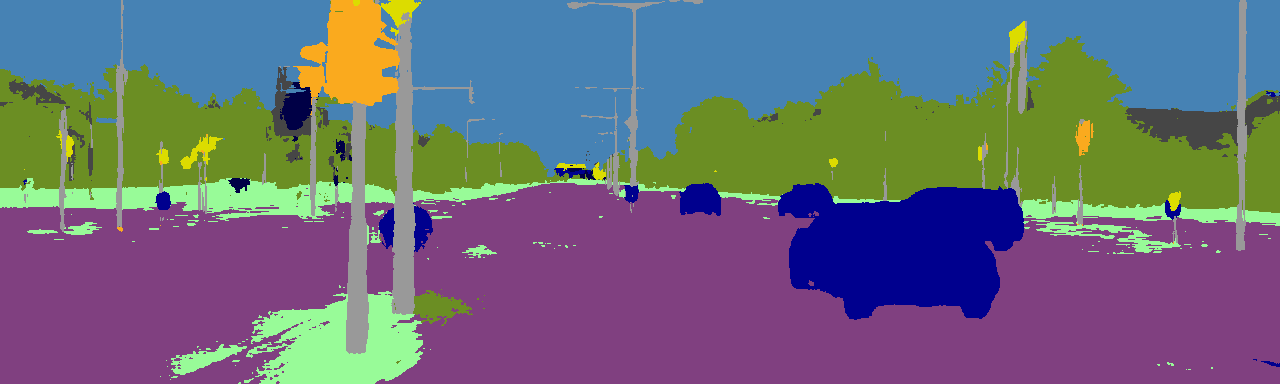}}
\subfloat{
\includegraphics[width=2.7cm,height=1.6cm]{
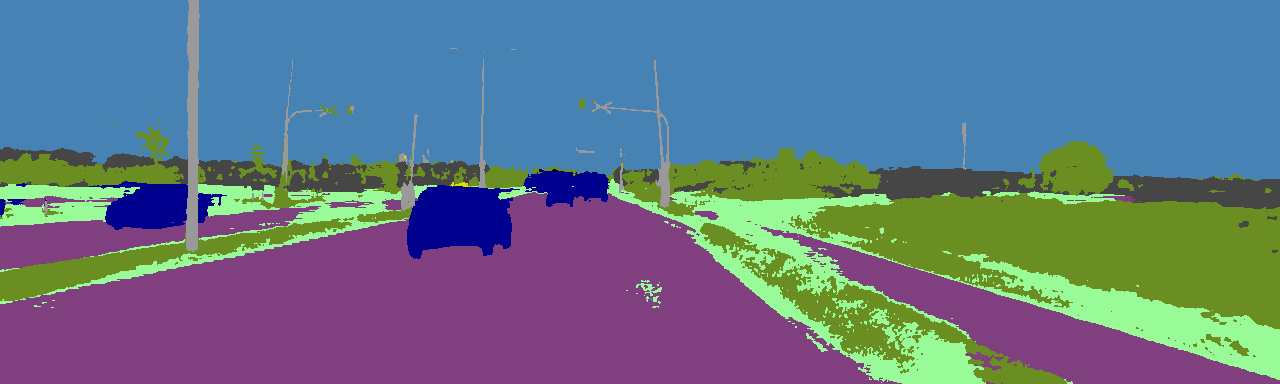}}
% \\ \vspace{-3mm}
% \raisebox{9mm}{\rotatebox[origin=c]{90}{\textit{GUDA+PL}}}
% \subfloat{
% \includegraphics[width=2.7cm,height=2.0cm]{images/cityscapes/16-logits_semantic_PL.png}}
% \subfloat{
% \includegraphics[width=2.7cm,height=2.0cm]{images/cityscapes/128-pred_semantic_PL.png}}
% \subfloat{
% \includegraphics[width=2.7cm,height=2.0cm]{images/depth/ddad/val-ddad-val-lidar-camera_01-784-inv_depth_20_0a331a4e.png}}
% \subfloat{
% \includegraphics[width=2.7cm,height=2.0cm]{images/cityscapes/264-pred_semantic_PL.png}}
% \subfloat{
% \includegraphics[width=2.7cm,height=2.0cm]{images/cityscapes/280-pred_semantic_PL.png}}
% \subfloat{
% \includegraphics[width=2.7cm,height=2.0cm]{images/cityscapes/336-pred_semantic_PL.png}}
\vspace{-2mm}
\caption{\textbf{GUDA semantic segmentation results} on \emph{Cityscapes}, \emph{DDAD} and \emph{KITTI}.}
\label{fig:baselines}
\end{figure*}

%% file: tables/depth_datasets.tex
\begin{table}[t!]
\vspace{-4mm}
\renewcommand{\arraystretch}{0.9}
\centering
{
\small
\setlength{\tabcolsep}{0.25em}
\begin{tabular}{c|l||cccc}
\toprule
& \textbf{Method} &
Abs.Rel$\downarrow$ &
Sqr.Rel$\downarrow$ &
RMSE$\downarrow$ &
$\delta<1.25$$\uparrow$ \\
\toprule

\parbox[t]{2mm}{\multirow{5}{*}{\rotatebox[origin=c]{90}{\textit{KITTI}}}}

& Source only$\dagger$ (R18) &
0.191 & 2.078 & 7.233 & 0.699 
\\
& Target only (R18) &
0.117 & 0.811 & 4.902 & 0.867
\\
& Fine-tune (R18) &
0.114 & 0.800 & 4.855 & 0.871
\\

\cmidrule{2-6}

& \VG{\textbf{GUDA$\dagger$ (R18) - PS}} &
0.114 & 0.875 & 4.808 & 0.871
\\
& \textbf{GUDA$\dagger$ (R18)} &
0.109 & 0.762 & 4.606 & 0.879
\\
& \textbf{GUDA$\dagger$} &
\textbf{0.107} & \textbf{0.714} & \textbf{4.421} & \textbf{0.883}
\\

\midrule
\midrule

\parbox[t]{2mm}{\multirow{5}{*}{\rotatebox[origin=c]{90}{\textit{DDAD}}}}

& Source only$\dagger$ (R18) &
0.233 & 7.429 & 18.498 & 0.620
\\
& Target only (R18) & 
0.203 & 6.999 & 16.844 & 0.748
\\
& Fine-tune (R18) &
0.173 & 4.846 & 16.025 & 0.747
\\

\cmidrule{2-6}

& \VG{\textbf{GUDA$\dagger$ (R18) - PS}} &
0.166 & 3.556 & 16.004 & 0.769
\\
& \textbf{GUDA$\dagger$ (R18)} &
0.158 & 3.332 & 15.112 & 0.778
\\
& \textbf{GUDA$\dagger$} &
\textbf{0.147} & \textbf{2.922} & \textbf{14.452} & \textbf{0.809}
\\
\bottomrule

\end{tabular}
}
\vspace{-2mm}
\caption{\textbf{Depth estimation results on \emph{KITTI} and \emph{DDAD}}. \textit{R18} indicates a ResNet18~\cite{he2016deep} backbone, \emph{Source only} and \emph{Target only} indicate only synthetic or real-world, and \emph{Fine-tune} indicates synthetic pre-training followed by real-world fine-tuning. \VG{\emph{PS} indicates the removal of the partially-supervised photometric loss} (Sec. \ref{sec:partial-sup-loss}). The symbol $\dagger$ indicates a scale-aware model \VG{(no test-time median-scaling)}.}
\label{tab:depth_datasets}
\vspace{-3mm}
\end{table}

%% file: figures/depth.tex
\begin{figure}[t!]
\vspace{-8mm}
\centering

% \subfloat{
% \includegraphics[width=0.15\textwidth,height=1.1cm]{images/depth/ddad/val-ddad-sem-lidar-camera_01-24-rgb_20_5a15a091.png}}
% \subfloat{
% \includegraphics[width=0.15\textwidth,height=1.1cm]{images/depth/ddad/val-ddad-sem-lidar-camera_01-24-inv_depth_20_3304e168.png}}
% \subfloat{
% \includegraphics[width=0.15\textwidth,height=1.1cm]{images/depth/ddad/val-ddad-sem-lidar-camera_01-24-calc_normals_20_82e5ce91.png}}

% \vspace{-3mm}

% \subfloat{
% \includegraphics[width=0.15\textwidth,height=1.1cm]{images/depth/ddad/val-ddad-val-lidar-camera_01-1568-rgb_20_2bb082ba.png}}
% \subfloat{
% \includegraphics[width=0.15\textwidth,height=1.1cm]{images/depth/ddad/val-ddad-val-lidar-camera_01-1568-inv_depth_5_1e3f6ca2.png}}
% \subfloat{
% \includegraphics[width=0.15\textwidth,height=1.1cm]{images/depth/ddad/val-ddad-val-lidar-camera_01-1568-calc_normals_20_7d059010.png}}

% \vspace{-3mm}

\subfloat{
\includegraphics[width=0.15\textwidth,height=1.1cm]{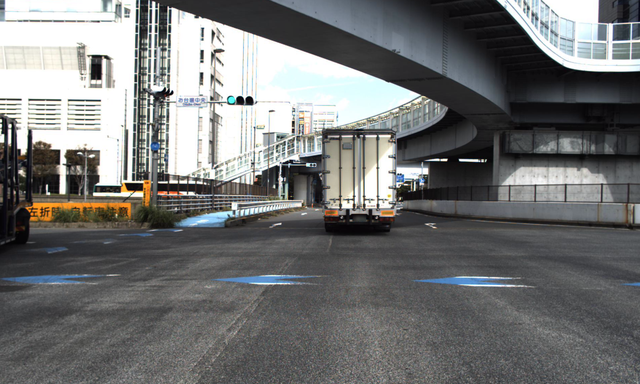}}
\subfloat{
\includegraphics[width=0.15\textwidth,height=1.1cm]{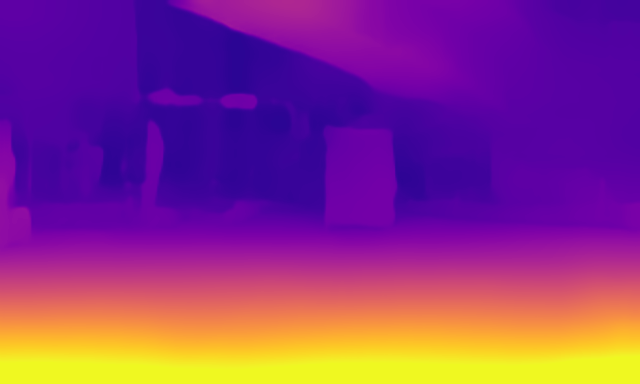}}
\subfloat{
\includegraphics[width=0.15\textwidth,height=1.1cm]{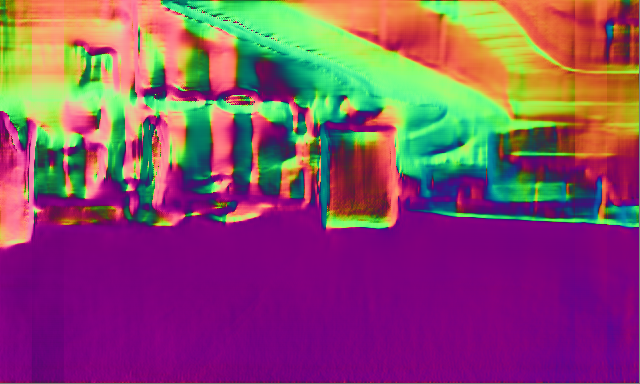}}

% \vspace{-3mm}

% \subfloat{
% \includegraphics[width=0.15\textwidth,height=1.1cm]{images/depth/ddad/val-ddad-val-lidar-camera_01-784-rgb_20_efe69ff8.png}}
% \subfloat{
% \includegraphics[width=0.15\textwidth,height=1.1cm]{images/depth/ddad/val-ddad-val-lidar-camera_01-784-inv_depth_20_0a331a4e.png}}
% \subfloat{
% \includegraphics[width=0.15\textwidth,height=1.1cm]{images/depth/ddad/val-ddad-val-lidar-camera_01-784-calc_normals_20_398049f0.png}}

\vspace{-3mm}

% \subfloat{
% \includegraphics[width=0.15\textwidth,height=1.1cm]{images/kitti_depth/media_images_val-KITTI_raw-eigen_test_files-velodyne-272-rgb_0_e964ab8c.png}}
% \subfloat{
% \includegraphics[width=0.15\textwidth,height=1.1cm]{images/kitti_depth/media_images_val-KITTI_raw-eigen_test_files-velodyne-272-inv_depths_10_23ce63ac.png}}
% \subfloat{
% \includegraphics[width=0.15\textwidth,height=1.1cm]{images/kitti_depth/media_images_val-KITTI_raw-eigen_test_files-velodyne-272-calc_normals_10_85dff18d.png}}

% \vspace{-3mm}

\subfloat{
\includegraphics[width=0.15\textwidth,height=1.1cm]{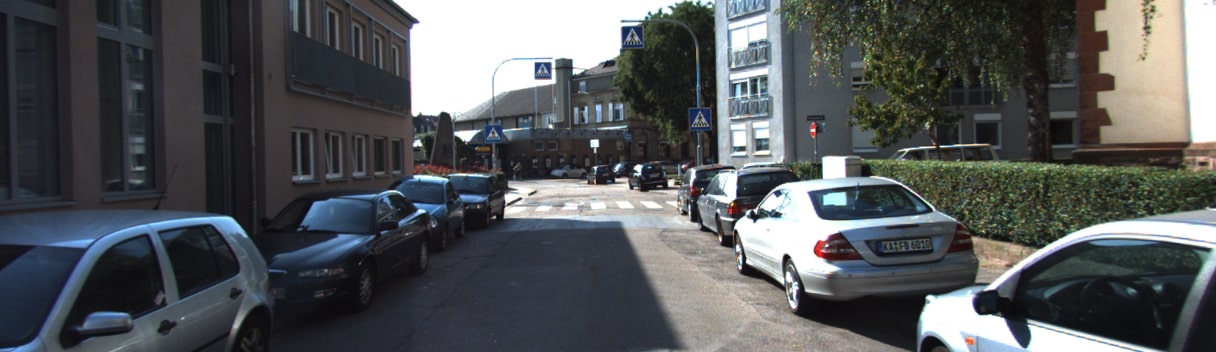}}
\subfloat{
\includegraphics[width=0.15\textwidth,height=1.1cm]{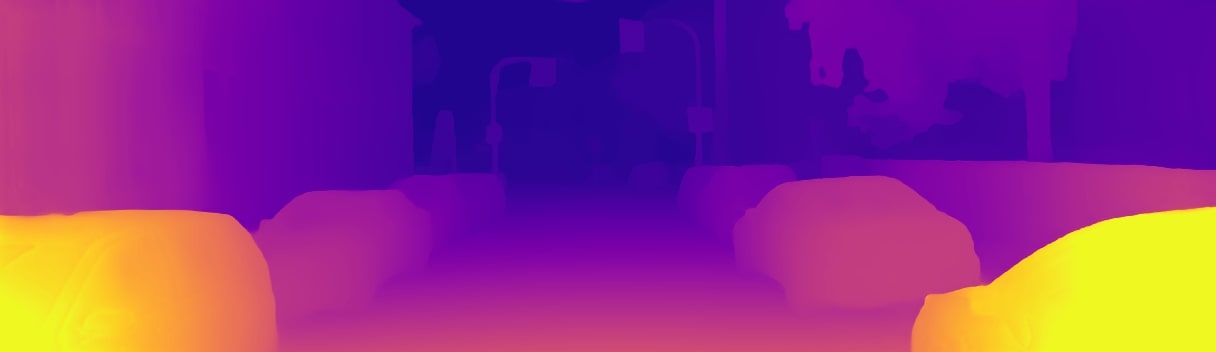}}
\subfloat{
\includegraphics[width=0.15\textwidth,height=1.1cm]{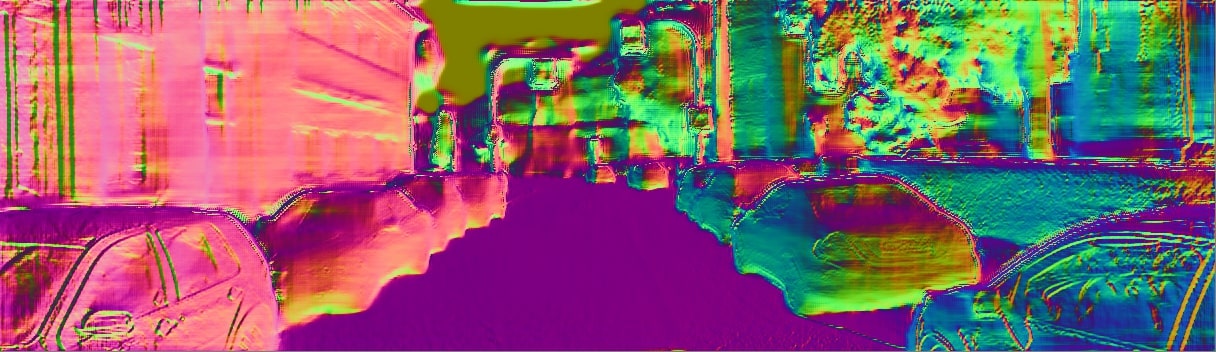}}

\vspace{-3mm}

\subfloat{
\includegraphics[width=0.15\textwidth,height=1.1cm]{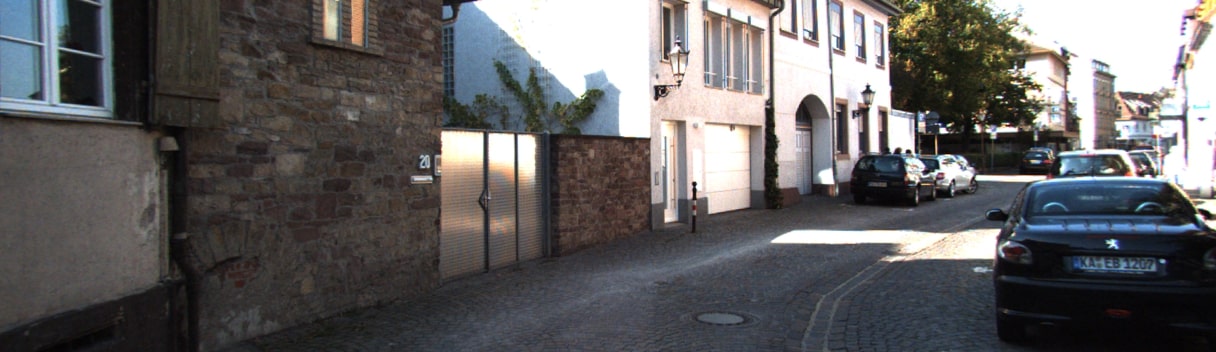}}
\subfloat{
\includegraphics[width=0.15\textwidth,height=1.1cm]{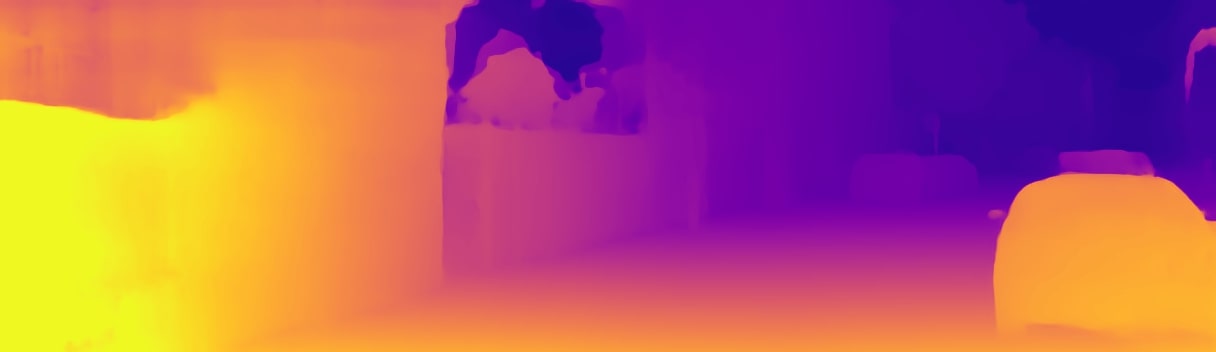}}
\subfloat{
\includegraphics[width=0.15\textwidth,height=1.1cm]{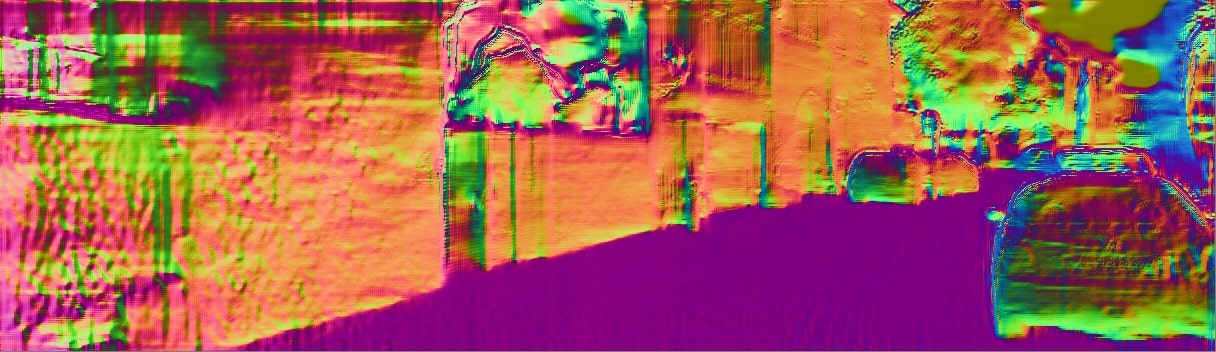}}

\vspace{-3mm}

\setcounter{subfigure}{0}

\subfloat[Input image]{
\includegraphics[width=0.15\textwidth,height=1.1cm]{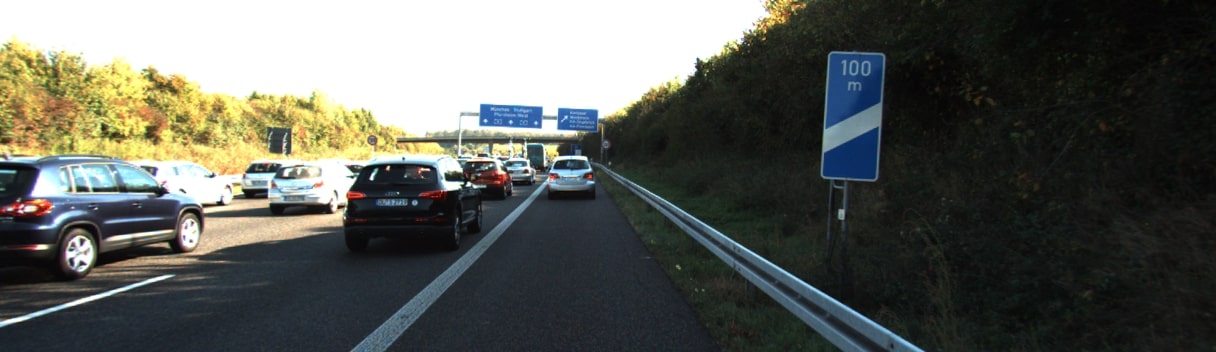}}
\subfloat[Pred. depth]{
\includegraphics[width=0.15\textwidth,height=1.1cm]{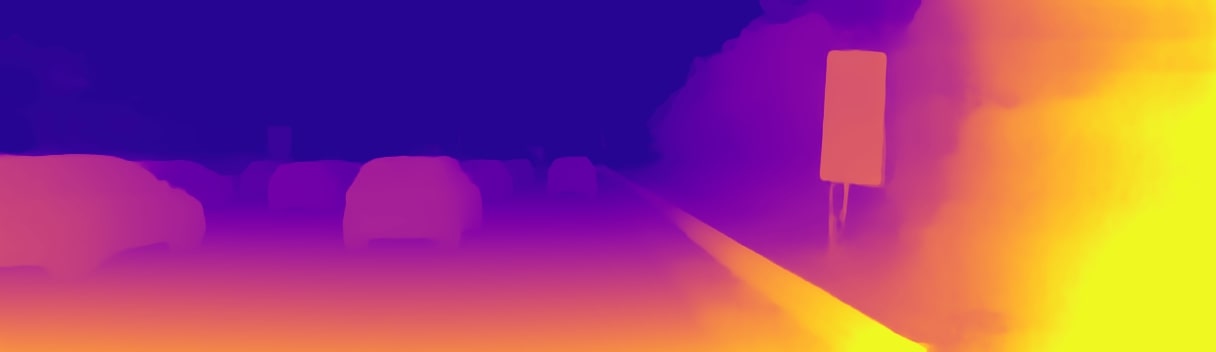}}
\subfloat[Pred. normals]{
\includegraphics[width=0.15\textwidth,height=1.1cm]{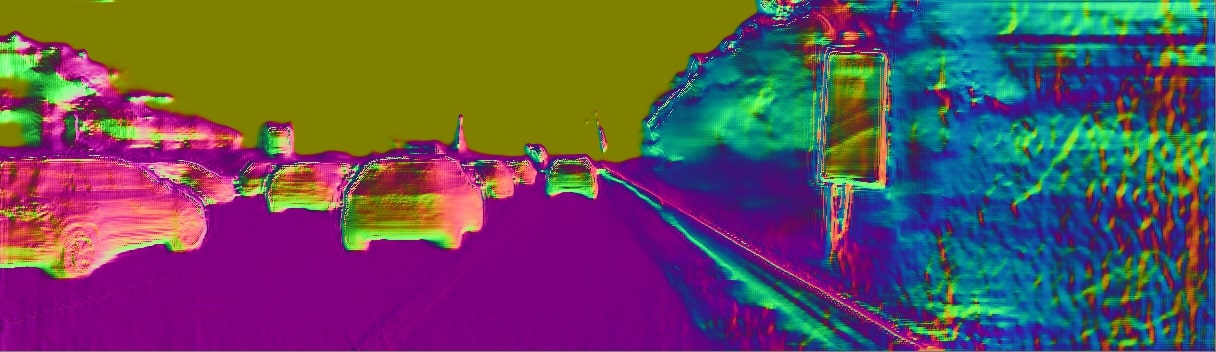}}
\vspace{-2mm}
\caption{\textbf{GUDA depth estimation results} on \emph{KITTI} and \emph{DDAD}. Our proposed mixed-batch training schedule produces much sharper and consistent depth maps, especially at longer ranges and in ``invalid" areas such as the sky.}
\label{fig:kitti_depth}
\vspace{-4mm}
\end{figure}

% \begin{figure}[t!]
% \centering
% \subfloat{
% \includegraphics[width=0.15\textwidth,height=2.2cm]{example-image-a}}
% \subfloat{
% \includegraphics[width=0.15\textwidth,height=2.2cm]{example-image-a}}
% \subfloat{
% \includegraphics[width=0.15\textwidth,height=2.2cm]{example-image-a}}
% \\ \vspace{-3mm}
% \subfloat{
% \includegraphics[width=0.15\textwidth,height=2.2cm]{example-image-a}}
% \subfloat{
% \includegraphics[width=0.15\textwidth,height=2.2cm]{example-image-a}}
% \subfloat{
% \includegraphics[width=0.15\textwidth,height=2.2cm]{example-image-a}}
% \\ \vspace{-3mm}
% \setcounter{subfigure}{0}
% \subfloat[Input image]{
% \includegraphics[width=0.15\textwidth,height=2.2cm]{example-image-a}}
% \subfloat[Pred. depth]{
% \includegraphics[width=0.15\textwidth,height=2.2cm]{example-image-a}}
% \subfloat[Pred. normals]{
% \includegraphics[width=0.15\textwidth,height=2.2cm]{example-image-a}}
% \caption{\textbf{Depth estimation results on the DDAD dataset}. }
% \label{fig:ddad_depth}
% \end{figure}

%% file: sections/06conclusion.tex
We introduce self-supervised monocular depth estimation as a proxy task for unsupervised sim-to-real transfer of semantic segmentation models. Our Geometric Unsupervised Domain Adaptation method, GUDA, combines dense synthetic labels with self-supervision from real-world unlabeled videos to bridge the sim-to-real domain gap. Although depth estimation is fundamentally a geometric task, we show it improves semantic representation transfer without any real-world semantic labels. Our multi-task self-supervised method outperforms other UDA approaches, while also improving monocular depth estimation. Furthermore, by introducing self-trained pseudo-labels as an extra source of supervision, we establish a new state of the art on this task.
Finally, we show that our method scales well with both the quantity and quality of synthetic data, highlighting its potential to eventually close the sim-to-real gap in challenging visual conditions like driving scenes.

%This paper introduces geometric supervision as a new tool for unsupervised domain adaptation (UDA). We show that the self-supervised photometric loss, commonly used to train monocular depth estimation models from raw videos, can be used to adapt features learned with supervision from virtual datasets through mixed-batch training. 
%By training a multi-task depth and semantic network, in combination with self-learned pseudo-labels, our proposed GUDA approach establishes a new state of the art in UDA for semantic segmentation in the SYNTHIA to Cityscapes benchmark. 
%We also show that mixed-batch training is a better domain adaptation tool for depth estimation than standard fine-tuning from pre-trained models, both improving results and enabling scale transfer between datasets.

%% file: suppmat.tex
\section{Network Architectures}

In Tab.~\ref{tab:sharednet} we describe in details the shared depth and semantic segmentation network used in our experiments. This architecture is based on recent developments in monocular depth estimation \cite{monodepth2}.
%, since this is the primary source of domain adaptation in GUDA.
Note that our proposed algorithm can be generalized to any multi-scale backbones.
We leave the exploration of architectures more suitable to jointly predict semantic segmentation~\cite{deeplab,CY2016Attention,PC2015Weak} and monocular depth~\cite{monodepth2, monodepth17} for future work. For the shared backbone we use a ResNet101~\cite{he2016deep} encoder, that produces feature maps with varying number of channels at increasingly lower resolutions (\#1, \#2, \#3, \#4, \#5 in Tab.~\ref{tab:sharednet}). These feature maps are used as skip connections for both the depth and the semantic segmentation decoders, through a series of convolutional layers followed by bilinear upsampling. For the depth decoder, at the final four upsampling stages (\#10, \#13, \#16, \#19 in Tab.~\ref{tab:sharednet}) an inverse depth layer is used to produce estimates within a minimum and maximum depth range:
\begin{equation}
    \frac{1}{d_{u,v}} = \frac{1}{d_{max}} + \Big( \frac{1}{d_{min}} - \frac{1}{d_{max}} \Big) Sigmoid(f_{u,v})
\label{eq:inv_depth}
\end{equation}
All four scales are used to calculate the self-supervised photometric loss (with results averaged per-batch, per-scale and per-pixel), and only the final scale is used to calculate the supervised depth loss. During inference, only the final scale is used as depth prediction estimates. The semantic network is similar, with the difference that the outputs at each of the upsampling stages (\#9, \#11, \#13, \#15 in Tab.~\ref{tab:sharednet}) are instead concatenated (after bilinear upsampling to the highest resolution) and processed using a final convolutional layer to produce a $C$-dimensional logits vector for each pixel.

\input{suppmat/decoders}

\input{suppmat/pd_figure}

\input{suppmat/posenet}

\input{suppmat/qualitative}

Our pose network is described in Tab. \ref{tab:posenet}, and follows closely \cite{monodepth2}. It uses a ResNet18 backbone as encoder, followed by four convolutional layers with 256 channels. Finally, a global pooling layer outputs a 6-dimensional vector, containing $(x,y,z)$ translation and \emph{(roll, pitch, yaw)} rotation. We have experimented with a shared encoder for depth, semantic segmentation and pose, however as pointed out in \cite{monodepth2} performance degraded in this configuration.

\section{Parallel Domain}

\input{suppmat/pd}

\input{tables/semantic_synthia_cs}

\input{tables/vkitti2_kitti}

\input{tables/pd_ddad}

\section{Qualitative Results}

In Fig. \ref{fig:qualitative} we present semantic pointclouds estimated using GUDA+PL for unsupervised domain adaptation from \emph{Parallel Domain} to \emph{Cityscapes}. Because our multi-task network (Tab. \ref{tab:sharednet}) produces both depth and semantic segmentation estimates, we can lift the predicted semantic labels to 3D space using depth estimates and camera intrinsics. Each pixel is assigned a 3D coordinate in the camera frame of reference, as well as RGB colors and semantic logits. We emphasize that no real-world labels (depth maps or semantic classes) were used at any point during the training of this network, only image sequences. All labeled information was obtained from synthetic datasets, and adapted to better align with real-world data using our proposed GUDA approach to geometric unsupervised domain adaptation. 

\section{Detailed Tables}

We also present detailed tables to complement some results from the main paper. In particular, Table \ref{tab:synthia_cs} expands Table 1 from the main paper, showing per-class results on the \emph{Cityscapes} dataset of the various methods we use as comparison to validate the improvements of our proposed GUDA approach. Similarly, Tables \ref{tab:vkitti2_kitti} and \ref{tab:pd_ddad} expand Figures 5 and 6 from the main paper, showing respectively GUDA results from our \emph{VKITTI2} to \emph{KITTI} and \emph{PD} to \emph{DDAD} experiments relative to \emph{source-only} and \emph{DANN} \cite{dann}.

%% file: suppmat/decoders.tex
\begin{table}[!t]%
\small
\centering
\resizebox{0.9\linewidth}{!}{
\begin{tabular}[b]{l|l|c}
\toprule
& \textbf{Layer Description} & \textbf{Out. Dimension} \\ 
\toprule
& RGB image & 3$\times$H$\times$W \\ 
\midrule
\multicolumn{3}{c}{\textbf{ResNet101 Encoder}} \\
\midrule
\textbf{\#1} & {Intermediate Features \#1} & 256$\times$H/2$\times$W/2 \\
\textbf{\#2} & {Intermediate Features \#2} & 256$\times$H/4$\times$W/4 \\
\textbf{\#3} & {Intermediate Features \#3} & 512$\times$H/8$\times$W/8 \\
\textbf{\#4} & {Intermediate Features \#4} & 1024$\times$H/16$\times$W/16 \\
\textbf{\#5} & {Latent Features} & 2048$\times$H/32$\times$W/32 \\
\midrule
\multicolumn{3}{c}{\textbf{Depth Decoder}}  \\
\midrule
\#6 & Conv2d (\#5) $\rightarrow$ ELU $\rightarrow$ Upsample & 256$\times$H/16$\times$W/16 \\
\#7 & Conv2d (\#6 $\oplus$ \#4) $\rightarrow$ ELU & 256$\times$H/16$\times$W/16  \\
\#8 & Conv2d (\#7) $\rightarrow$ ELU $\rightarrow$ Upsample & 128$\times$H/8$\times$W/8  \\
\#9 & Conv2d (\#8 $\oplus$ \#3) $\rightarrow$ ELU & 128$\times$H/8$\times$W/8  \\
\textbf{\#10} & Conv2d (\#9) $\rightarrow$ InvDepth & 1$\times$H/8$\times$W/8  \\
\#11 & Conv2d (\#9) $\rightarrow$ ELU $\rightarrow$ Upsample & 64$\times$H/4$\times$W/4  \\
\#12 & Conv2d (\#11 $\oplus$ \#2) $\rightarrow$ ELU & 64$\times$H/4$\times$W/4  \\
\textbf{\#13} & Conv2d (\#12) $\rightarrow$ InvDepth & 1$\times$H/4$\times$W/4  \\
\#14 & Conv2d (\#12) $\rightarrow$ ELU $\rightarrow$ Upsample & 32$\times$H/2$\times$W/2  \\
\#15 & Conv2d (\#14 $\oplus$ \#1) $\rightarrow$ ELU & 32$\times$H/2$\times$W/2  \\
\textbf{\#16} & Conv2d (\#15) $\rightarrow$ InvDepth & 1$\times$H/2$\times$W/2  \\
\#17 & Conv2d (\#15) $\rightarrow$ ELU $\rightarrow$ Upsample & 16$\times$H$\times$W  \\
\#18 & Conv2d (\#17) $\rightarrow$ ELU & 16$\times$H$\times$W  \\
\textbf{\#19} & Conv2d (\#18) $\rightarrow$ InvDepth & 1$\times$H$\times$W  \\
\midrule
\multicolumn{3}{c}{\textbf{Semantic Decoder}}  \\
\midrule
\#6 & Conv2d (\#5) $\rightarrow$ ELU $\rightarrow$ Upsample & 256$\times$H/16$\times$W/16 \\
\#7 & Conv2d (\#6 $\oplus$ \#4) $\rightarrow$ ELU & 256$\times$H/16$\times$W/16  \\
\#8 & Conv2d (\#7) $\rightarrow$ ELU $\rightarrow$ Upsample & 128$\times$H/8$\times$W/8  \\
\#9 & Conv2d (\#8 $\oplus$ \#3) $\rightarrow$ ELU & 128$\times$H/8$\times$W/8  \\
\#10 & Conv2d (\#9) $\rightarrow$ ELU $\rightarrow$ Upsample & 64$\times$H/4$\times$W/4  \\
\#11 & Conv2d (\#10 $\oplus$ \#2) $\rightarrow$ ELU & 64$\times$H/4$\times$W/4  \\
\#12 & Conv2d (\#11) $\rightarrow$ ELU $\rightarrow$ Upsample & 32$\times$H/2$\times$W/2  \\
\#13 & Conv2d (\#12 $\oplus$ \#1) $\rightarrow$ ELU & 32$\times$H/2$\times$W/2  \\
\#14 & Conv2d (\#13) $\rightarrow$ ELU $\rightarrow$ Upsample & 16$\times$H$\times$W  \\
\#15 & Conv2d (\#14) $\rightarrow$ ELU & 16$\times$H$\times$W  \\
\textbf{\#16} & Conv2d (\#9 $\oplus$ \#11 $\oplus$ \#13 $\oplus$ \#15) & C$\times$H$\times$W  \\
\bottomrule
\end{tabular}
}
\\
%%%%%%%%%%%%%%%%%%%%%%%%%%%%%%%%%%%%%%%%%%%%%%%%%%%
\caption{
\textbf{Depth and semantic segmentation multi-task network}. We use a ResNet101 backbone as encoder, that outputs intermediate features at different resolutions. These intermediate features are used as skip connections in different stages of the semantic and depth decoders. ELU are Exponential Linear Units \cite{clevert2016fast}, \emph{Upsample} denotes bilinear interpolation, \emph{InvDepth} is an inverse depth layer (Eq. \ref{eq:inv_depth}), and $\oplus$ denotes feature concatenation.
}
%%%%%%%%%%%%%%%%%%%%%%%%%%%%%%%%%%%%%%%%%%%%%%%%%%%
\label{tab:sharednet}
\end{table}

%% file: suppmat/pd_figure.tex
\begin{figure*}[t!]
\subfloat{
\includegraphics[width=0.2\textwidth]{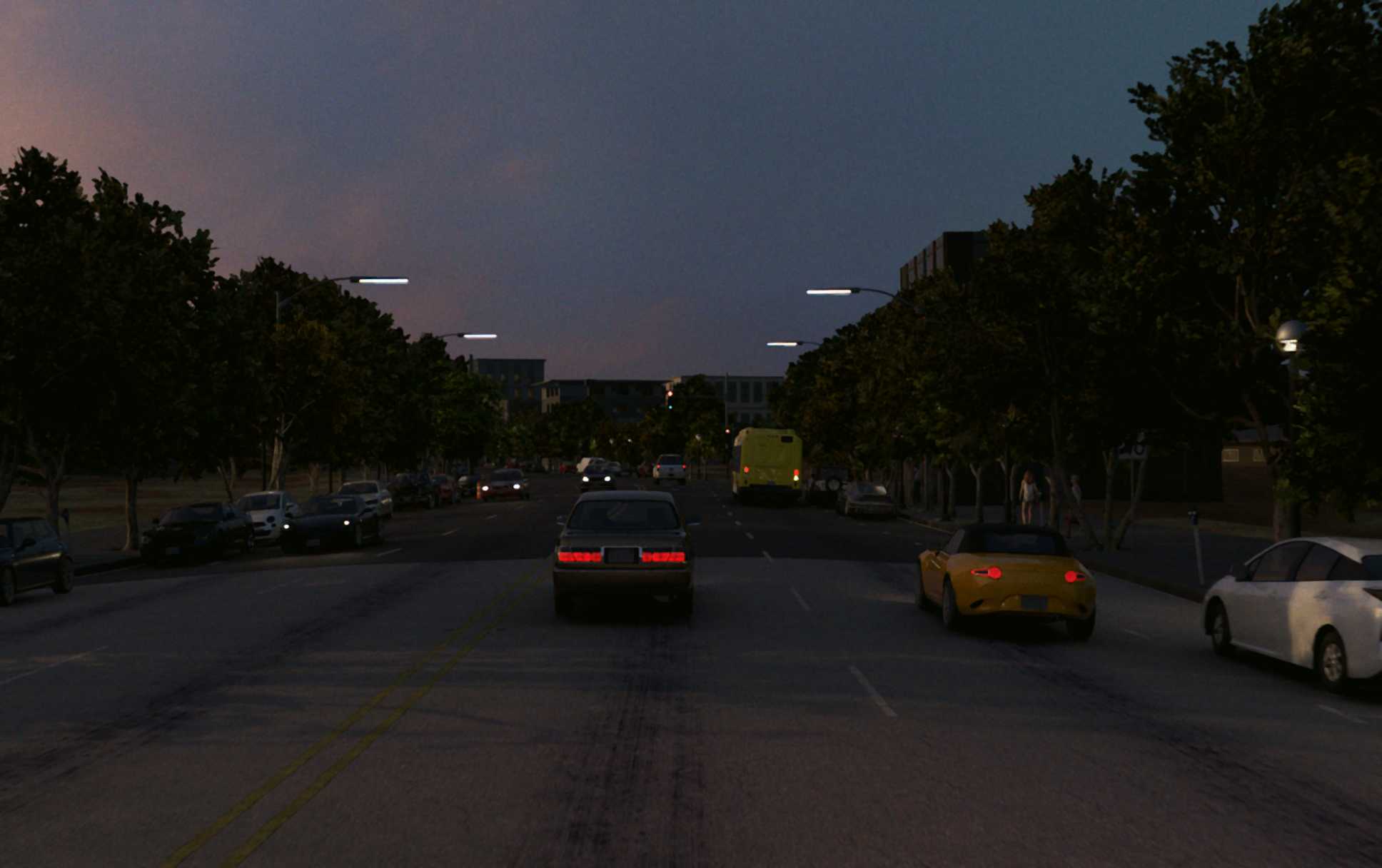}
\includegraphics[width=0.2\textwidth]{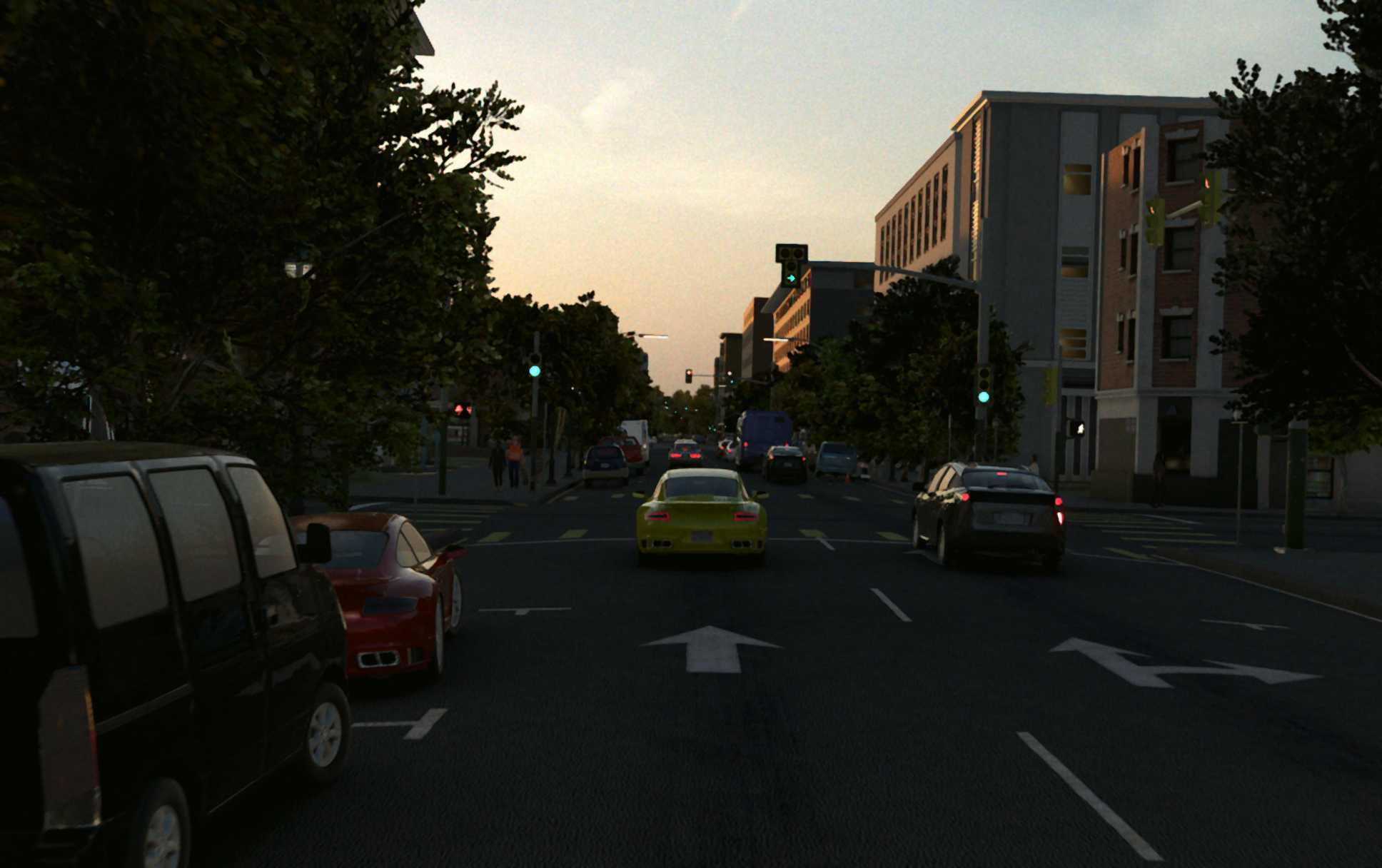}
\includegraphics[width=0.2\textwidth]{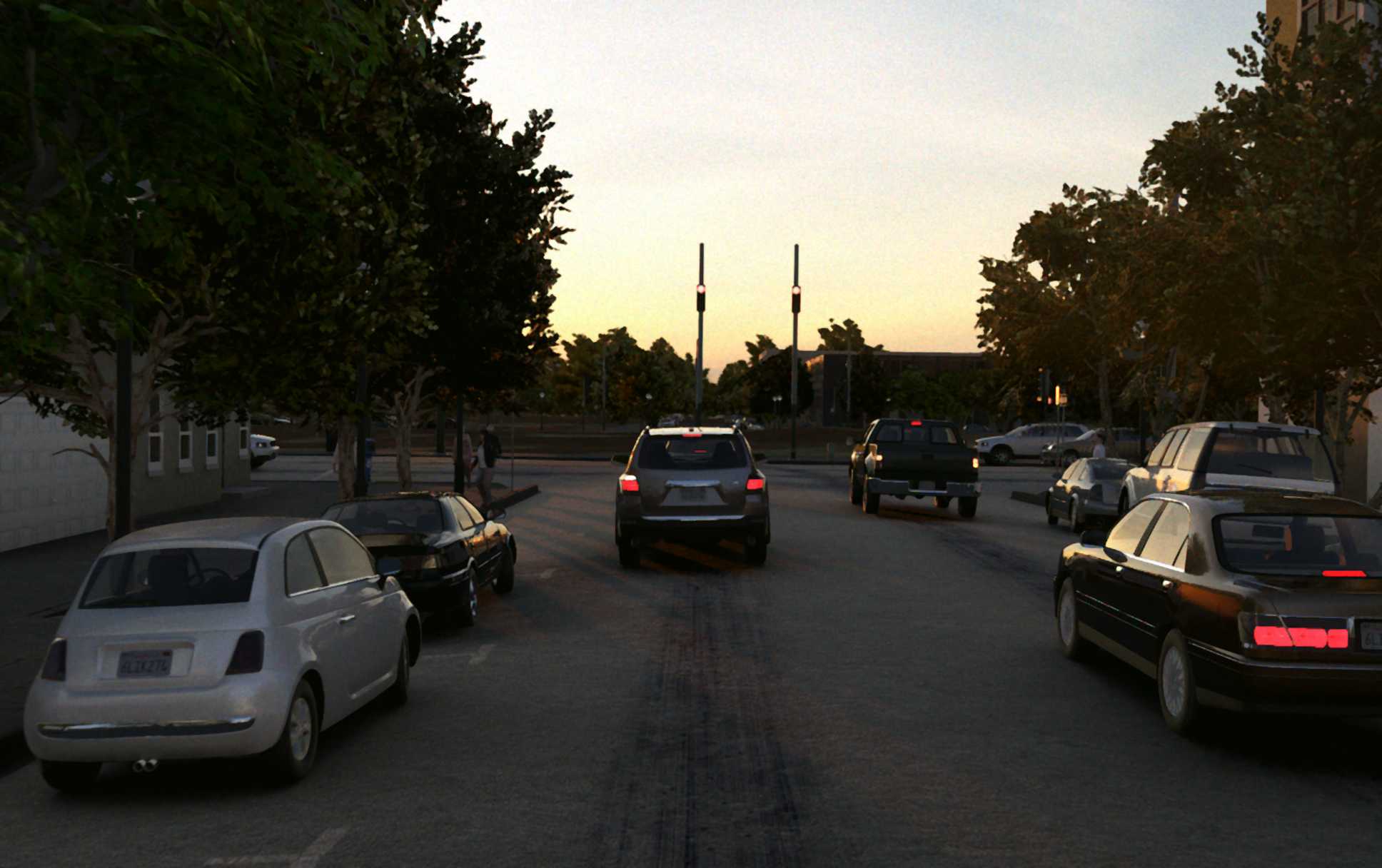}
\includegraphics[width=0.2\textwidth]{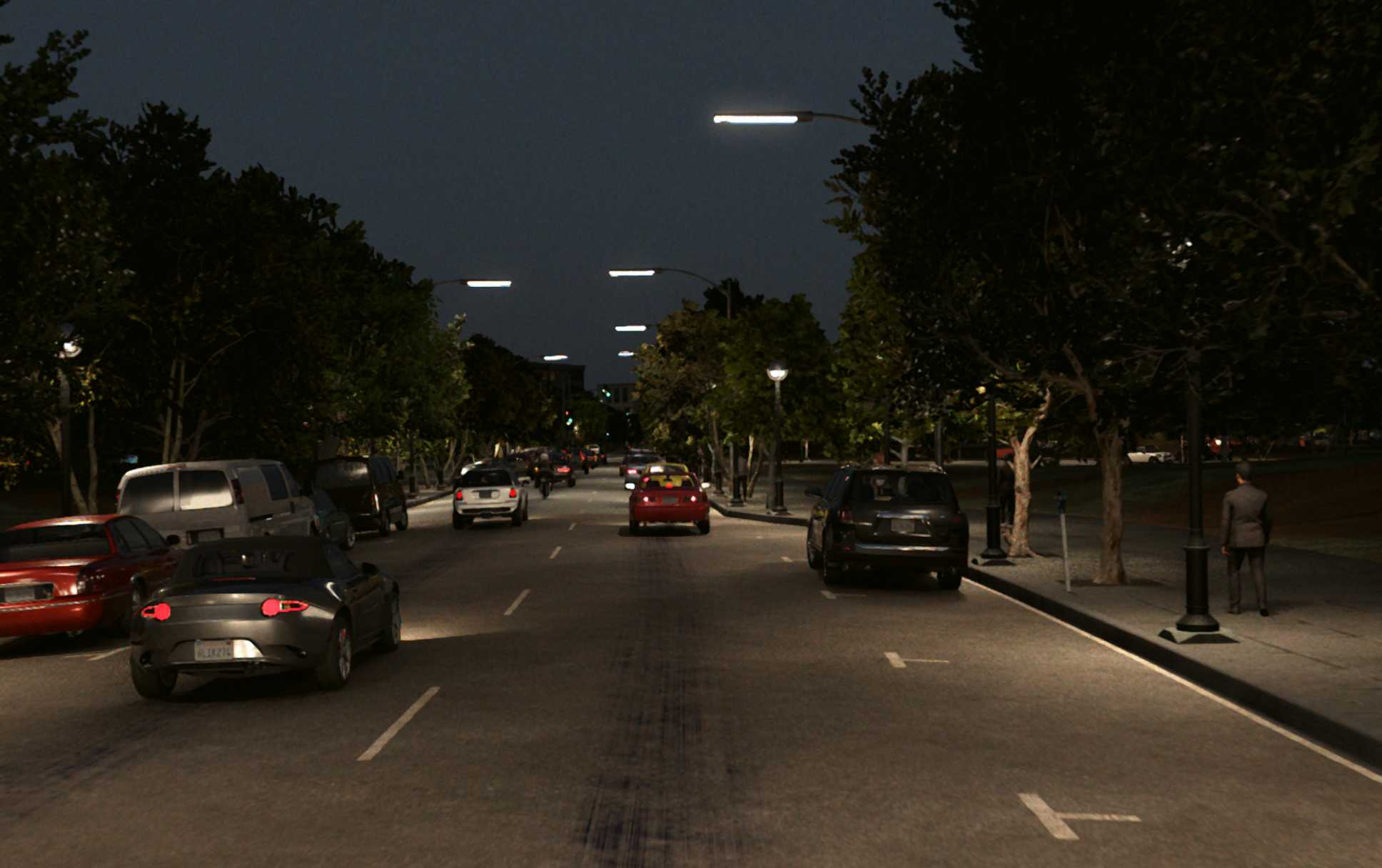}
\includegraphics[width=0.2\textwidth]{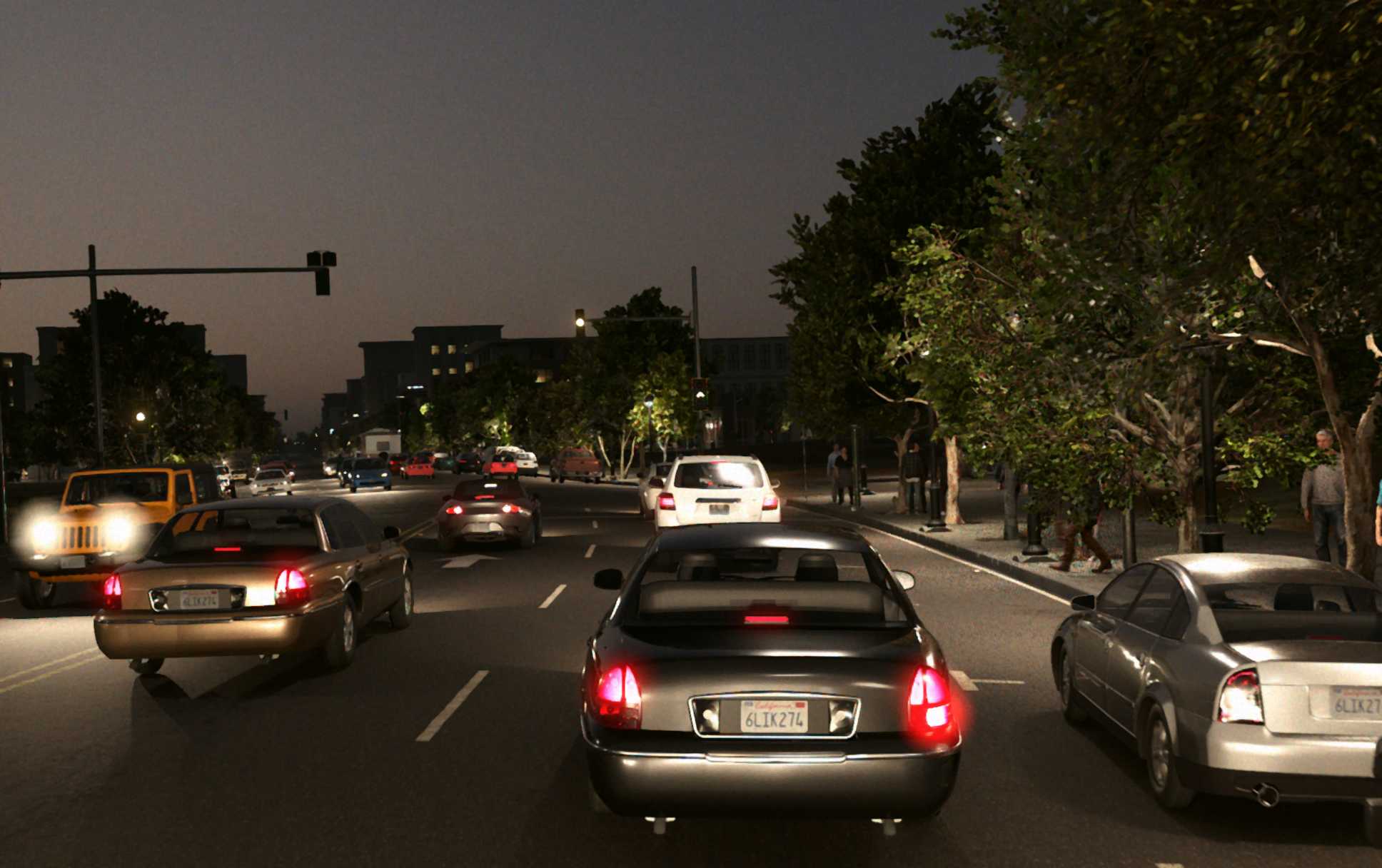}
}\vspace{-3mm} \\ 
\subfloat{
\includegraphics[width=0.2\textwidth]{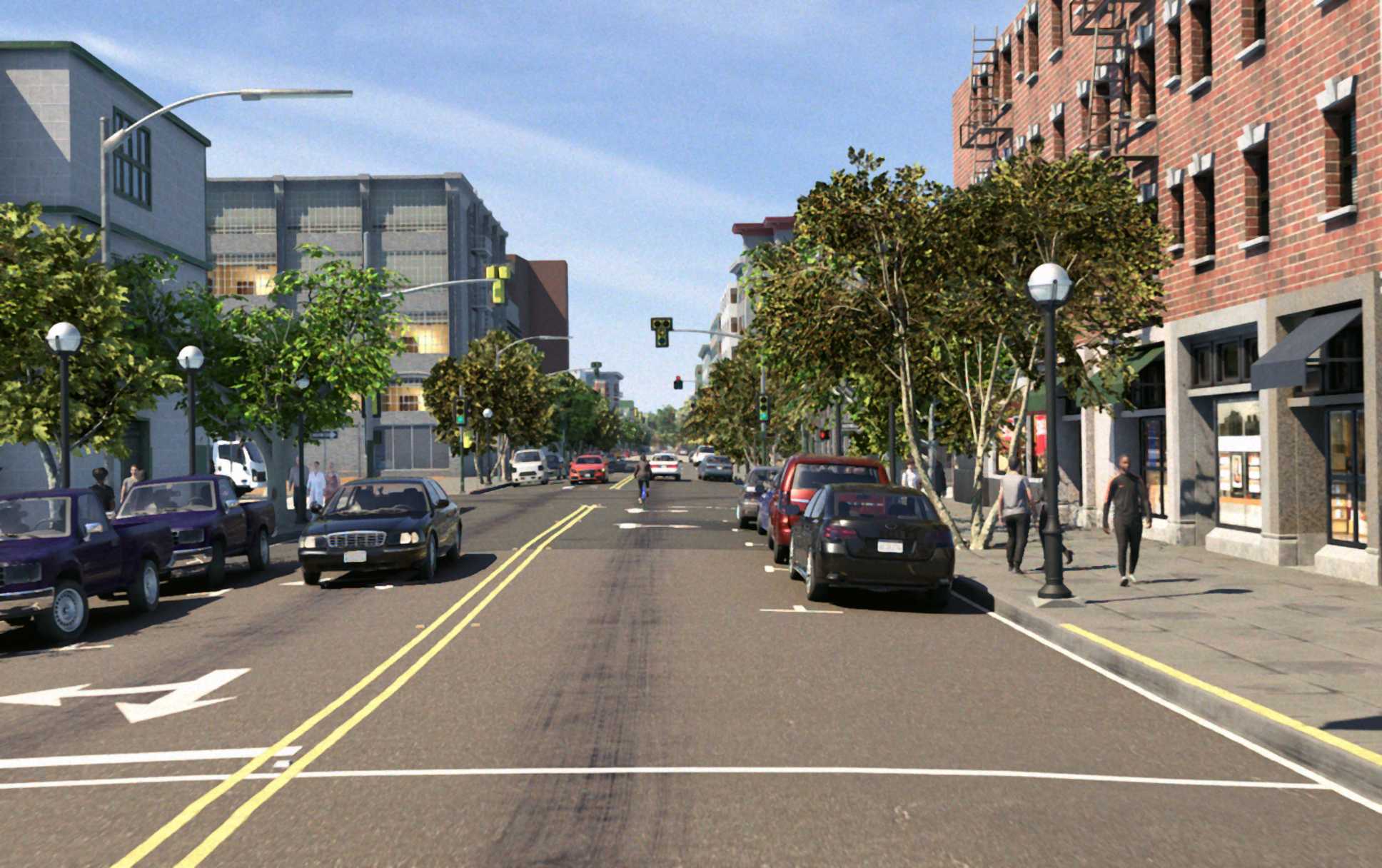}
\includegraphics[width=0.2\textwidth]{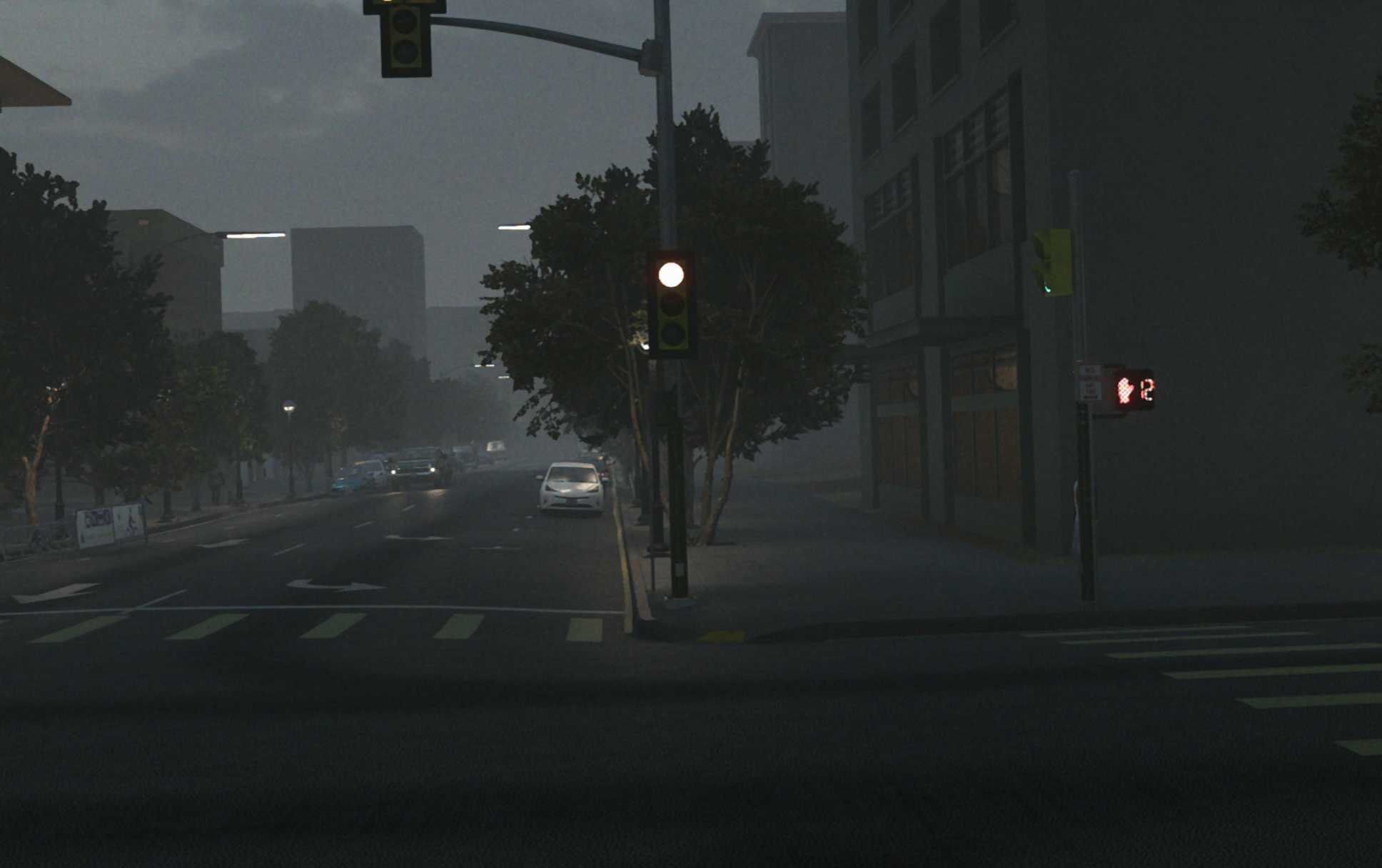}
\includegraphics[width=0.2\textwidth]{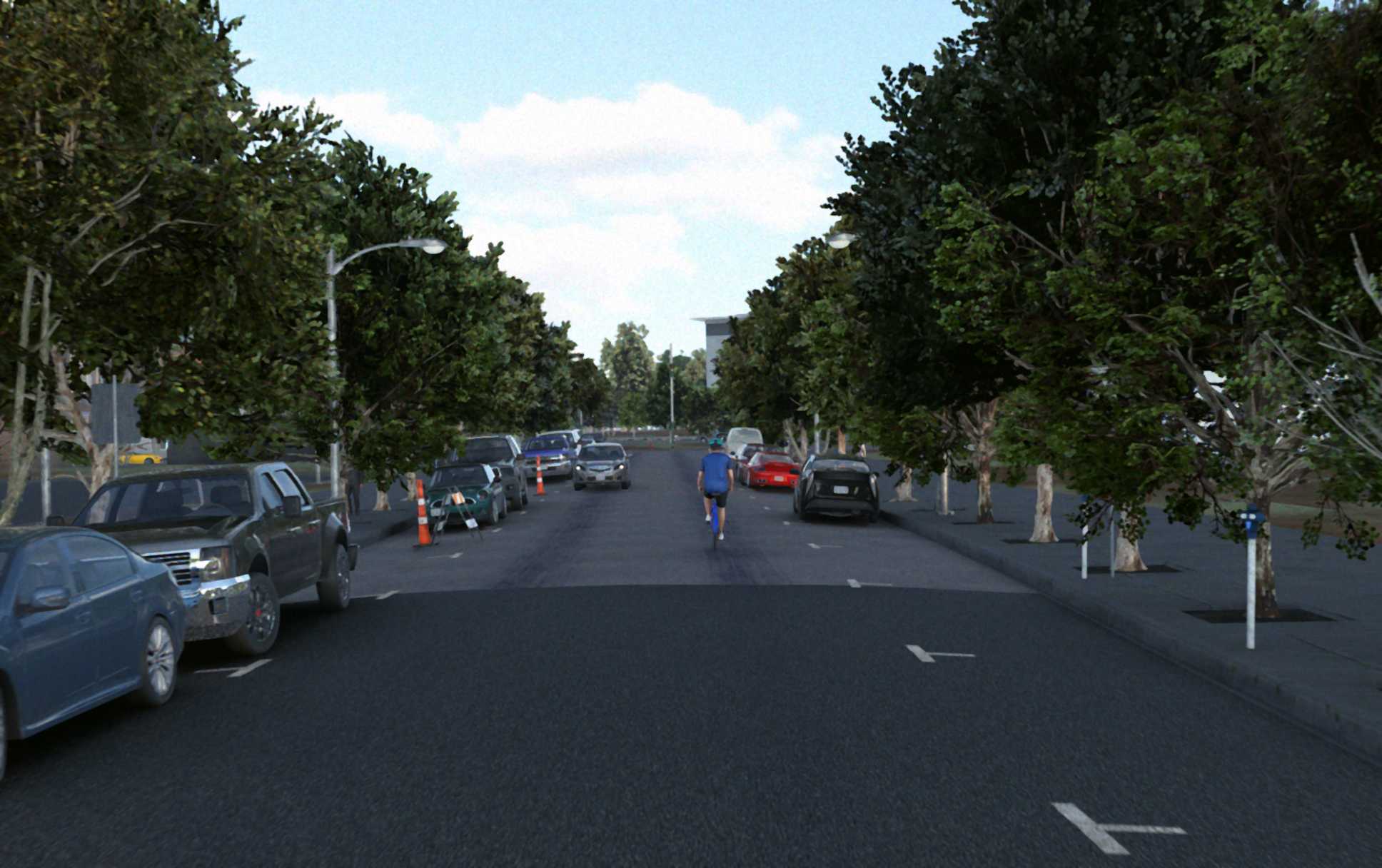}
\includegraphics[width=0.2\textwidth]{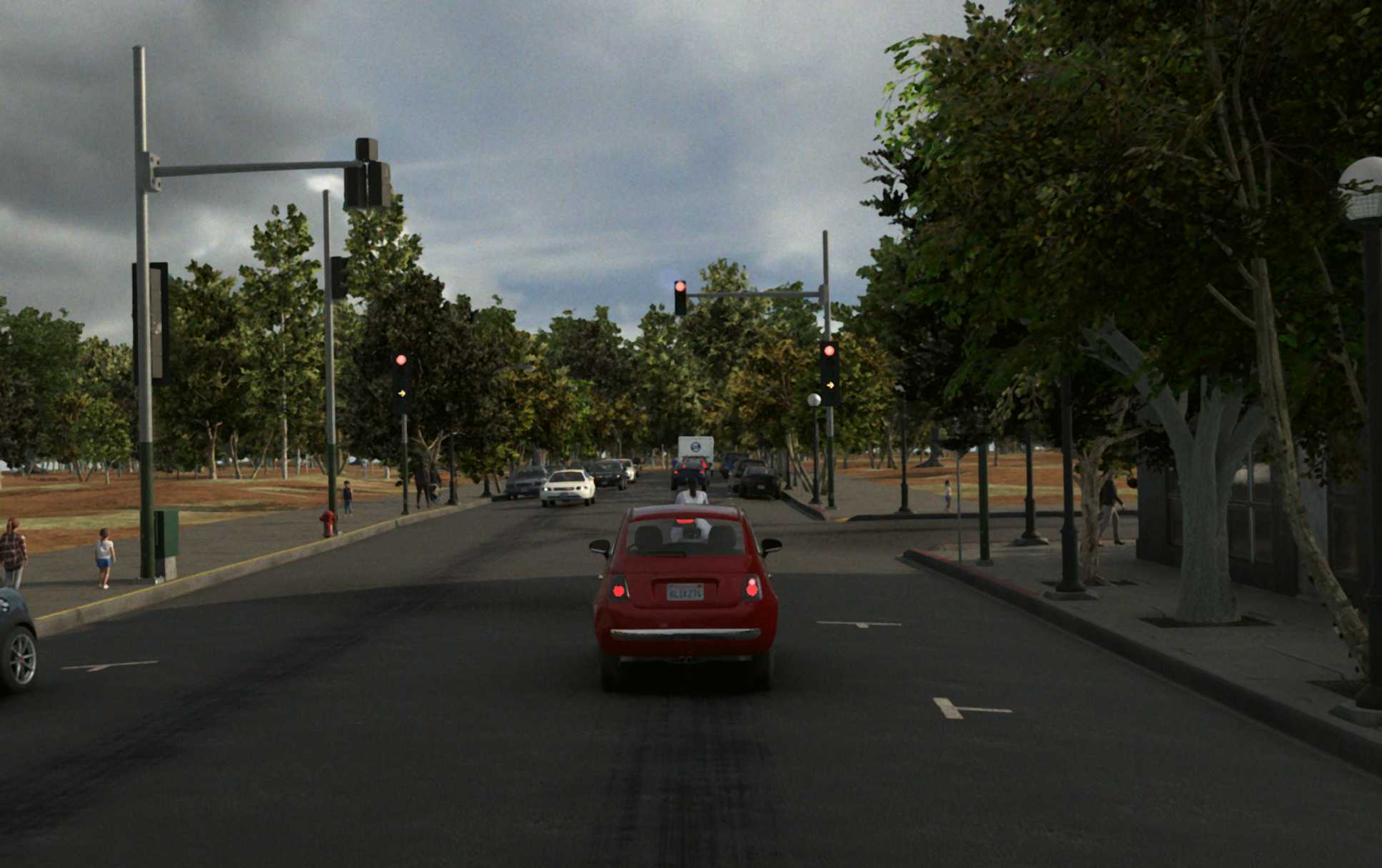}
\includegraphics[width=0.2\textwidth]{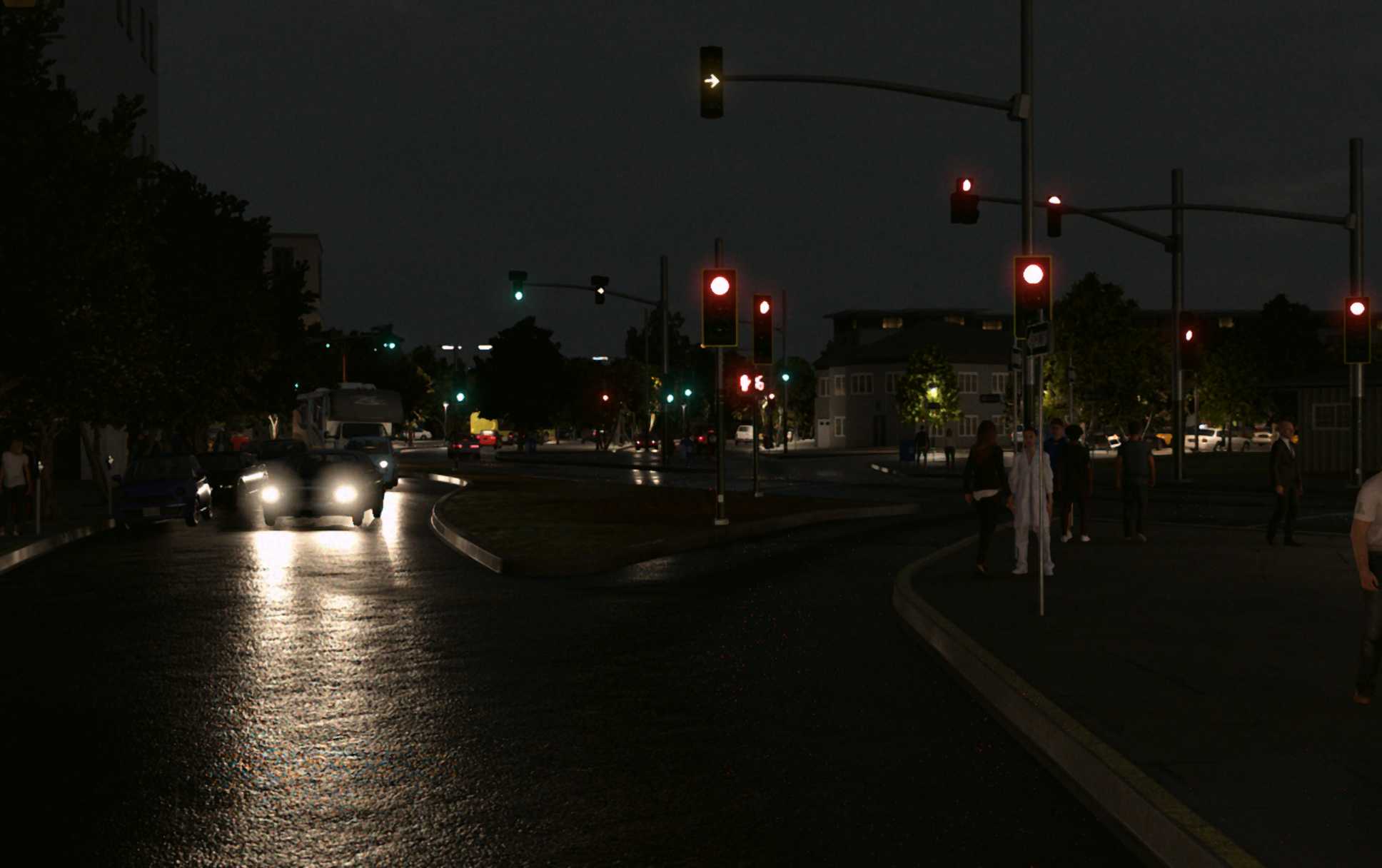}
}\vspace{-3mm} \\ 
\subfloat{
\includegraphics[width=0.2\textwidth]{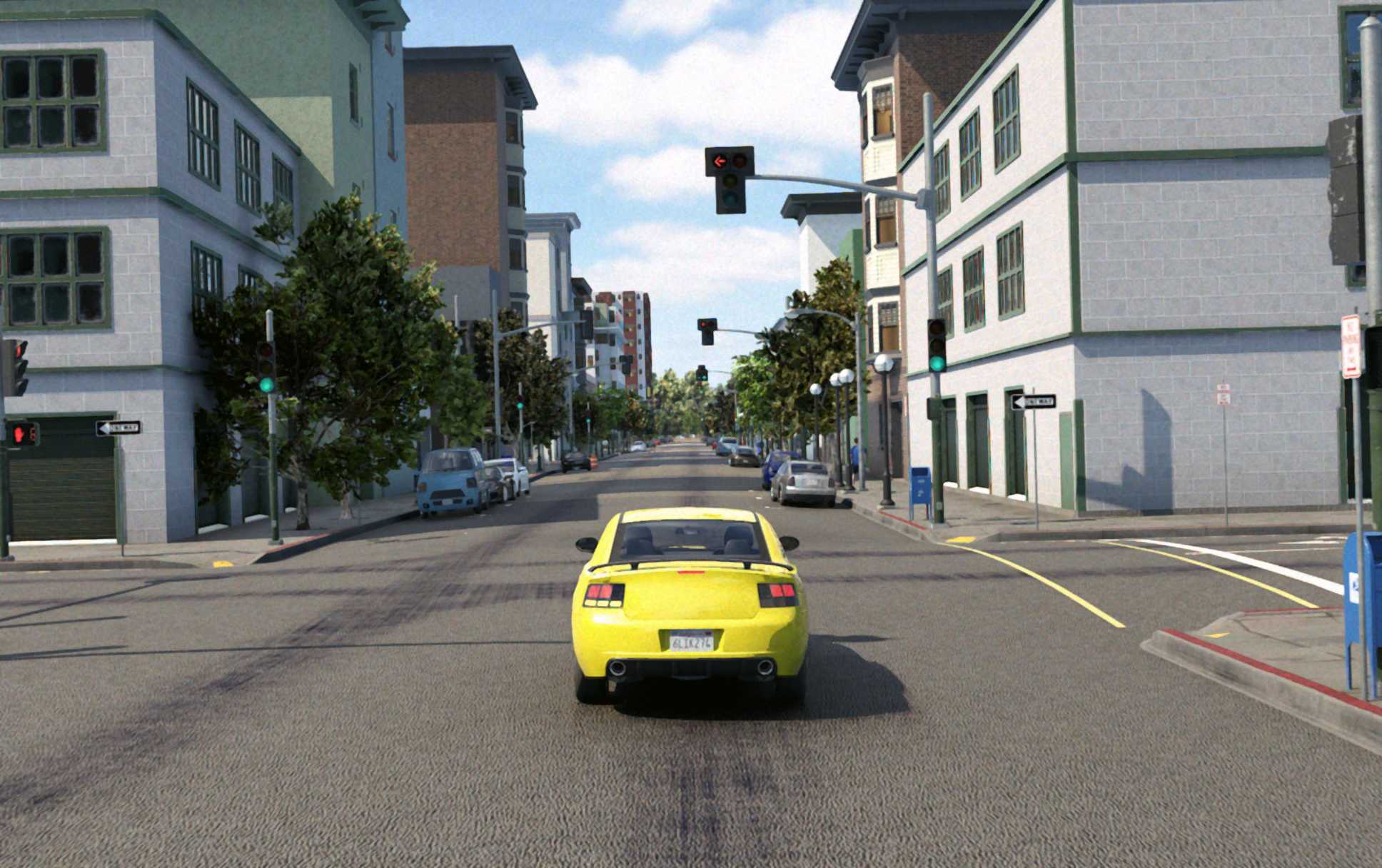}
\includegraphics[width=0.2\textwidth]{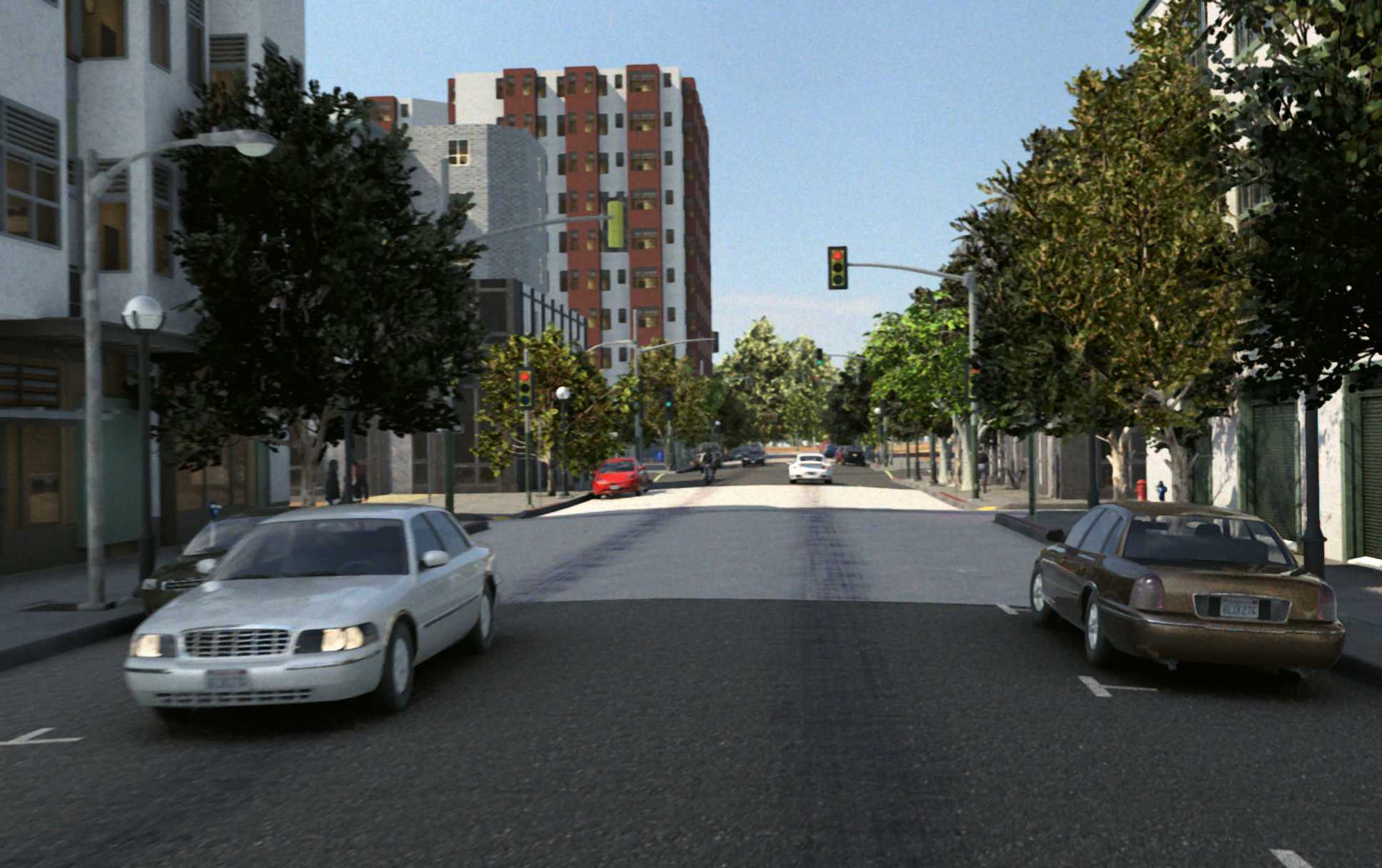}
\includegraphics[width=0.2\textwidth]{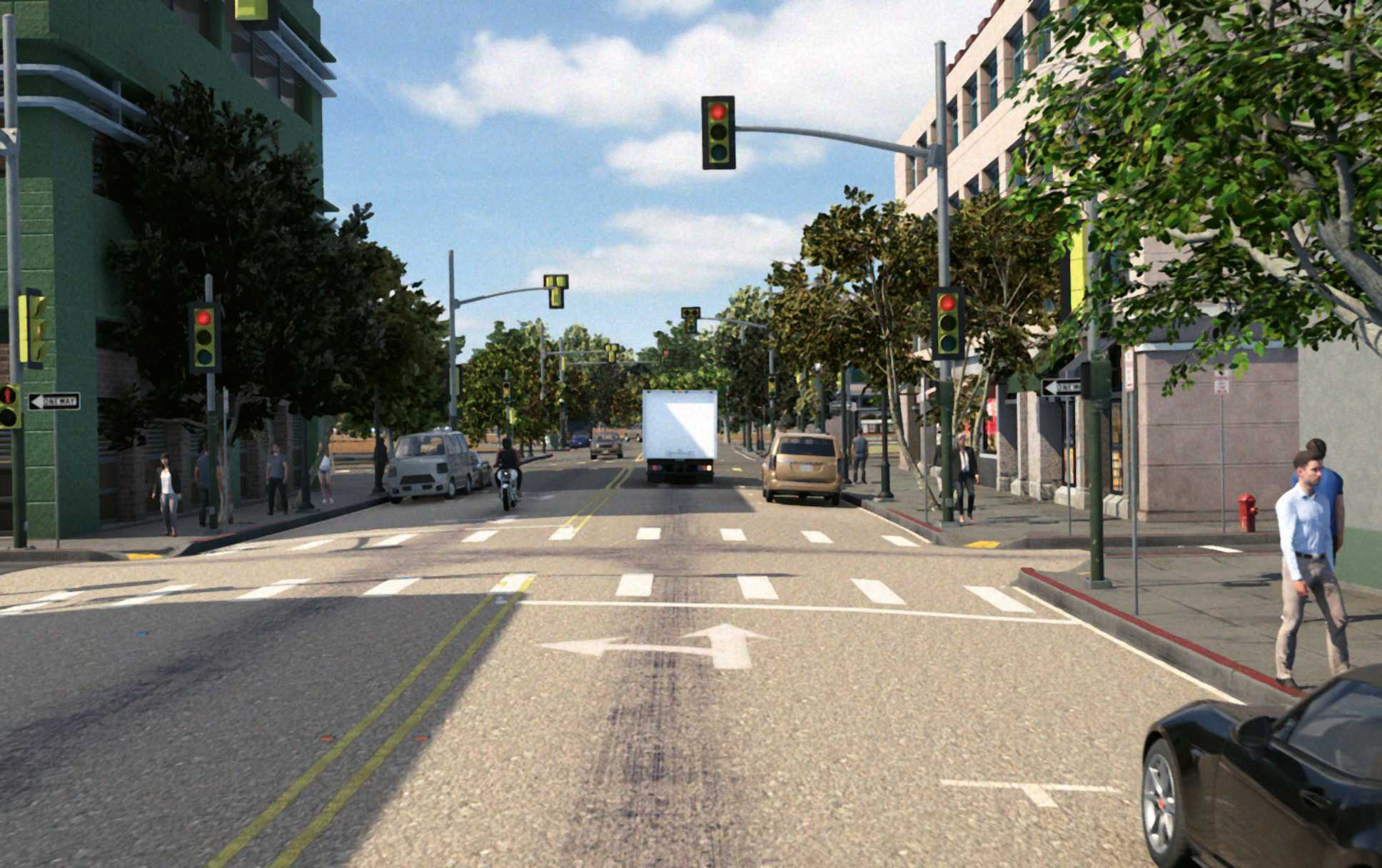}
\includegraphics[width=0.2\textwidth]{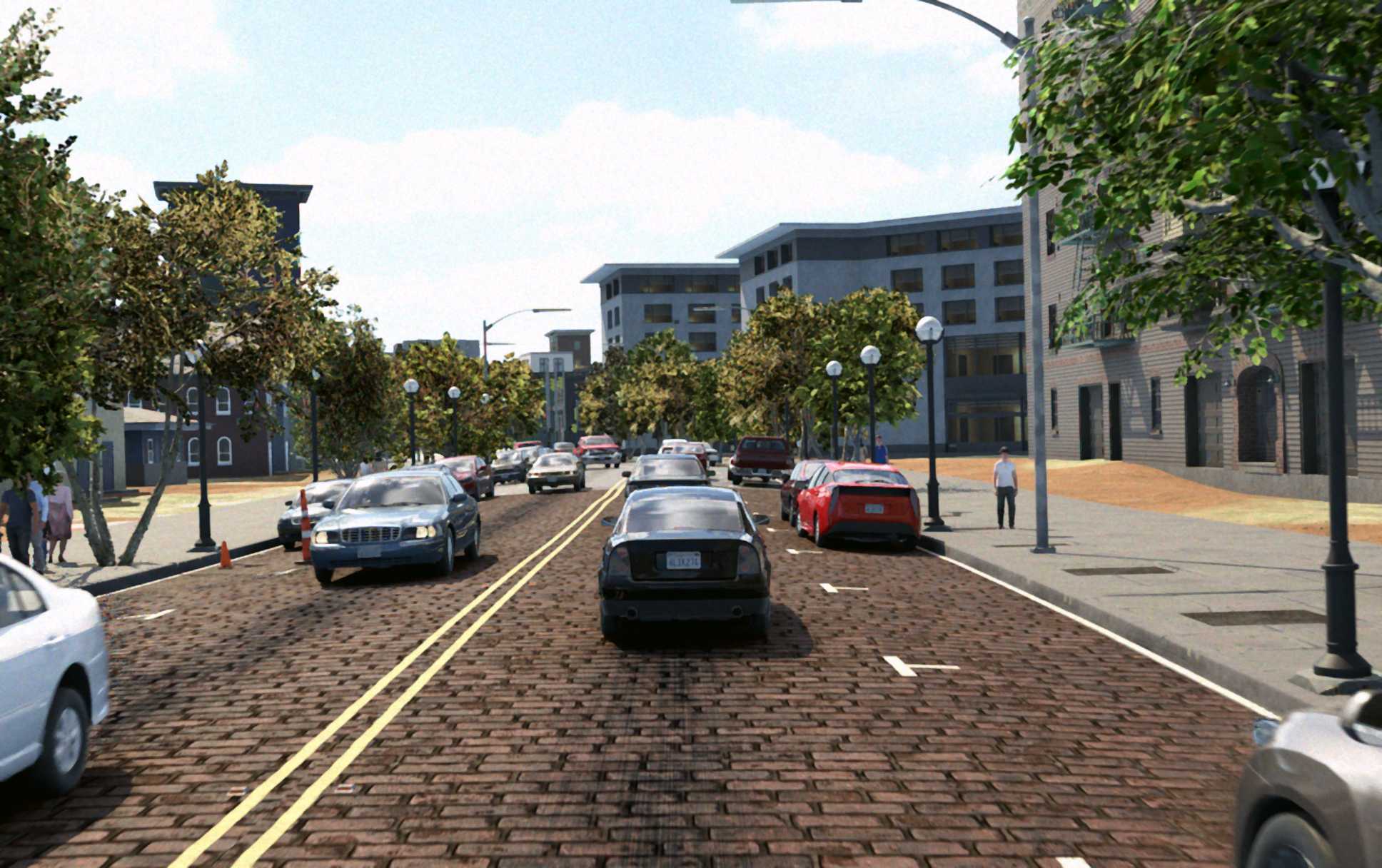}
\includegraphics[width=0.2\textwidth]{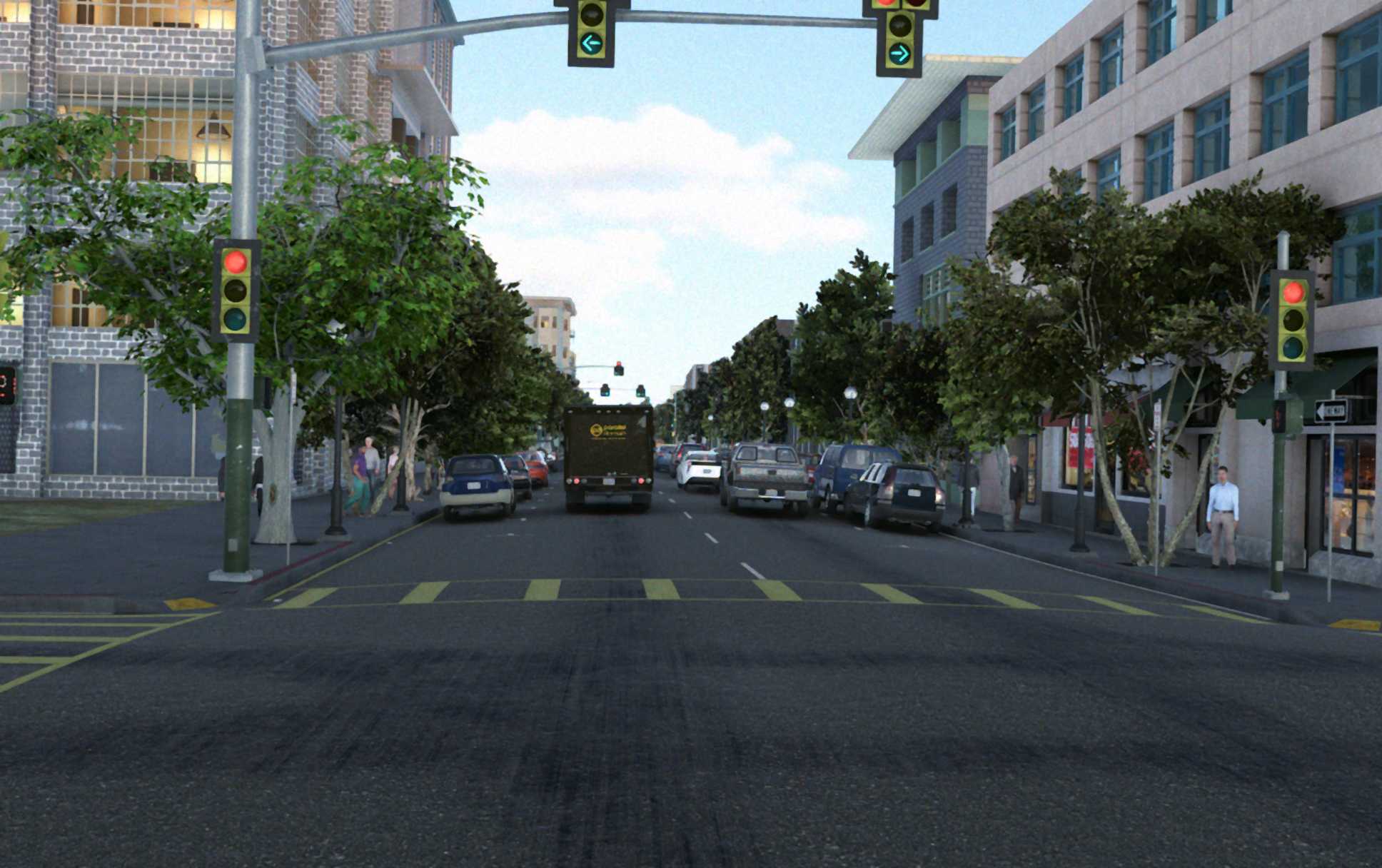}
}\vspace{-3mm} \\ 
\subfloat{
\includegraphics[width=0.2\textwidth]{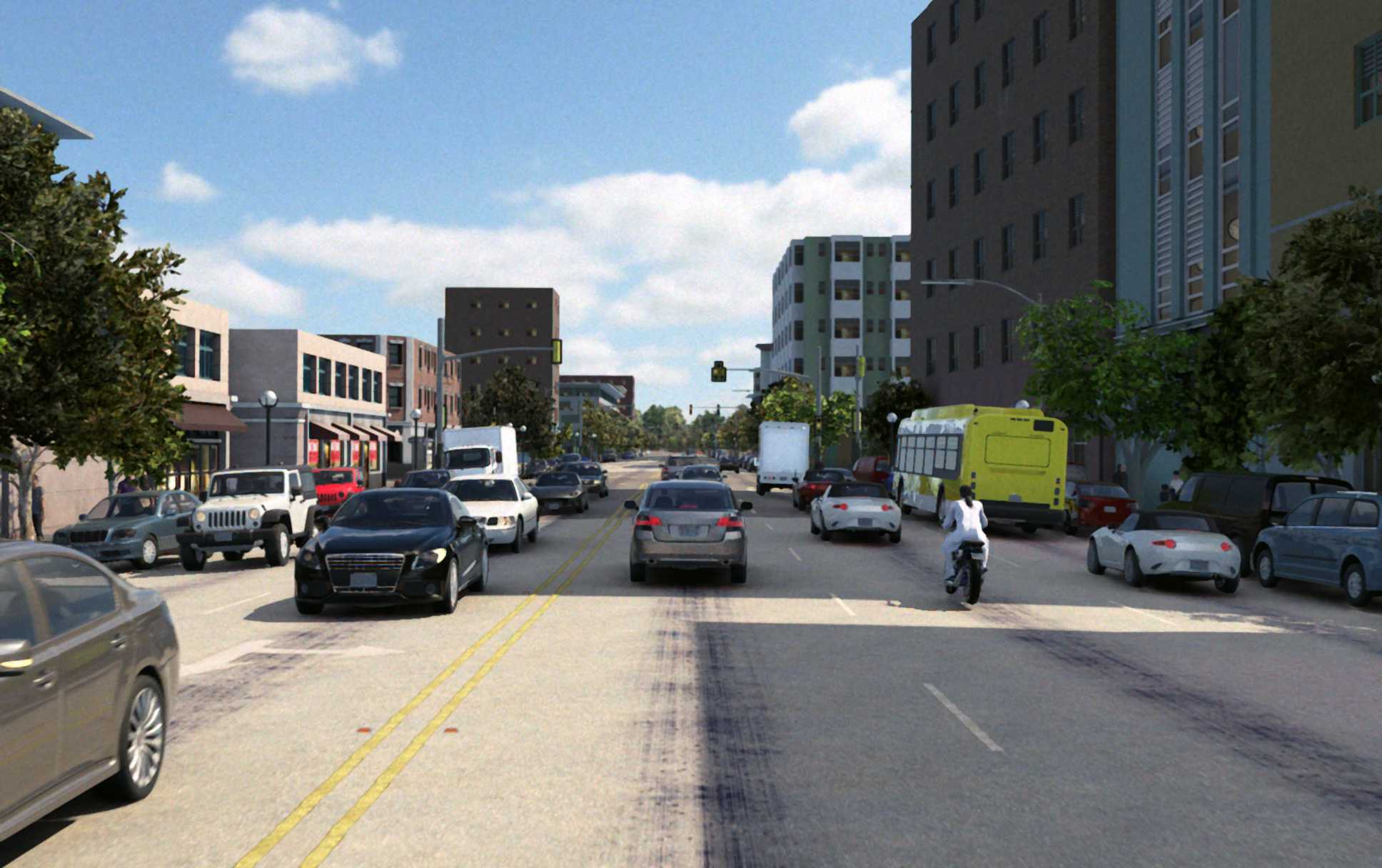}
\includegraphics[width=0.2\textwidth]{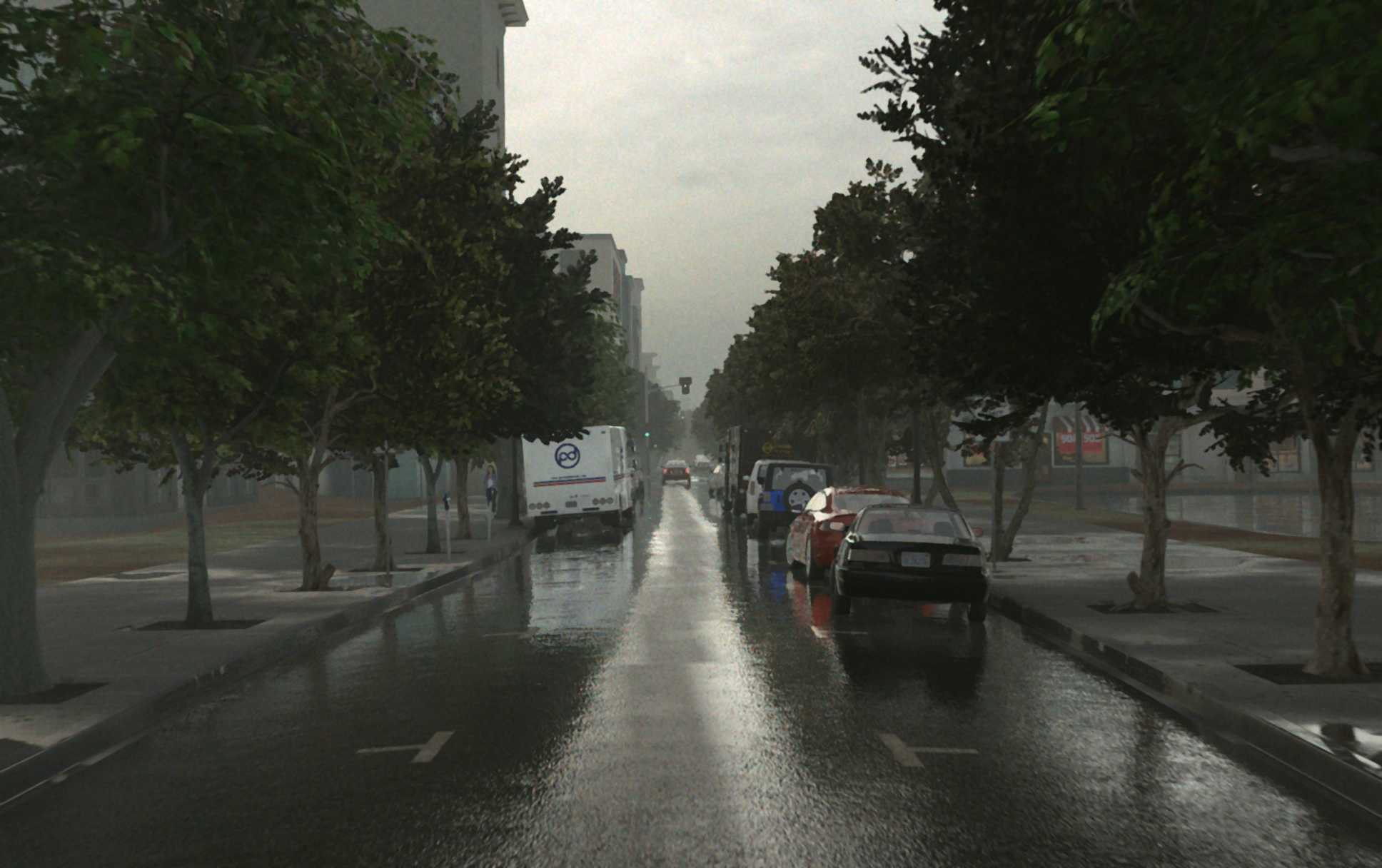}
\includegraphics[width=0.2\textwidth]{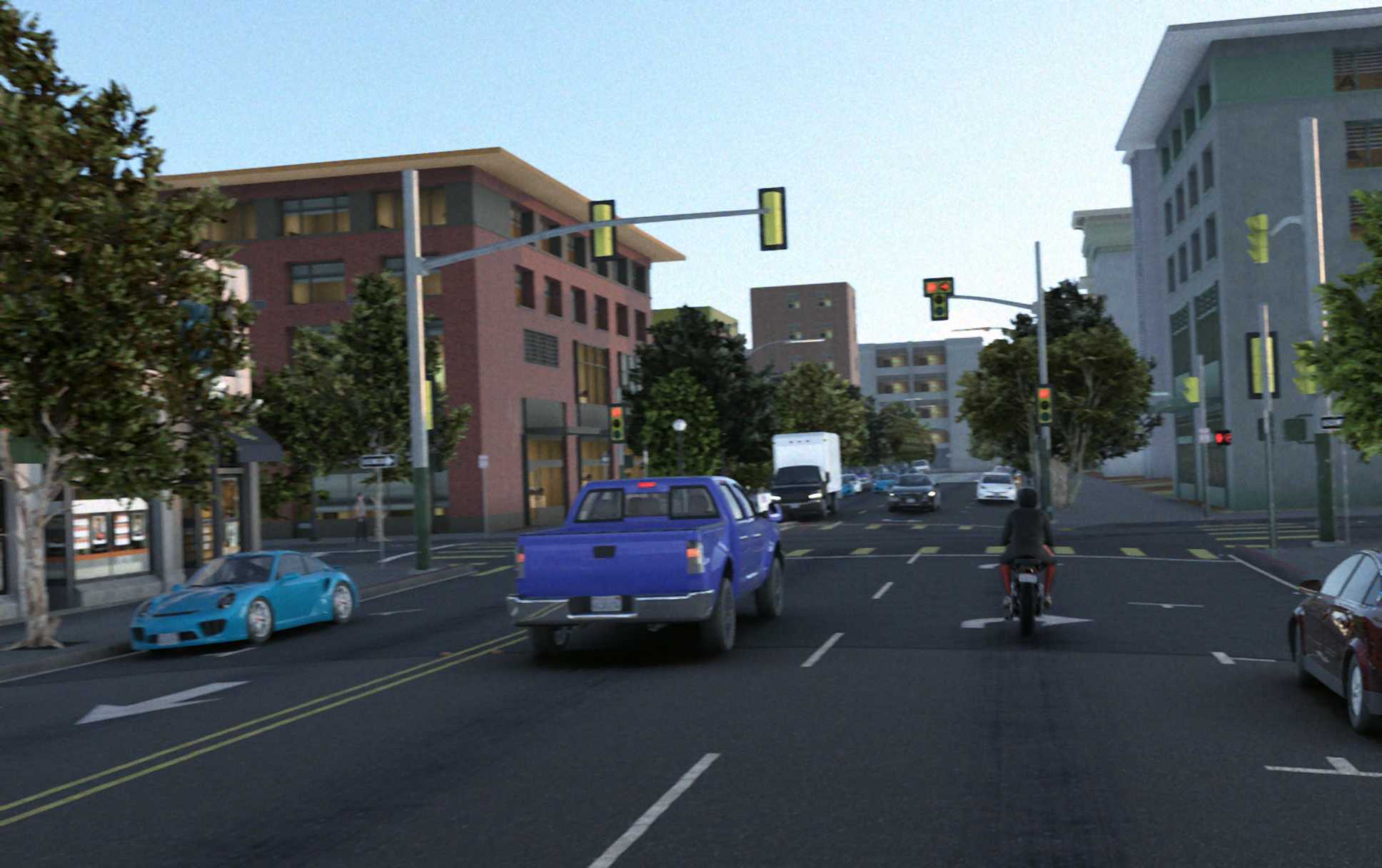}
\includegraphics[width=0.2\textwidth]{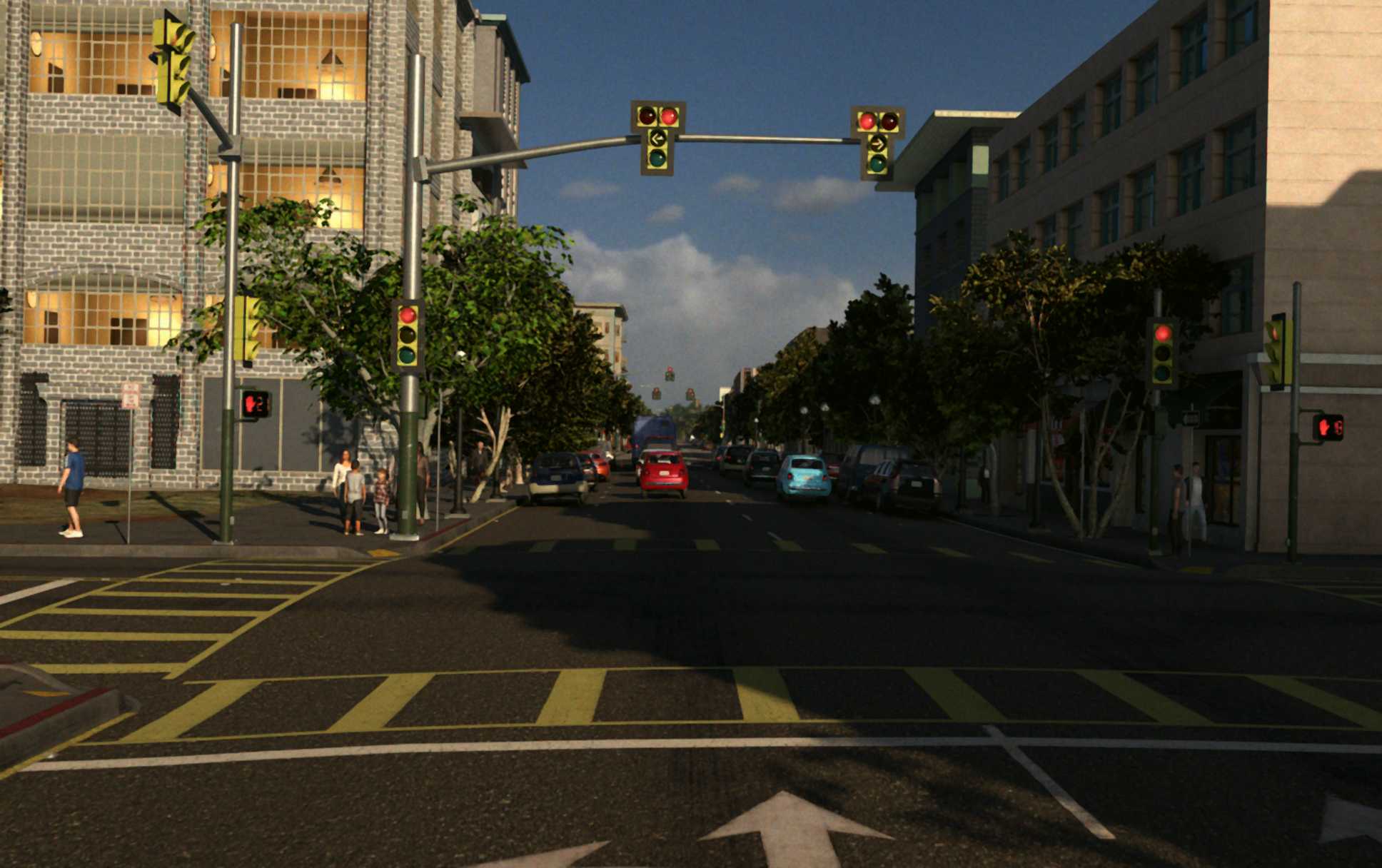}
\includegraphics[width=0.2\textwidth]{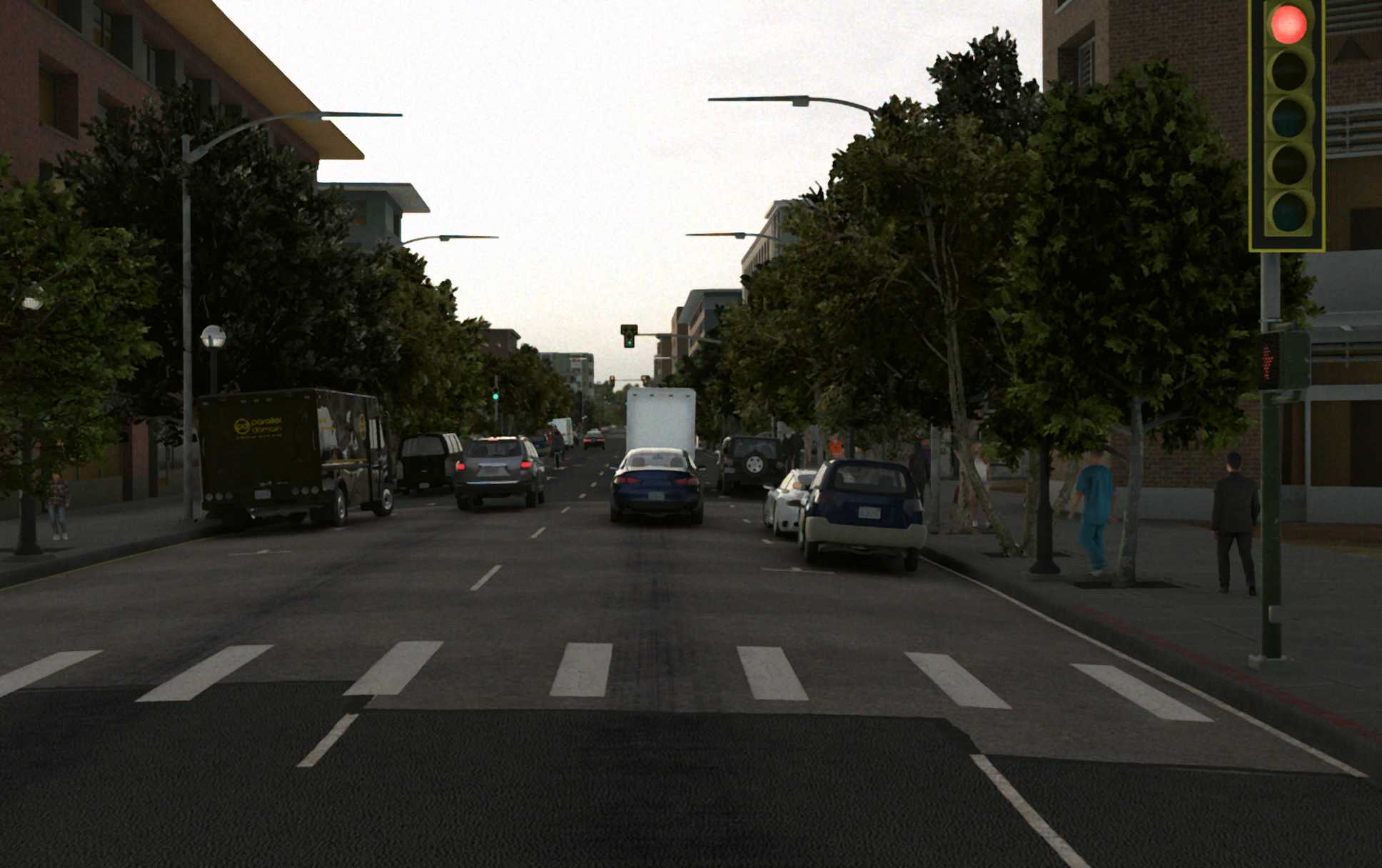}
}\vspace{-3mm} \\ 
\subfloat{
\includegraphics[width=0.2\textwidth]{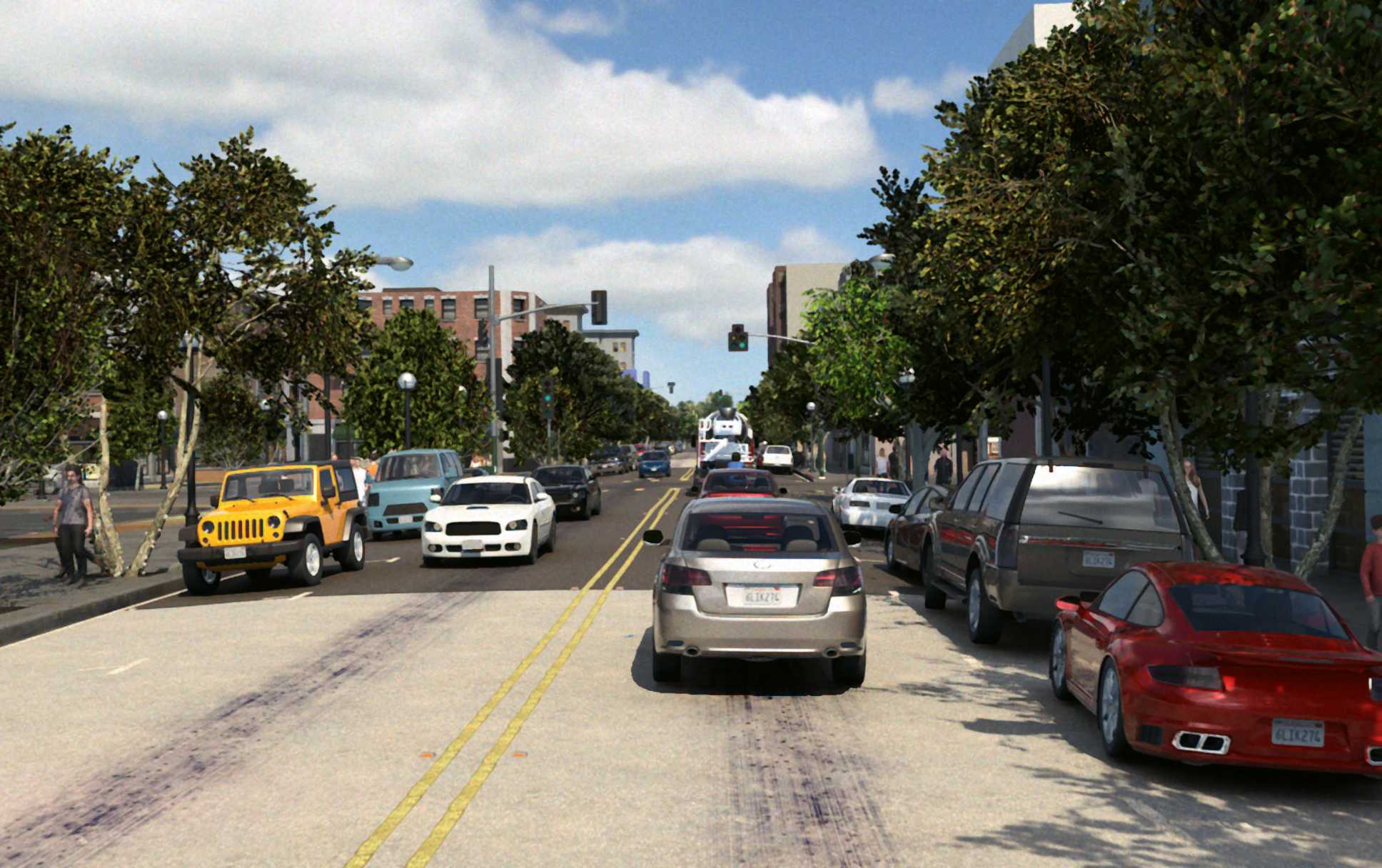}
\includegraphics[width=0.2\textwidth]{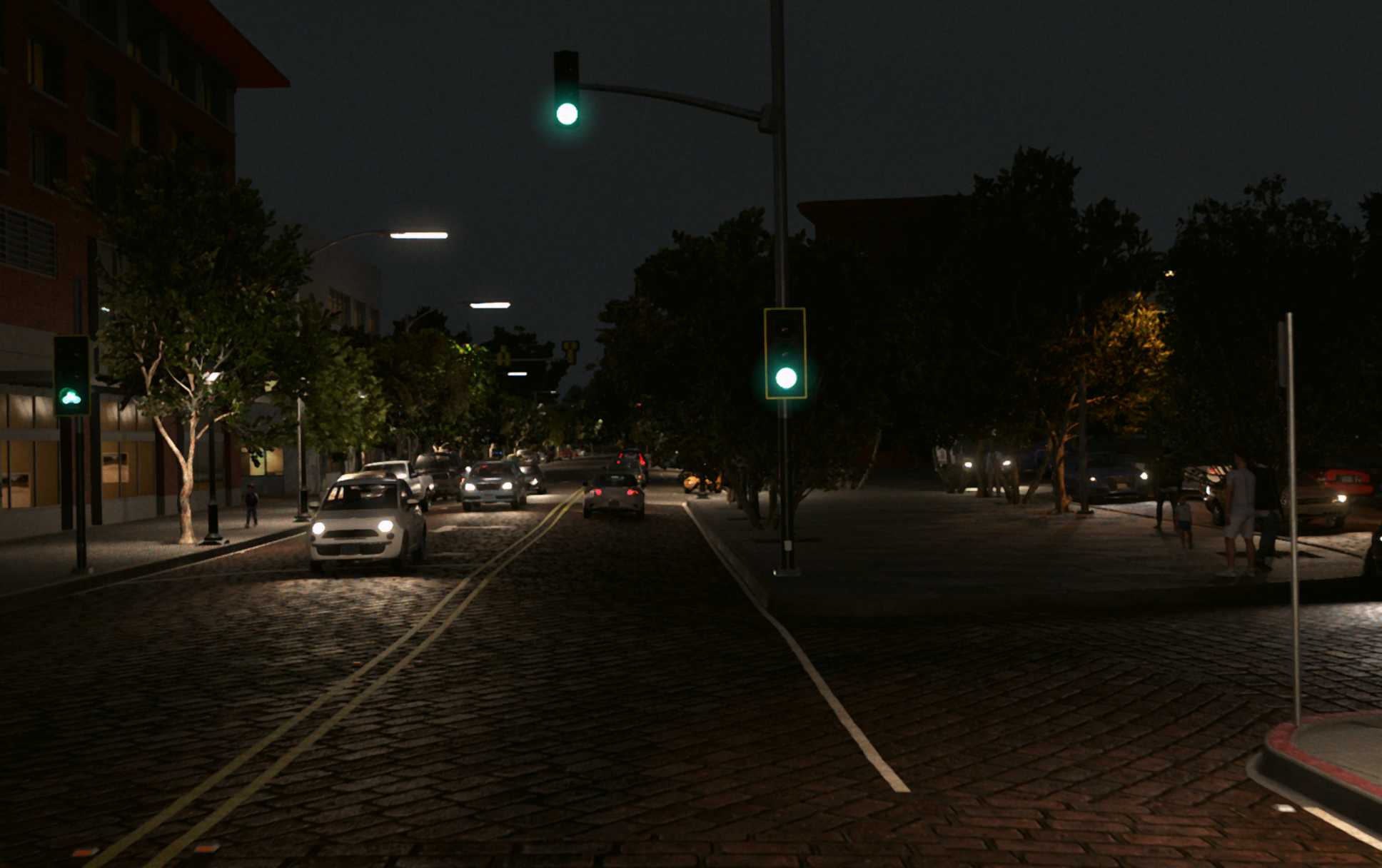}
\includegraphics[width=0.2\textwidth]{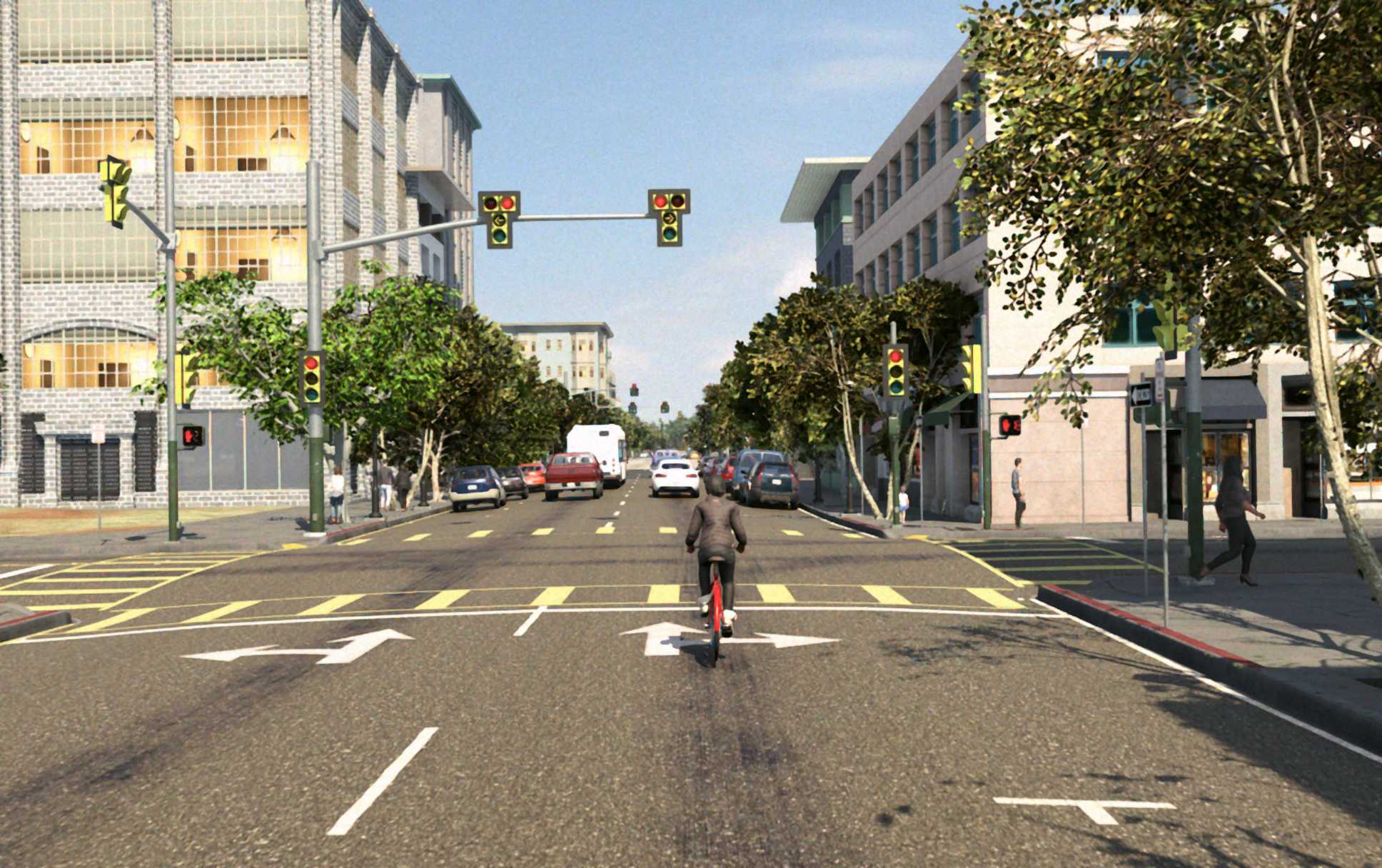}
\includegraphics[width=0.2\textwidth]{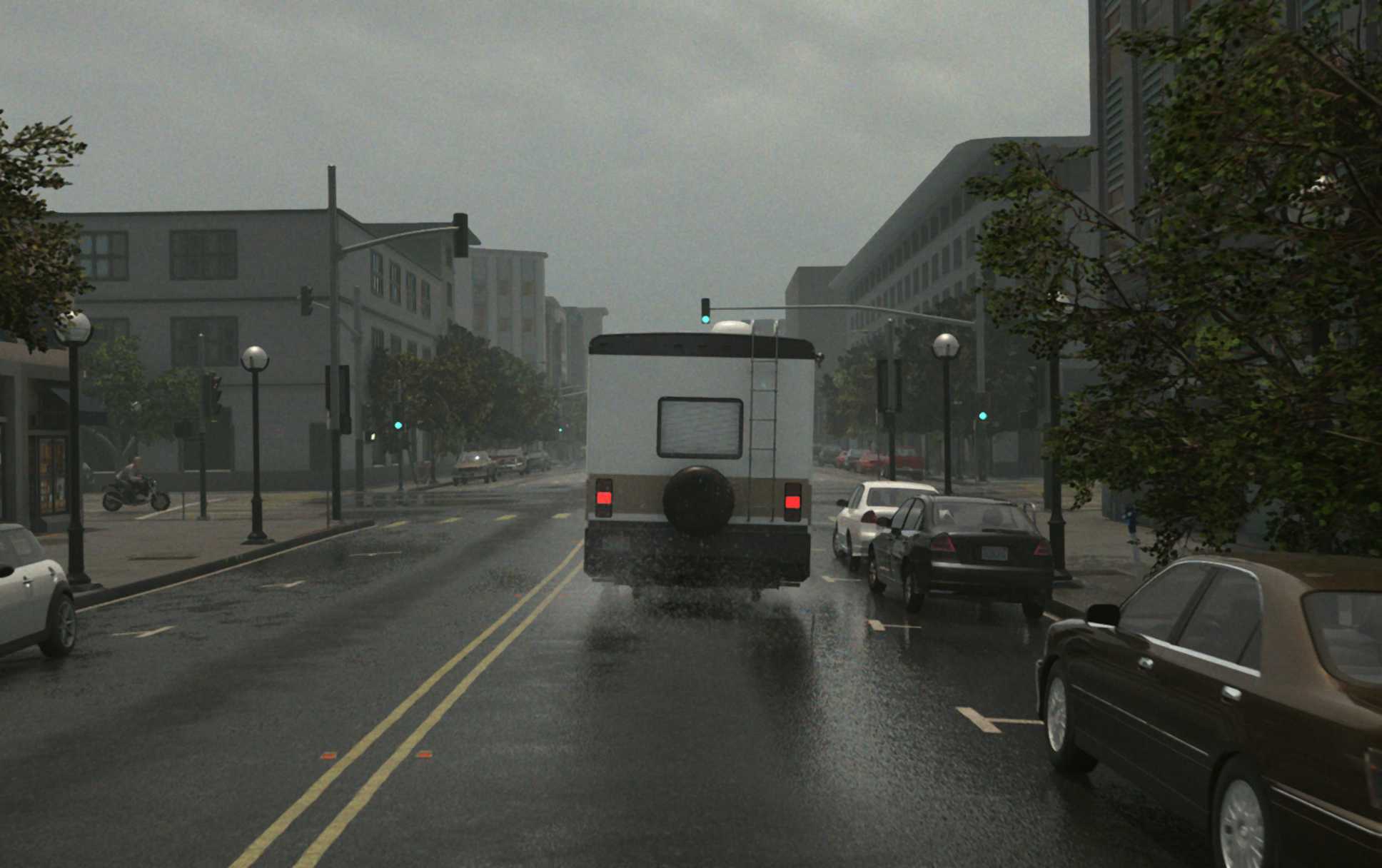}
\includegraphics[width=0.2\textwidth]{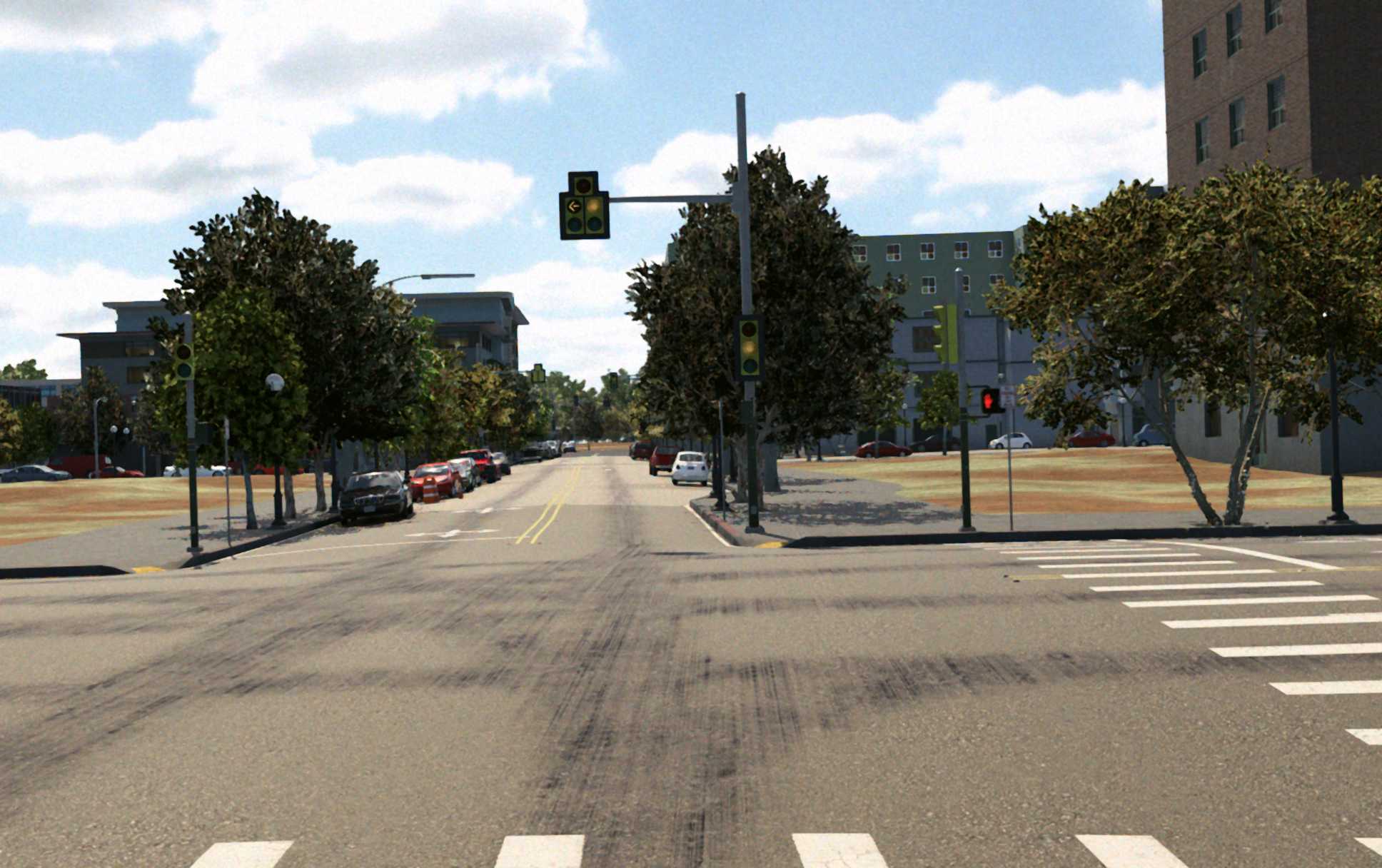}
}
    % \centering
    % \includegraphics[width=0.95\textwidth]{example-image-a}
    \caption{\textbf{The \emph{Parallel Domain} dataset}: sample images.}
    \label{fig:supplementary-pd-viz1}
\end{figure*}

%% file: suppmat/posenet.tex
\begin{table}[!t]%
\small
\centering
\resizebox{0.9\linewidth}{!}{
\begin{tabular}[b]{l|l|c}
\toprule
& \textbf{Layer Description} & \textbf{Out. Dimension} \\ 
\toprule
& 2 Stacked RGB images & 6$\times$H$\times$W \\ 
\midrule
\multicolumn{3}{c}{\textbf{ResNet18 Encoder}} \\
\midrule
\#1 & Latent Features & 256$\times$H/8$\times$W/8 \\
\midrule
\multicolumn{3}{c}{\textbf{Pose Decoder}}  \\
\midrule
\#2 & Conv2d $\rightarrow$ ReLU & 256$\times$H/8$\times$W/8  \\
\#3 & Conv2d $\rightarrow$ ReLU & 256$\times$H/8$\times$W/8  \\
\#4 & Conv2d $\rightarrow$ ReLU & 256$\times$H/8$\times$W/8  \\
\textbf{\#5} & Conv2d $\rightarrow$ Global Pooling & 6       \\
\bottomrule
\end{tabular}
}
\\
%%%%%%%%%%%%%%%%%%%%%%%%%%%%%%%%%%%%%%%%%%%%%%%%%%%
\caption{
\textbf{Pose network}. Two concatenated RGB images are used as input for a ResNet18 encoder (the first convolutional layer is duplicated to account for that). The output is a 6-dimensional vector estimating the rigid transformation between frames (translation and rotation in Euler angles).
}
%%%%%%%%%%%%%%%%%%%%%%%%%%%%%%%%%%%%%%%%%%%%%%%%%%%
\label{tab:posenet}
\end{table}

%% file: suppmat/qualitative.tex
\begin{figure*}[ht!]
\includegraphics[width=0.49\textwidth,height=4.0cm]{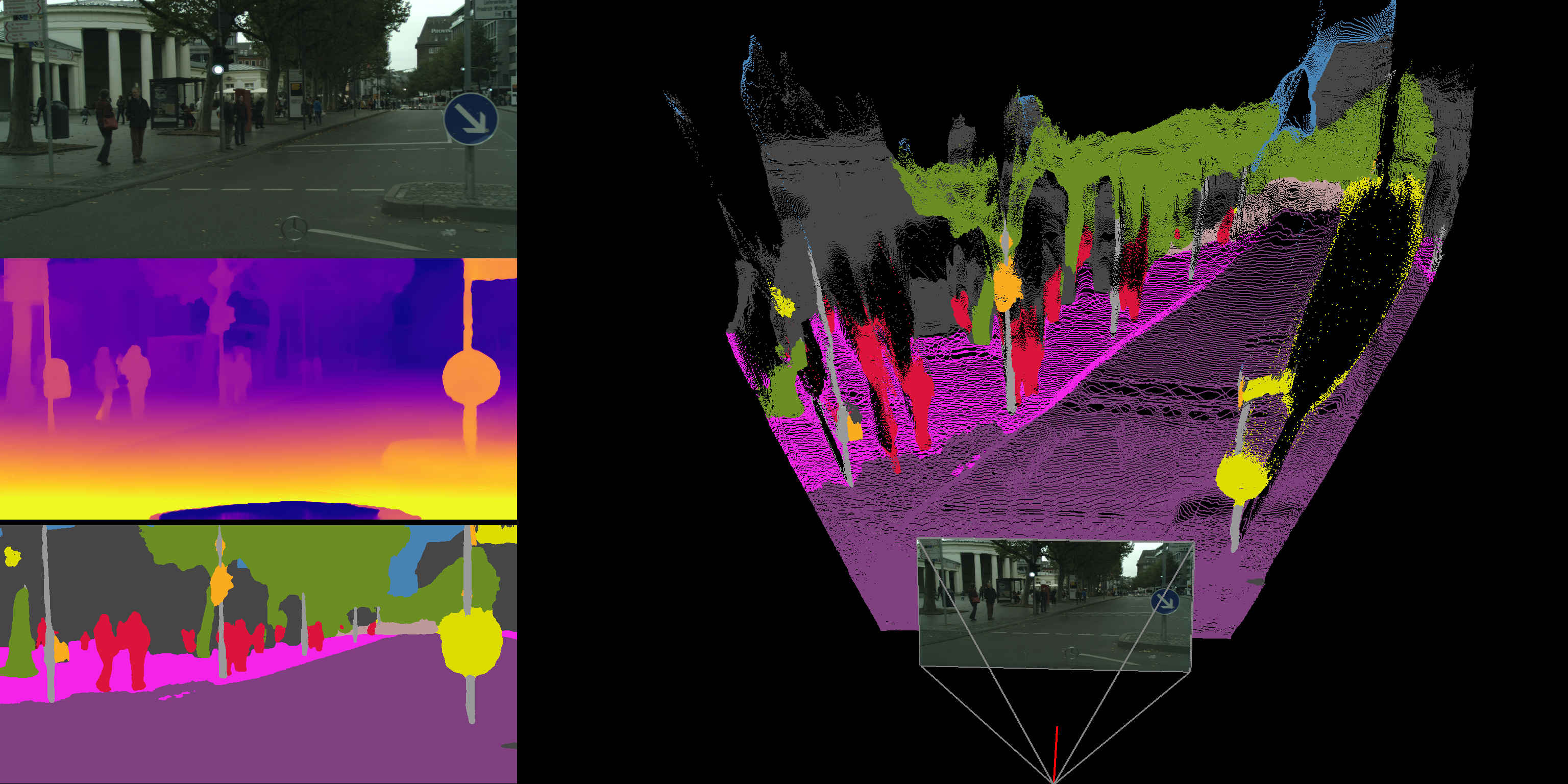}
\includegraphics[width=0.49\textwidth,height=4.0cm]{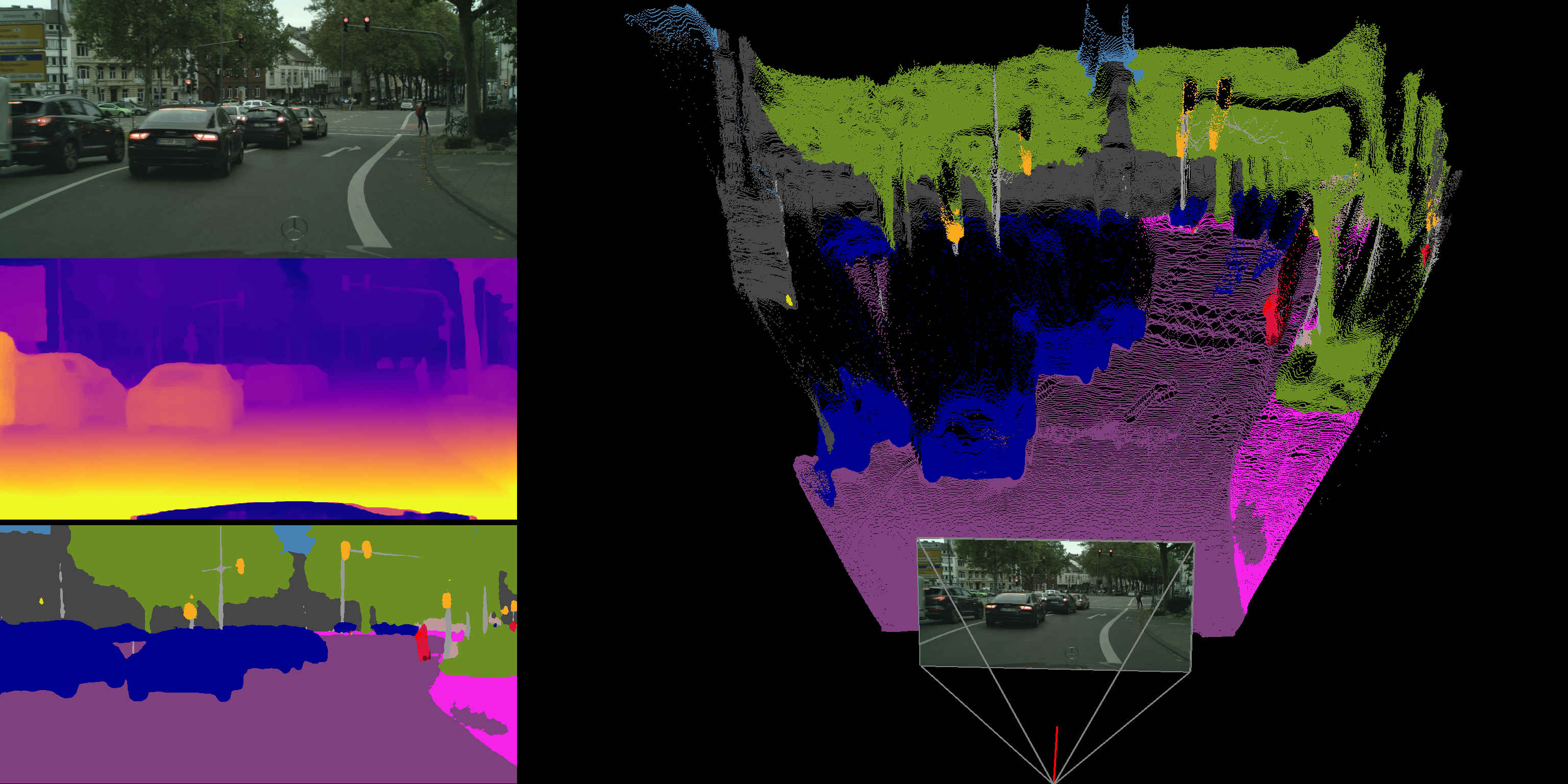}
\\
\includegraphics[width=0.49\textwidth,height=4.0cm]{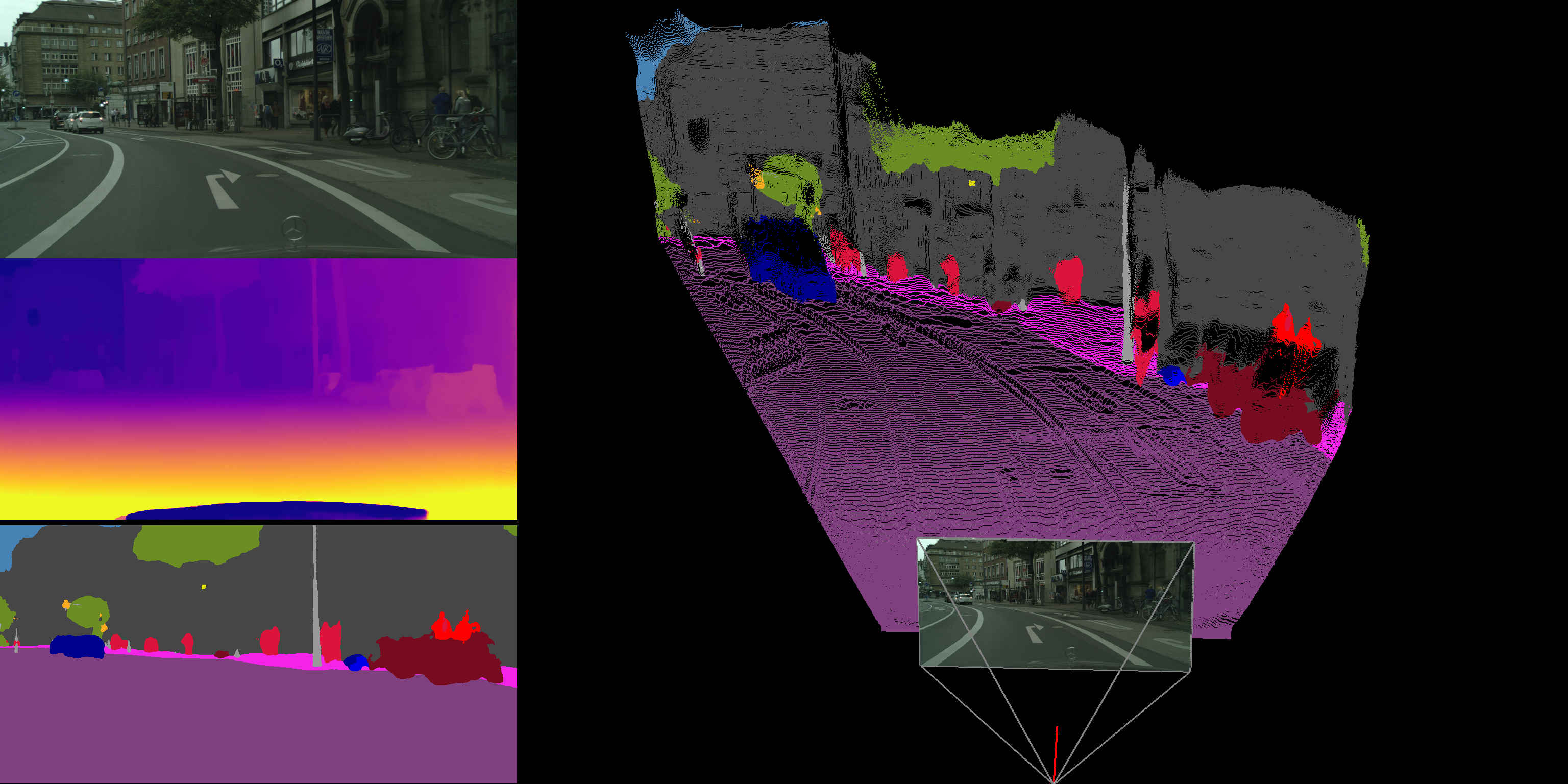}
\includegraphics[width=0.49\textwidth,height=4.0cm]{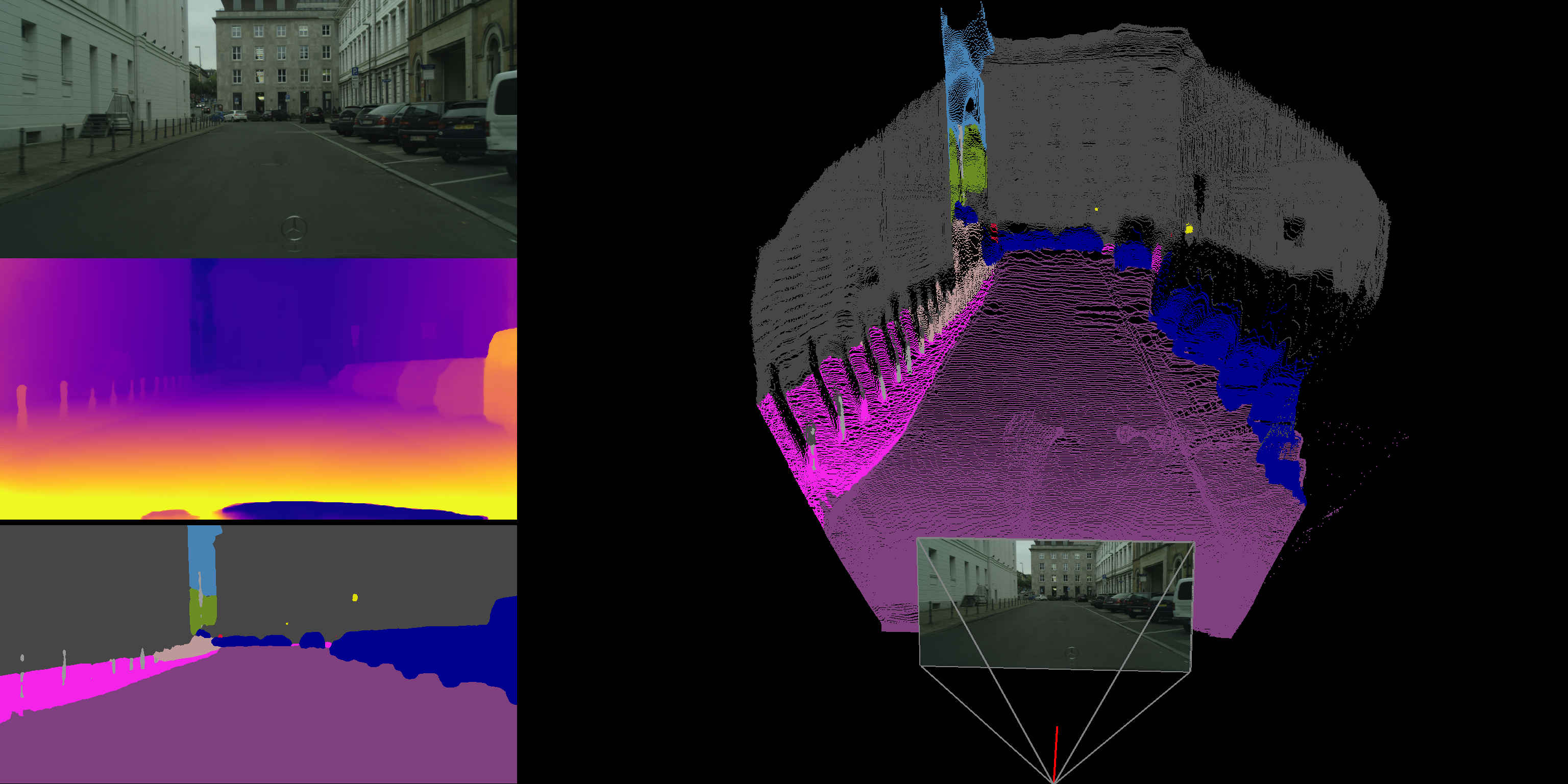}
\\ 
\includegraphics[width=0.49\textwidth,height=4.0cm]{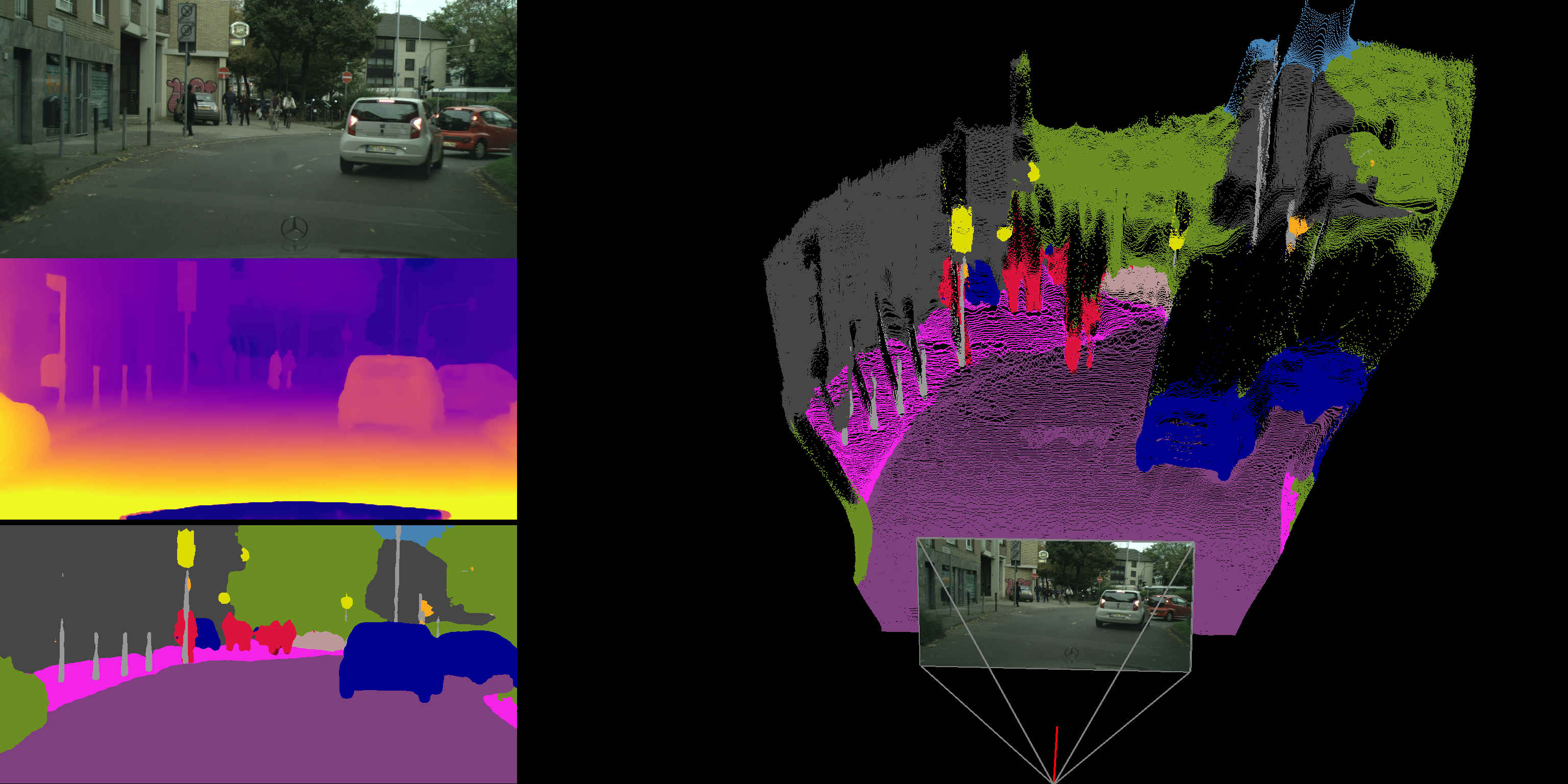}
\includegraphics[width=0.49\textwidth,height=4.0cm]{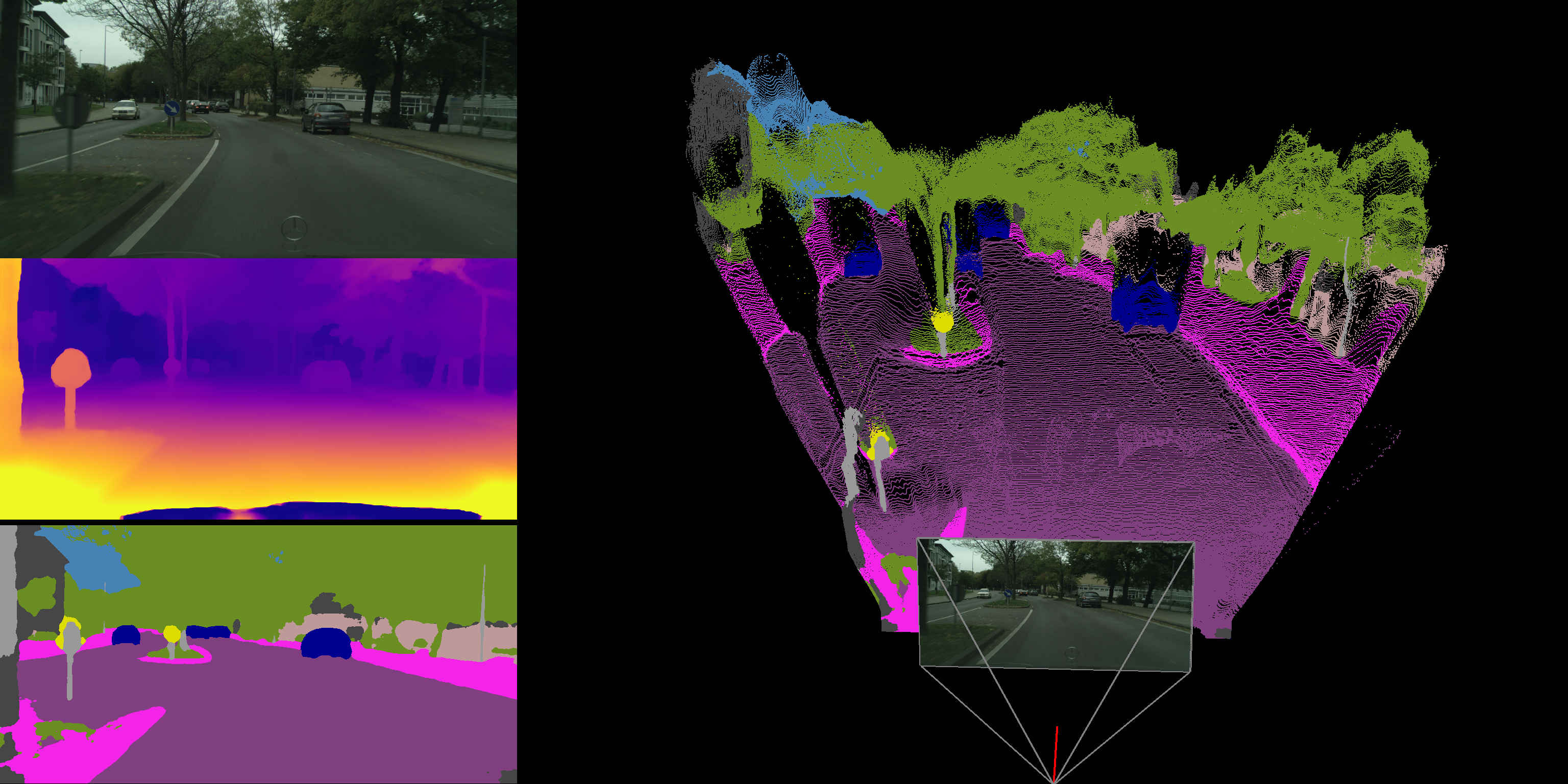}
\\ 
\includegraphics[width=0.49\textwidth,height=4.0cm]{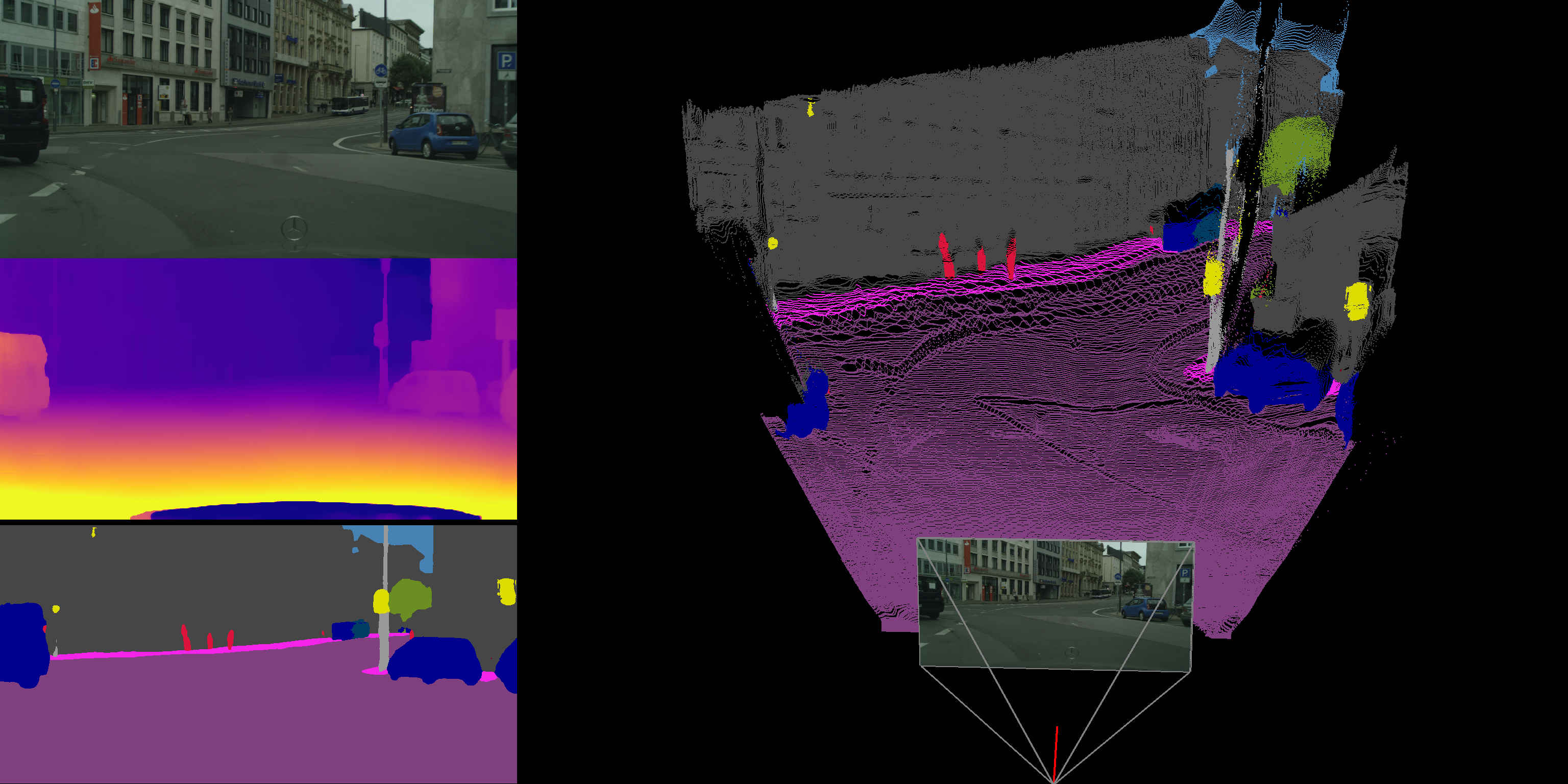}
\includegraphics[width=0.49\textwidth,height=4.0cm]{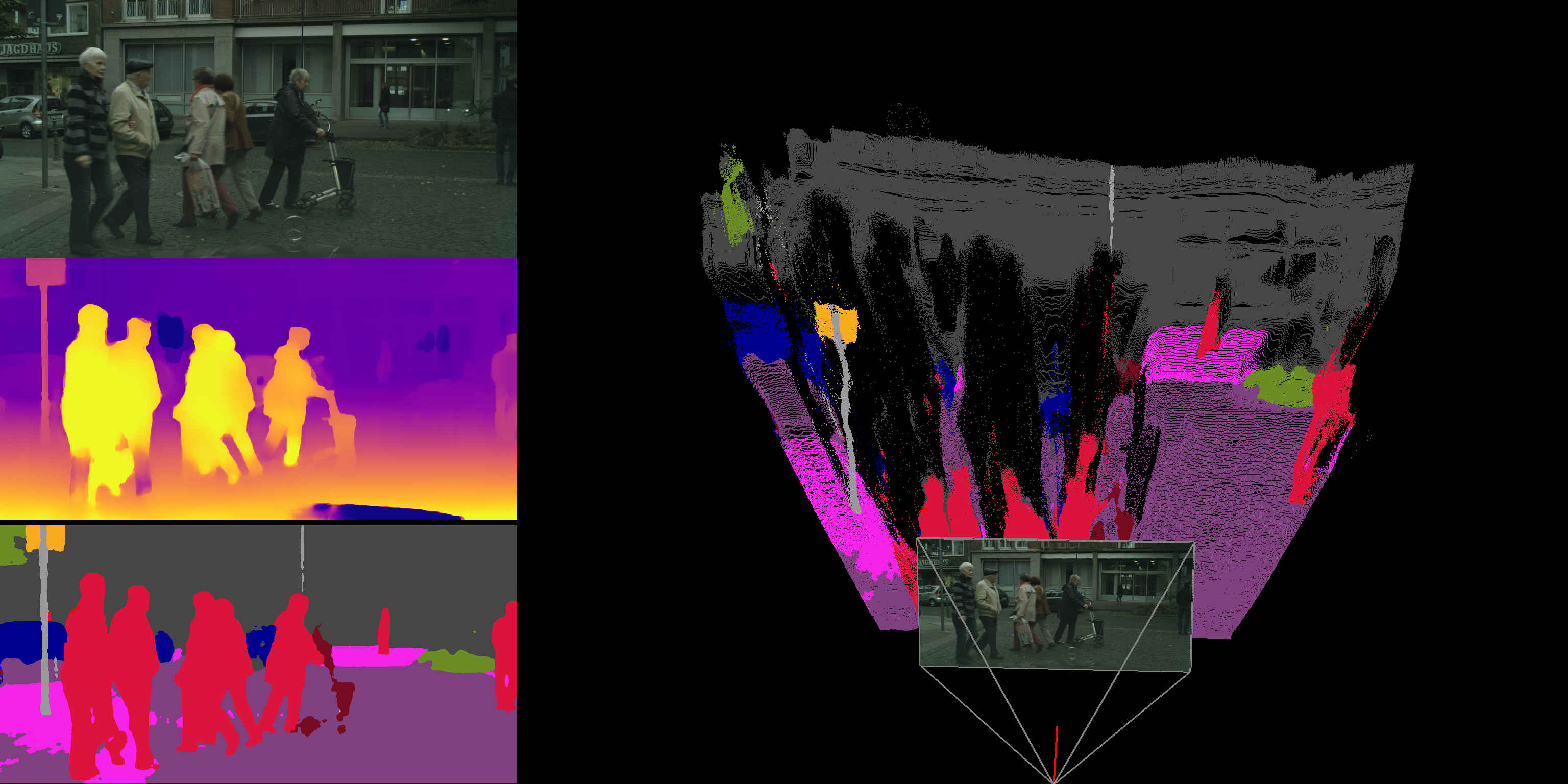}
\\ 
\includegraphics[width=0.49\textwidth,height=4.0cm]{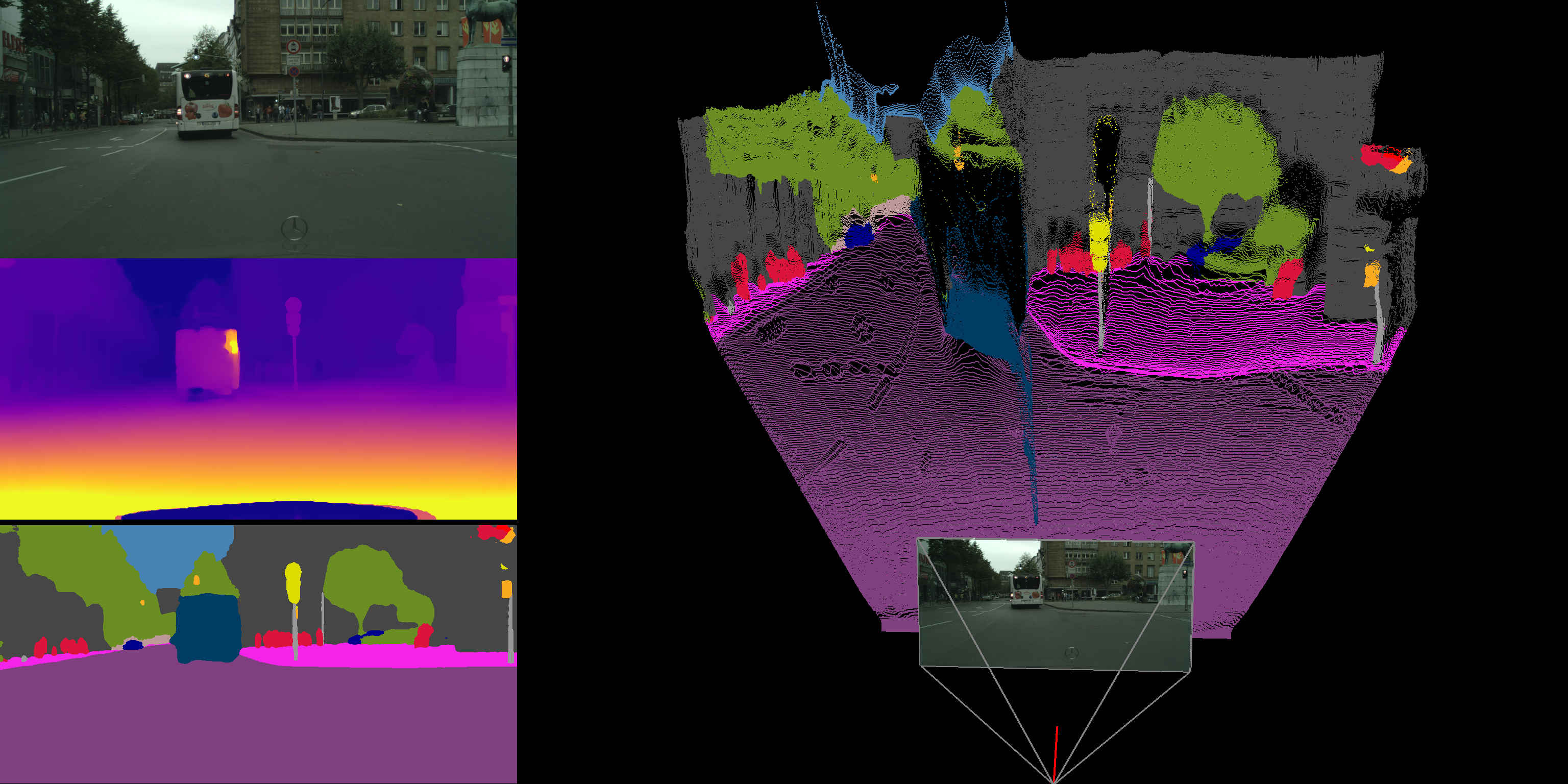}
\includegraphics[width=0.49\textwidth,height=4.0cm]{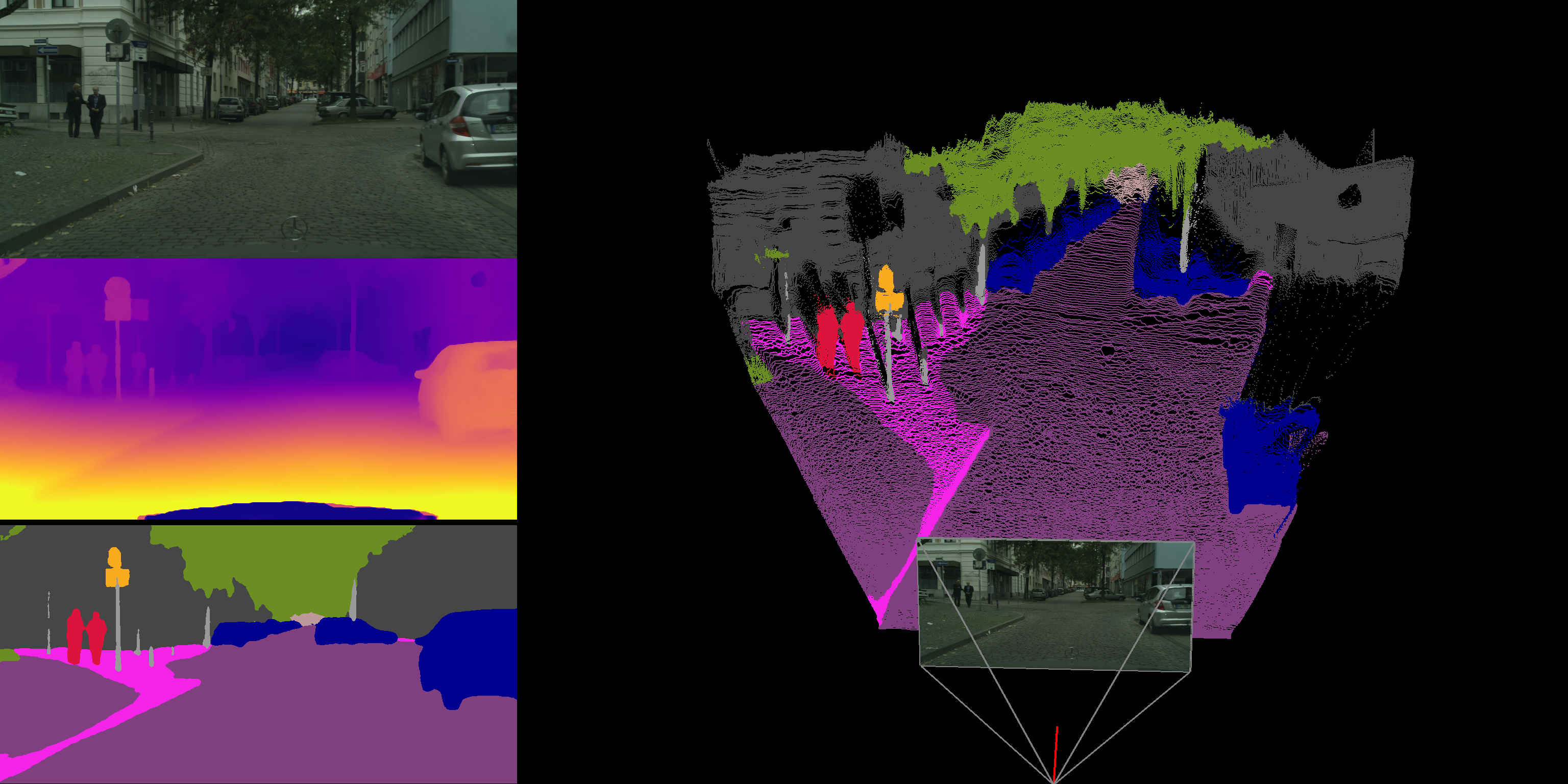}
\caption{\textbf{Qualitative depth and semantic segmentation results}, using GUDA+PL to perform unsupervised domain adaptation from \emph{Parallel Domain} to \emph{Cityscapes}. The same multi-task network was used to generate depth and semantic segmentation estimates, that were combined into a 3D pointcloud using camera intrinsics. No real-world labels (depth or semantic) were used during training.}
\label{fig:qualitative}
\end{figure*}

%% file: suppmat/pd.tex
This dataset is procedurally generated using the \textit{Parallel Domain} synthetic data generation service~\cite{parallel_domain}. It contains 5000 10-frame sequences, for a total of $50000$ frames. Each frame consists of an RGB image from a front-facing vehicle-mounted camera along with associated per-pixel depth and semantic segmentation labels. The dataset consists of urban and highway environments with varying number of agents, time of day, and weather conditions. We present reference images from the dataset in Fig.~\ref{fig:supplementary-pd-viz1}. Each image is rendered with a $1936\times1216$ resolution. The high degree of fidelity and perceptual quality allows us to investigate the following questions: (i) how does the quality of the simulation affect the \emph{sim-to-real} domain gap; and (ii) can we decrease the \emph{sim-to-real} domain gap with additional synthetic data. As reported in the main paper, Tab. $\color{red}1$ and Fig. $\color{red}7$, we conclude that high quality synthetic data can indeed help narrow the \textit{sim-to-real} gap, and the gap is further narrowed as additional data is made available.

%% file: tables/semantic_synthia_cs.tex
\begin{table*}[t!]
\renewcommand{\arraystretch}{0.9}
\centering
{
\small
\setlength{\tabcolsep}{0.35em}
\begin{tabular}{l||c|c|c|c|c|c|c|c|c|c|c|c|c|c|c|c||c|c}
\toprule
\multirow{3}{*}{\textbf{Method}}
& \multirow{3}{*}{\rotatebox{90}{Road}}
& \multirow{3}{*}{\rotatebox{90}{S.walk}}
& \multirow{3}{*}{\rotatebox{90}{Build.}} 
& \multirow{3}{*}{\rotatebox{90}{Wall*}}
& \multirow{3}{*}{\rotatebox{90}{Fence*}}
& \multirow{3}{*}{\rotatebox{90}{Pole*}}
& \multirow{3}{*}{\rotatebox{90}{T.Light}}
& \multirow{3}{*}{\rotatebox{90}{T.Sign}}
& \multirow{3}{*}{\rotatebox{90}{Vegt.}}
& \multirow{3}{*}{\rotatebox{90}{Sky}}
& \multirow{3}{*}{\rotatebox{90}{Person}}
& \multirow{3}{*}{\rotatebox{90}{Rider}}
& \multirow{3}{*}{\rotatebox{90}{Car}}
& \multirow{3}{*}{\rotatebox{90}{Bus}}
& \multirow{3}{*}{\rotatebox{90}{Motor.}}
& \multirow{3}{*}{\rotatebox{90}{Bike}}
& \multirow{3}{*}{\rotatebox{90}{mIoU}}
& \multirow{3}{*}{\rotatebox{90}{mIoU*}}
\\
& & & & & & & & & & & & & & & & & &
\\
& & & & & & & & & & & & & & & & & &
\\
% \midrule
%\parbox[t]{2mm}{\multirow{17}{*}{\rotatebox[origin=c]{90}{}}}

% \bottomrule
% \multicolumn{10}{l}{\textbf{asdfasdfasf}}
\toprule
Source (SY)
& 70.2 & 35.0 & 74.7 & 2.1 & 0.2 & 27.8 & 1.7 & 4.4 & 76.9 & 83.4 & 44.4 & 9.9 & 51.3 & 7.9 & 4.0 & 12.8 & 31.7 & 36.7 \\
Source (PD)
& 85.5 & 39.4 & 70.6 & 0.0 & 0.8 & 37.6 & 25.4 & 11.9 & 79.9 & 80.9 & 47.0 & 25.0 & 70.1 & 10.7 & 9.8 & 15.3 & 38.1 & 44.0 \\
Target 
& 97.1 & 82.9 & 90.6 & 47.3 & 51.7 & 57.1 & 60.8 & 72.5 & 91.6 & 93.3 & 75.8 & 54.3 & 93.4 & 77.5 & 48.5 & 71.9 & 72.9 & 77.8 \\

% \toprule
\toprule
\multicolumn{19}{l}{\textbf{(a) Comparison with other depth-based UDA methods (SYNTHIA $\rightarrow$ Cityscapes)}} \\
\toprule

% Source 
% & 70.2 & 35.0 & 74.7 & 2.1 & 0.2 & 27.8 & 1.7 & 4.4 & 76.9 & 83.4 & 44.4 & 9.9 & 51.3 & 7.9 & 4.0 & 12.8 & 31.7 & 36.7 \\

% Source (PD)
% & 80.3 & 46.8 & 62.8 & 0.0 & 0.8 & 34.2 & 29.8 & 11.9 & 76.4 & 83.0 & 45.3 & 22.4 & 75.1 & 8.1 & 9.7 & 13.2 & 37.5 & 43.4 \\

% \cmidrule{1-19}
% adversarial +  depth: cycle-gan plue depth as auxilary task 
SPIGAN \cite{spigan}     
& 71.1 & 29.8 & 71.4 & 3.7 & \underline{0.3} & 33.2 & 6.4 & 15.6 & 81.2 & 78.9 & 52.7 & 13.1  & 75.9 & 25.5 & \underline{10.0} & 20.5 & 36.8 & 42.4 \\
% adversarial + depth: using depth to adapt the input in GAN
GIO-Ada~\cite{Chen_2019_CVPR} & 78.3 & 29.2 & 76.9 & 11.4 & \underline{0.3} & \underline{26.5} & \underline{10.8} & \underline{17.2} &\underline{81.7} & 81.9 & 45.8 & \underline{15.4} & 68.0 & 15.9 & 7.5 & 30.4 & 37.3 & 43.0 \\
% adversarial + depth: depth+segmentation+feature fusion network. adversarial at output.
DADA \cite{vu2019dada}     
& \textbf{89.2} & \underline{44.8} & \textbf{81.4} & 6.8 & \underline{0.3} & 26.2 & 8.6 & 11.1 & \textbf{81.8} & \underline{84.0} & \underline{54.7} & \textbf{19.3} & \underline{79.7} & \textbf{40.7} & \textbf{14.0} & \textbf{38.8} & \underline{42.6} & \underline{49.8} \\

\cmidrule{1-19}

\textbf{GUDA} 
& \underline{85.4} & \textbf{49.5} & \underline{80.8} & \textbf{13.8} & \textbf{0.9} & \textbf{36.2} & \textbf{21.8} & \textbf{35.2} & 78.8 & \textbf{84.7} & \textbf{59.9} & 13.5 & \textbf{84.0} & \underline{33.8} & 2.8 & \underline{30.9} & \textbf{44.5} & \textbf{50.9} \\

% \cmidrule{1-19}

% Target 
% & 97.1 & 82.9 & 90.6 & 47.3 & 51.7 & 57.1 & 60.8 & 72.5 & 91.6 & 93.3 & 75.8 & 54.3 & 93.4 & 77.5 & 48.5 & 71.9 & 72.9 & 77.8 \\

\toprule
\multicolumn{19}{l}{\textbf{(b) Comparison with other UDA methods (SYNTHIA $\rightarrow$ Cityscapes)}} \\
\toprule

% Source 
% & 70.2 & 35.0 & 74.7 & 2.1 & 0.2 & 27.8 & 1.7 & 4.4 & 76.9 & 83.4 & 44.4 & 9.9 & 51.3 & 7.9 & 4.0 & 12.8 & 31.7 & 36.7 \\

% \cmidrule{1-19}

%AdaptPatch \cite{adapt_patch}
%& 82.2 & 39.4 & 79.4 & --- & --- & --- & 9.9 & 10.5 & 78.2 & 80.5 & 53.5 & 19.6  & 67.0 & 29.5 & 21.6 & 31.3 & --- & 45.9 \\
% self-supervision for computer vision tasks
Xu et al. \cite{xu2019self}     
& --- & --- & --- & --- & --- & --- & --- & --- & ---& --- & --- & --- & --- & --- &--- & --- & 38.8 & --- \\
CLAN \cite{clan}       
& 81.3 & 37.0 & 80.1 & --- & --- & --- & 16.1 & 13.7 & 78.2 & 81.5 & 53.4 & 21.2 & 73.0  & 32.9 & 22.6 & 30.7 & --- & 47.8 \\
%AdvEnt \cite{vu2018advent}
%& 87.0 & 44.1 & 79.7 & 9.6 & 0.6 & 24.3 & 4.8 & 7.2 & 80.1 & 83.6 & 56.4 & 23.7 & 72.7  & 32.6 & 12.8 & 33.7 & 40.8 & 47.6 \\
% CBST:class balanced self-training, spacial priors, psudo-label
CBST \cite{cbst}&53.6& 23.7 &75.0& 12.5& 0.3& 36.4& 23.5& 26.3& 84.8& 74.7& 67.2& 17.5& 84.5 &28.4& 15.2& 55.8 &42.5& 48.4\\
% Learning from Scale-Invariant Example
% Self-training, pseudo labels are used. Training on original and scaled test image and penalized the difference.
% CAG \cite{cag}
% & 82.9 & 43.1 & 78.1 & 9.3 & 0.6 & 28.2 & 9.1 & 14.4 & 77.0 & 83.5 & 58.1 & 25.9 & 71.9 & 38.0 & 29.4 & 31.2 & 42.6 & 49.4 \\
% confidence regularized self-training.
CRST \cite{crst}     
& 67.7 & 32.2 & 73.9 & 10.7 & \underline{1.6} & \underline{37.4} & 22.2 & \textbf{31.2} & 80.8 & 80.5 & 60.8 & 29.1 & 82.8 & 25.0 & 19.4 & 45.3 & 43.8 & 50.1 \\
ESL\cite{Saporta2020ESLES}     
&84.3 &39.7 &79.0& 9.4 &0.7& 27.7& 16.0 &14.3& 78.3& 83.8& 59.1& 26.6& 72.7 &35.8& 23.6 &45.8 &43.5 &50.7\\
% 
% Using Fourier transformation as input style transfer, no adversarial learning.
FDA \cite{fda}      
& 79.3 & 35.0 & 73.2 & --- & --- & --- & 19.9 & 24.0 & 61.7 & 82.6 & 61.4 & 31.1 & 83.9 & 40.8 & \textbf{38.4} & 51.1 & --- & 52.5 \\
% using pseudo label to select source image with less domain gap. Then fine-tune model with selected images. self-training paradigm.
CCMD \cite{ccmd}     
& 79.6 & 36.4 & 80.6 & 13.3 & 0.3 & 25.5 & 22.4 & 14.9 & 81.8 & 77.4 & 56.8 & 25.9 & 80.7 & 45.3 & \underline{29.9} & 52.0 & 45.2 & 52.6 \\

Yang et al. \cite{Yang2020LabelDrivenRF}     
& 85.1& 44.5& 81.0 & --- & --- & --- &16.4 &15.2& 80.1 &84.8& 59.4 &31.9& 73.2 &41.0& 32.6& 44.7 &53.1 \\

% Regularize the model using auxilary classifier. 
% can we call it self-supervised on target domain?
USAMR \cite{usamr}      
& 83.1 & 38.2 & 81.7 & 9.3 & 1.0 & 35.1 & \underline{30.3} & 19.9 & 82.0 & 80.1 & 62.8 & 21.1 & 84.4 & 37.8 & 24.5 & \underline{53.3} & 46.5 & 53.8 \\
IAST \cite{iast}     
& 81.9 & 41.5 & \underline{83.3} & \underline{17.7} & \textbf{4.6} & 32.3 & \textbf{30.9} & \underline{28.8} & \textbf{83.4} & \textbf{85.0} & \underline{65.5} & \textbf{30.8} & \underline{86.5} & 38.2 & \underline{33.1} & 52.7 & \underline{49.8} & \underline{57.0} \\

\cmidrule{1-19}

% \cmidrule{1-19}

% \textbf{GUDA (PD)} 
% & 94.3 & 70.3 & 84.1 & 0.00 & 12.4 & 46.4 & 45.5 & 42.4 & 84.7 & 85.4 & 64.7 & 30.1 & 86.9 & 36.9 & 9.6 & 13.5 & 50.5 & 57.6 \\

% \cmidrule{1-19}

\textbf{GUDA+PL} 
& \textbf{88.1} & \textbf{53.0} & \textbf{84.0} & \textbf{22.0} & 1.4 & \textbf{39.6} & 28.2 & 24.8 & \underline{82.7} & 81.5 & \textbf{65.5} & 22.7 & \textbf{89.3} & \textbf{50.5} & 25.1 & \textbf{57.5} & \textbf{51.0} & \textbf{57.9} \\

% \cmidrule{1-19}

% Target 
% & 97.1 & 82.9 & 90.6 & 47.3 & 51.7 & 57.1 & 60.8 & 72.5 & 91.6 & 93.3 & 75.8 & 54.3 & 93.4 & 77.5 & 48.5 & 71.9 & 72.9 & 77.8 \\

\toprule
\multicolumn{19}{l}{\textbf{(c) Comparison with the state of the art (Varying Sources $\rightarrow$ Cityscapes)}} \\
\toprule

% Source(PD)
% & 85.5 & 39.4 & 70.6 & 0.0 & 0.8 & 37.6 & 25.4 & 11.9 & 79.9 & 80.9 & 47.0 & 25.0 & 70.1 & 10.7 & 9.8 & 15.3 & 38.1 & 44.0 \\

% \cmidrule{1-19}
% UDA through self-supervised. GTA-> cs, CYCADA + self-supervision
UDAS \cite{sun2019unsupervised}  &
86.6& 37.8& 80.8& 29.7&16.4 &28.9&30.9 &22.2 &37.1 &76.9&60.1 &7.8 &84.1&  32.1&  23.2& 13.3&44.3
 & 49.2\\
USAMR (G5) \cite{usamr}      
& 90.5 & 35.0 & 84.6 & \underline{34.3} & \underline{24.0} & \underline{36.8} & \underline{44.1} & \underline{42.7} & 84.5 & 82.5 & \underline{63.1} & \underline{34.4} & 85.8 & 38.2 & 27.1 & \underline{41.8} & 53.1 & 58.0 \\
IAST (G5) \cite{iast}      
& \textbf{94.1} & \textbf{58.8} & \underline{85.4} & \textbf{39.7} & \textbf{29.2} & 25.1 & 43.1 & 34.2 & \underline{84.8} & \textbf{88.7} & 62.7 & 30.3 & \underline{87.6} & \underline{50.3} & \textbf{35.2} & 40.2 & \underline{55.6} & \underline{61.2} \\

\cmidrule{1-19}

\textbf{GUDA(PD)+PL(G5)} 
& \underline{92.9} & \underline{50.5} & \textbf{86.0} & 17.9 & \underline{24.0} & \textbf{45.4} & \textbf{50.9} & \textbf{44.5} & \textbf{87.7} & \underline{87.0} & \textbf{66.6} & \textbf{36.9} & \textbf{89.5} & \textbf{52.1} & \underline{28.5} & \textbf{54.0} & \textbf{57.2} & \textbf{63.2} \\

% \toprule
% \toprule

% Target 
% & 97.1 & 82.9 & 90.6 & 47.3 & 51.7 & 57.1 & 60.8 & 72.5 & 91.6 & 93.3 & 75.8 & 54.3 & 93.4 & 77.5 & 48.5 & 71.9 & 72.9 & 77.8 \\

\bottomrule

\end{tabular}
}
\caption{\textbf{Semantic segmentation results on \emph{Cityscapes}} using different unsupervised domain adaptation (UDA) methods and synthetic datasets. The \emph{mIoU} metric considers all 16 classes, and \emph{mIoU*} considers only the 13 classes present in SYNTHIA (removing the ones marked with \emph{*}). \emph{Source} shows results without any adaptation, and \emph{Target} shows results with semantic supervision on the target domain. Synthetic datasets include: \emph{SYNTHIA (SY)}, \emph{Parallel Domain (PD)}, and \emph{GTA5 (G5)}.} 
\label{tab:synthia_cs}
% \vspace{-3mm}
\end{table*}

%% file: tables/vkitti2_kitti.tex
\begin{table*}[t!]
\vspace{5mm}
\centering
{
\small
\setlength{\tabcolsep}{0.45em}
\begin{tabular}{l|c|c|c|c|c|c|c|c|c|c|c}
\toprule
\multirow{3}{*}{\textbf{Method}} 
& \multirow{3}{*}{\rotatebox{90}{Road}}
& \multirow{3}{*}{\rotatebox{90}{Building}}
& \multirow{3}{*}{\rotatebox{90}{Pole}}
& \multirow{3}{*}{\rotatebox{90}{T. Light}}
& \multirow{3}{*}{\rotatebox{90}{T. Sign}}
& \multirow{3}{*}{\rotatebox{90}{Vegetat.}}
& \multirow{3}{*}{\rotatebox{90}{Terrain}}
& \multirow{3}{*}{\rotatebox{90}{Sky}}
& \multirow{3}{*}{\rotatebox{90}{Car}}
& \multirow{3}{*}{\rotatebox{90}{Truck}}
& \multirow{3}{*}{\rotatebox{90}{\textbf{mIoU}}}
\\
& & & & & & & & & & &
\\
& & & & & & & & & & &
\\
\midrule
Source 
& 64.9 & 28.3 & 37.8 & 18.8 & 11.7 & 63.7 & 21.6 & 78.7 & 55.3 & 1.5 & 38.6
\\ 
DANN  
& 70.3 & 49.4 & 39.5 & 28.0 & 22.2 & 67.0 & 23.1 & 82.0 & 69.4 & 5.1 & 45.6
\\
\textbf{GUDA} 
& \textbf{86.8} & \textbf{72.7} & \textbf{46.2} & \textbf{41.4} & \textbf{44.6} & \textbf{77.3} & \textbf{29.1} & \textbf{88.5} & \textbf{86.1} & \textbf{9.8} & \textbf{58.25}
\\
\bottomrule

\end{tabular}
}
\caption{\textbf{Semantic segmentation results on \textit{VKITTI2} $\rightarrow$ \textit{KITTI}}, using GUDA and DANN \cite{dann}.}
\label{tab:vkitti2_kitti}
\end{table*}

%% file: tables/pd_ddad.tex
\begin{table*}[t!]
\vspace{5mm}
\centering
{
\small
\setlength{\tabcolsep}{0.45em}
\begin{tabular}{l|c|c|c|c|c|c|c|c|c|c|c|c|c|c|c|c}
\toprule
\multirow{3}{*}{\textbf{Method}} 
& \multirow{3}{*}{\rotatebox{90}{Road}}
& \multirow{3}{*}{\rotatebox{90}{S.walk}}
& \multirow{3}{*}{\rotatebox{90}{Build.}}
& \multirow{3}{*}{\rotatebox{90}{Pole}}
& \multirow{3}{*}{\rotatebox{90}{T.Light}}
& \multirow{3}{*}{\rotatebox{90}{T.Sign}}
& \multirow{3}{*}{\rotatebox{90}{Vegetat.}}
& \multirow{3}{*}{\rotatebox{90}{Sky}}
& \multirow{3}{*}{\rotatebox{90}{Person}}
& \multirow{3}{*}{\rotatebox{90}{Rider}}
& \multirow{3}{*}{\rotatebox{90}{Car}}
& \multirow{3}{*}{\rotatebox{90}{Truck}}
& \multirow{3}{*}{\rotatebox{90}{Bus}}
& \multirow{3}{*}{\rotatebox{90}{Motor.}}
& \multirow{3}{*}{\rotatebox{90}{Bike}}
& \multirow{3}{*}{\rotatebox{90}{\textbf{mIoU}}}
\\
& & & & & & & & & & & & & & &
\\
& & & & & & & & & & & & & & &
\\
\midrule
Source 
& 93.9 & 30.7 & 49.3 & 35.7 & 50.7 & 20.8 & 87.2 & 89.3 & 10.0 & 28.7 & 63.2 & 38.4 & 14.3 & 8.5 & 7.3 & 41.9
\\ 
DANN  
& 95.3 & 36.1 & 53.0 & 35.6 & 52.8 & 20.7 & 88.3 & 90.3 & 15.2 & 38.7 & 67.5 & 44.1 & 36.5 & 19.5 & 11.1 & 47.0
\\
\textbf{GUDA} 
& \textbf{96.1} & \textbf{48.0} & \textbf{58.9} & \textbf{37.1} & \textbf{55.8} & \textbf{22.0} & \textbf{89.6} & \textbf{93.0} & \textbf{30.6} & \textbf{54.8} & \textbf{70.8} & \textbf{47.2} & \textbf{58.7} & \textbf{41.6} & \textbf{29.6} & \textbf{55.6}
\\
\bottomrule

\end{tabular}
}
\caption{\textbf{Semantic segmentation results on \textit{Parallel Domain} $\rightarrow$ \textit{DDAD}}, using GUDA and DANN \cite{dann}.}
\label{tab:pd_ddad}
\end{table*}